\documentclass[10pt,twocolumn,letterpaper]{article} 
\usepackage[pagenumbers]{iccv}  

\definecolor{iccvblue}{rgb}{0.21,0.49,0.74}
\usepackage[pagebackref,breaklinks,colorlinks,allcolors=iccvblue]{hyperref}

\usepackage[linesnumbered,ruled,vlined]{algorithm2e}
\usepackage[utf8]{inputenc}    
\usepackage{amsmath,amssymb}   
\usepackage{pifont}           
\usepackage{graphicx}
\usepackage{multirow}
\usepackage{makecell}
\usepackage{colortbl}
\usepackage{adjustbox}
\usepackage{booktabs}  
\usepackage{caption}
\usepackage[accsupp]{axessibility}

\definecolor{lightyellow}{HTML}{FFFF00}
\definecolor{lightblue}{HTML}{BDD7EE}   
\definecolor{lightred}{HTML}{F8CBAD}    
\definecolor{lightgreen}{HTML}{C5E0B4}  
\definecolor{lightpurple}{HTML}{F3EAFD} 
\definecolor{lightgrey}{HTML}{D9D9D9}
\definecolor{darkred}{HTML}{8B0000}
\definecolor{darkgreen}{HTML}{006400}

\vspace{-10mm}
\title{DuET: \underline{Du}al Incremental Object Detection via \underline{E}xemplar-Free \underline{T}ask Arithmetic}

\author{
Munish Monga$^{1,2}$ \quad Vishal Chudasama$^1$ \quad Pankaj Wasnik$^{1,*}$ \quad Biplab Banerjee$^2$ \\ 
$^1$Sony Research India \quad $^2$Indian Institute of Technology, Bombay\\ 
{\tt\small \{munish.monga, vishal.chudasama1, pankaj.wasnik\}@sony.com, getbiplab@gmail.com} 
}

\newcommand{\cmark}{\ding{51}}
\newcommand{\xmark}{\ding{55}}
\newcommand{\deltaa}[1]{\ifnum #1>0 {\color{red}\textuparrow\ #1} \else \ifnum #1<0 {\color{blue}\textdownarrow\ #1} \else #1\fi\fi}

\newcommand{\greentick}{\textcolor{green}{\ding{51}}} 
\newcommand{\redcross}{\textcolor{red}{\ding{55}}}

\begin{document}
\twocolumn[{
    \renewcommand\twocolumn[1][]{#1} 
    \vspace{-6mm}
    \maketitle 
    \vspace{-12mm}
    \begin{center}
        \centering
        \includegraphics[width=\linewidth]{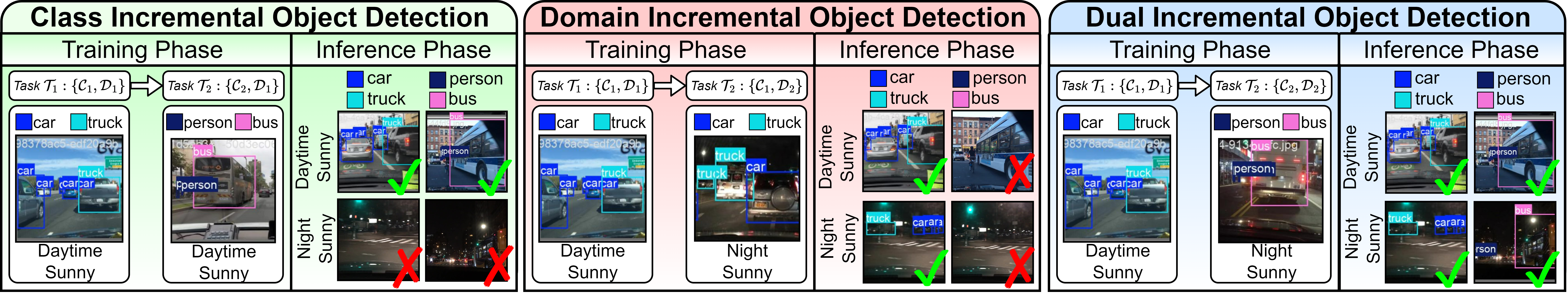}
        \vspace{-8mm}
        \captionof{figure}{\textbf{Comparison of Class Incremental Object Detection (CIOD), Domain Incremental Object Detection (DIOD), and our proposed Dual Incremental Object Detection (DuIOD), where \greentick~: the scenario is addressed, \redcross~: the scenario is NOT addressed.} \emph{\textbf{CIOD (left)}} incrementally learns new categories (\(\mathcal{C}_1 \rightarrow \mathcal{C}_2\)) in a fixed domain but struggles in unseen domains. \emph{\textbf{DIOD (middle)}}, adapts to new domains (\(\mathcal{D}_1 \rightarrow \mathcal{D}_2\)) but does not detect new classes. In contrast, \emph{\textbf{DuIOD (right)}} addresses both challenges by learning new object categories while generalizing across evolving domains (zoomed-in for best view).}
        \label{fig:teaser}
    \end{center}
}]
\renewcommand{\thefootnote}{\fnsymbol{footnote}}
\footnotetext[1]{Corresponding author}
\renewcommand{\thefootnote}{\arabic{footnote}}
\begin{abstract}
Real-world object detection systems, such as those in autonomous driving and surveillance, must continuously learn new object categories and simultaneously adapt to changing environmental conditions. Existing approaches, Class Incremental Object Detection (CIOD) and Domain Incremental Object Detection (DIOD)—only address one aspect of this challenge. CIOD struggles in unseen domains, while DIOD suffers from catastrophic forgetting when learning new classes, limiting their real-world applicability. To overcome these limitations, we introduce Dual Incremental Object Detection (DuIOD), a more practical setting that simultaneously handles class and domain shifts in an exemplar-free manner. We propose DuET, a Task Arithmetic-based model merging framework that enables stable incremental learning while mitigating sign conflicts through a novel Directional Consistency Loss. Unlike prior methods, DuET is detector-agnostic, allowing models like YOLO11 and RT-DETR to function as real-time incremental object detectors. To comprehensively evaluate both retention and adaptation, we introduce the Retention-Adaptability Index (RAI), which combines the Average Retention Index (Avg RI) for catastrophic forgetting and the Average Generalization Index for domain adaptability into a common ground. Extensive experiments on the Pascal Series and Diverse Weather Series demonstrate DuET’s effectiveness, achieving a +13.12\% RAI improvement while preserving 89.3\% Avg RI on the Pascal Series (4 tasks), as well as a +11.39\% RAI improvement with 88.57\% Avg RI on the Diverse Weather Series (3 tasks), outperforming existing methods.

\end{abstract}
\vspace{-1em}
\section{Introduction}
\label{Introduction}
Real-world object detection systems must continuously learn new categories and adapt dynamically to evolving environments without necessitating full retraining. Traditional models, typically trained on static datasets, struggle to perform effectively in dynamic real-world scenarios like autonomous driving and surveillance, where they must recognize novel objects while handling environmental variations. For example, an autonomous vehicle must detect newly introduced road signs under diverse lighting and weather conditions—without forgetting past knowledge, a challenge known as catastrophic forgetting \cite{LwF, cat-forget1, shmelkov2017incremental}.

Existing approaches attempt to address this through Class Incremental Object Detection (CIOD) \cite{shmelkov2017incremental, CL-DETR, OW-DETR, VLM-PL, SDDGR}, which learns new object categories while assuming a fixed domain, and Domain Incremental Object Detection (DIOD) \cite{LDB}, which adapts to unseen domains while keeping object categories unchanged. However, these methods operate under restrictive assumptions—CIOD neglects domain shifts, while DIOD assumes a fixed set of object categories. As shown in Figure \ref{fig:teaser}, both approaches fail when faced with simultaneous class and domain shifts, a scenario frequently encountered in real-world applications. 

\begin{figure}[ht]
\begin{center}
\centerline{\includegraphics[width=\columnwidth]{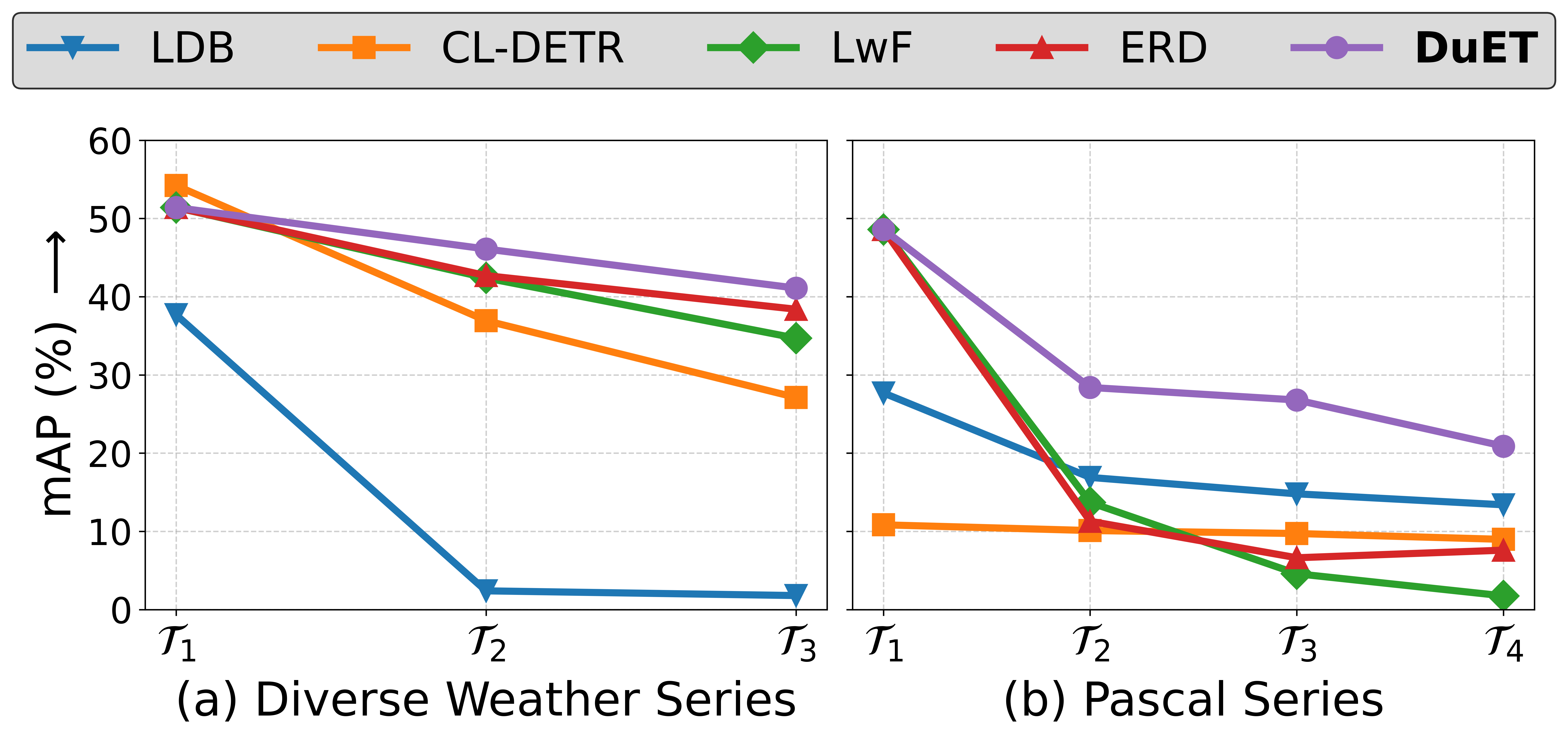}}
\vspace{-1em}
\caption{\textbf{Catastrophic forgetting of \(\mathcal{T}_1\) classes in multi-phase experiments for various methods on the DuIOD task.} The \emph{\textbf{left plot}} illustrates catastrophic forgetting in \textit{Night Sunny [1:2]} classes across three incremental tasks: Night Sunny [1:2] (\(\mathcal{T}_1\)) \(\rightarrow\) Daytime Sunny [3:4] (\(\mathcal{T}_2\)) \(\rightarrow\) Daytime Foggy [5:7] (\(\mathcal{T}_3\)) on the Diverse Weather Series dataset. Similarly, the \emph{\textbf{right plot}} depicts catastrophic forgetting in \textit{Watercolor [1:3]} classes across four incremental tasks: Watercolor [1:3] (\(\mathcal{T}_1\)) \(\rightarrow\) Comic [4:6] (\(\mathcal{T}_2\)) \(\rightarrow\) Clipart [7:13] (\(\mathcal{T}_3\)) \(\rightarrow\) VOC [14:20] (\(\mathcal{T}_4\)) on the \emph{Pascal Series} dataset.}
\vspace{-10mm}
\label{fig:cat-forget}
\end{center}
\end{figure}

To bridge this gap, we introduce a more practical yet underexplored setting, Dual Incremental Object Detection (DuIOD), where object detectors must incrementally learn new object categories across evolving domains without access to previous training data (exemplar-free). While some works, such as VIL \cite{VIL}, extend incremental learning to both class and domain shifts, they primarily focus on classification tasks and do not address key object detection challenges like spatial feature retention and bounding box refinement, nor do they tackle the background shift problem of IOD \cite{ABR_IOD, continual_survey}, where previously learned objects are misclassified as background if unannotated in later tasks, all of which are present in DuIOD. Recently, PD \cite{Purified_Distillation} partially addresses domain shifts in object detection but fails to generalize to the broader DuIOD problem. 

To tackle the challenges of DuIOD, we propose DuET, a simple yet effective Task Arithmetic (TA) \cite{TA_Paper_0} based incremental learning framework for robust object detection across evolving domains. TA paradigm was recently introduced by \cite{TA_Paper_0} for modifying pre-trained models via arithmetic operations on task vectors. Formally, a task vector is derived by subtracting the weights of a pre-trained model from those of the same model fine-tuned on a specific task. In DuET, we decompose model parameters into shared parameters, which persist across all incremental tasks, and task-specific parameters, which evolve incrementally. We compute task vectors for both previous and current tasks using the shared parameters. DuET framework is built upon two components: \textit{DuET Module} \& \textit{Incremental Head}. \textit{DuET Module} merges these task vectors through dynamic layer-wise retention and adaptation factors, effectively mitigating catastrophic forgetting. Meanwhile, the \textit{Incremental Head} concatenates task-specific parameters from both past and current tasks, enhancing generalization. Unlike prior methods, DuET is object detector-agnostic, seamlessly integrating with diverse detection backbones. Notably, it enables state-of-the-art real-time object detectors, such as YOLO11 \cite{YOLO11} and RT-DETR \cite{RT-DETR}, to function as real-time incremental object detectors.

Additionally, we introduced Directional Consistency Loss to further stabilize incremental updates, which mitigates weight update conflicts during model merging \cite{TIES_MERGING, DARE}. Moreover, to jointly quantify catastrophic forgetting and adaptation performance, we propose a new evaluation metric, the Retention-Adaptability Index (RAI). 

Our extensive experiments on \emph{Pascal Series} and \emph{Diverse Weather Series} datasets demonstrate the effectiveness of DuET in addressing the DuIOD task. As highlighted in Figure \ref{fig:cat-forget}, CIOD-based methods (e.g., CL-DETR \cite{CL-DETR}) suffer severe forgetting in unseen domains, while DIOD-based methods (e.g., LDB \cite{LDB}) fail to retain previously learned categories. Additionally, conventional incremental learning techniques such as LwF \cite{LwF} and ERD \cite{ERD} also struggle in this setting. In contrast, DuET significantly mitigates catastrophic forgetting and enables robust incremental learning across both class and domain shifts. We highlight our contributions as:

\noindent - To the best of our knowledge, DuET is the first method to tackle DuIOD using the concept of Task Merging. \\
\noindent - We propose a novel TA-based algorithm which is object detector agnostic, validated YOLO11 and RT-DETR, enabling real-time incremental object detection. \\
\noindent - We propose Directional Consistency Loss to mitigate sign conflicts during model merging, ensuring stable incremental learning. \\
\noindent - To quantify both catastrophic forgetting and adaptation performance, we introduce a new evaluation metric, the Retention-Adaptability Index.
\section{Related Works}
\noindent \textbf{Class Incremental Object Detection:}  
CIOD has evolved from early CNN-based methods \cite{pang2019libra, Faster_ILOD, ABR_IOD, cermelli2022modeling, shmelkov2017incremental, SID, RILOD, ERD} to Transformer-based approaches \cite{CL-DETR, OW-DETR, VLM-PL, kang2023alleviating, SDDGR}, improving generalization and stability during incremental updates. Initial efforts, such as \cite{shmelkov2017incremental}, introduced CIOD via knowledge distillation inspired by incremental classification methods like LwF \cite{LwF}. Subsequent works explored intermediate feature distillation \cite{hao2019end, chen2020incremental, chen2019new}, replay buffers \cite{shieh2020continual, Rodeo, SDDGR}, parameter isolation \cite{li2018incremental, zhang2021incremental}, and pseudo-labeling \cite{OW-DETR, RD-IOD, VLM-PL}.  

Transformer-based models, including CL-DETR \cite{CL-DETR}, SDDGR \cite{SDDGR}, and VLM-PL \cite{VLM-PL}, use \textit{Deformable DETR} \cite{DeformableDETR} to enhance CIOD. CL-DETR applies knowledge distillation on labels with exemplar replay, SDDGR integrates generative replay via stable diffusion, and VLM-PL employs vision-language pseudo-labeling. However, most CIOD methods rely on exemplars \cite{zhong2025replay, SDDGR, Rodeo} and assume a static domain. In contrast, DuET proposes an exemplar-free, unified approach that prevents \textit{catastrophic forgetting} while incrementally learning new objects across domains.  

\noindent \textbf{Domain Incremental Object Detection:}  
DIOD focuses on adapting to distribution shifts—such as changes in lighting, viewpoints, and weather—while maintaining a fixed object set. LDB \cite{LDB} tackles DIOD by learning domain biases with frozen base models and employing a nearest-mean classifier for domain selection, while other methods \cite{S-Prompts, DISC, PINA} primarily target \textit{Domain Incremental Learning} through prompt-based strategies, domain statistics, and cross-domain concept alignment.  

Although some DIOD methods support \textit{exemplar-free adaptation}, they assume a consistent class set across domains and struggle with \textit{inter-domain class confusion}. DuET overcomes these limitations by using \textbf{task arithmetic-based model merging} to incrementally learn class and domain shifts without relying on exemplars.  

\noindent \textbf{Task Arithmetic:}  
Task arithmetic has recently emerged as an effective way to modify pre-trained models through simple arithmetic operations on model weights. Task-Arithmetic \cite{TA_Paper_0} showed that \textit{task vectors}—differences between fine-tuned and pre-trained model weights—can steer model behaviour. Extensions like Fisher-Merging \cite{Fisher-merging} and RegMean \cite{RegMean} introduced importance-weighted schemes using Fisher information and inner-product matrices, but they require manual matrix computations. To resolve \textit{sign conflicts} during model merging, TIES-Merging \cite{TIES_MERGING} enforces orthogonality constraints, while DARE \cite{DARE} reduces interference using dropout and rescaling. EMR-Merging \cite{EMR-Merging} aligns task directions and amplitudes through a tuning-free strategy, and MagMax \cite{MAGMAX} mitigates forgetting through important parameters selection.  

Unlike standard TA or the variants mentioned above, which merge entire task vectors without distinction, DuET applies layer-wise Task merging with dynamically calculated retention-adaptation factors to prevent catastrophic forgetting while improving generalization. In Section \ref{ablations}, we benchmark DuET against existing model-merging baselines and demonstrate its superiority across multiple object detectors in the DuIOD task.  

\section{Methodology}
\label{Method}
\subsection{Problem formulation}
\label{problem_formulation}
In exemplar-free DuIOD, our goal is to develop a unified and robust object detection model that incrementally learns new object classes and adapts to evolving domain characteristics without retaining any past training data  (exemplars).  

Formally, we define a sequence of tasks as \( T \) = \( \{ \mathcal{T}_1, \mathcal{T}_2, \dots, \mathcal{T}_T \} \), where each task \( \mathcal{T}_t \) introduces a new set of object classes \( \mathcal{C}_t \) from a novel domain \( \mathcal{D}_t \). The cumulative set of classes and domains encountered up to task \( \mathcal{T}_t \) is denoted by: \( \mathcal{C}_{1:t} = \bigcup_{j=1}^{t} \mathcal{C}_j \) and \( \mathcal{D}_{1:t} = \bigcup_{j=1}^{t} \mathcal{D}_j \), respectively. 
Furthermore, following the typical CIOD \& DIOD assumptions \cite{shmelkov2017incremental, CL-DETR, OW-DETR, VLM-PL, SDDGR, LDB}, we ensure that:
\begin{equation}
    \mathcal{C}_t \cap \mathcal{C}_{1:t-1} = \emptyset, \quad \mathcal{D}_t \cap \mathcal{D}_{1:t-1} = \emptyset, \quad \forall \ t \geq 2
\end{equation}
\begin{equation}
    P(\mathcal{D}_t) \neq P(\mathcal{D}_{t'}) \ \forall \ t \neq t'.
\end{equation}
For each task \( \mathcal{T}_t \), we define the dataset as \( \{ \mathcal{X}_t, \mathcal{Y}_t \} \), where \( \mathcal{X}_t = \{ x_j^t \}_{j=1}^{\mathcal{N}_t} \) consists of input images from \( \mathcal{D}_t \) that contain only objects from \( \mathcal{C}_t \), and \( \mathcal{Y}_t = \{ y_j^t \}_{j=1}^{\mathcal{N}_t} \) provides bounding box annotations exclusively for \( \mathcal{C}_t \). While images in \( \mathcal{X}_t \) may include objects from previously learned classes \( \mathcal{C}_{1:t-1} \), these objects remain unlabeled and are treated as background. This results in a fundamental \textit{background shift} problem \cite{ABR_IOD, continual_survey}, where previously learned objects become part of the background in later tasks—an issue absent in prior works like \cite{VIL}.  

To maintain an exemplar-free setting, during task \( \mathcal{T}_t \), we have access only to the dataset \( \mathcal{D}_t \) and do not retain any images or annotations from previous domains \( \{ \mathcal{D}_1, \mathcal{D}_2, \dots, \mathcal{D}_{t-1} \} \). Our objective is to incrementally update the object detection model \( \mathcal{M}_{\theta_t} \), parameterized by \( \theta_t \), using only data from \( \mathcal{D}_t \), ensuring that \( \mathcal{M}_{\theta_t} \) can accurately detect objects from all learned classes \( \mathcal{C}_{1:t} \) across all encountered domains \( \mathcal{D}_{1:t} \).  

\subsection{Overview of DuET Approach}
\label{duet_approach}
\begin{figure*}[ht]
\begin{center}
\centerline{\includegraphics[width=\textwidth]{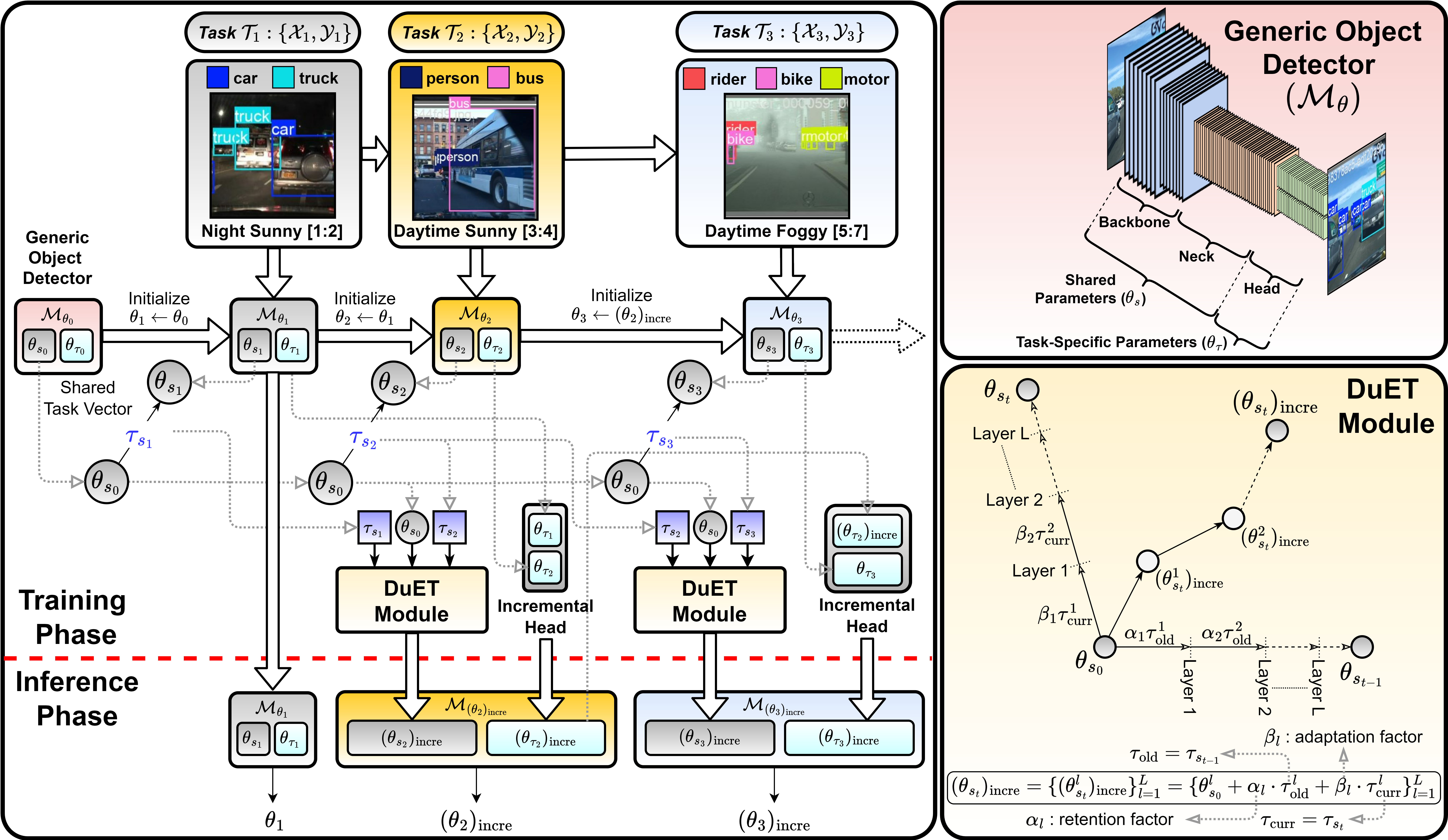}}
\vspace{-0.5em}
\caption{
\textbf{Overview of the proposed DuET framework for exemplar-free Dual Incremental Object Detection (DuIOD).}
\emph{\textbf{Left:}} To setup DuIOD task, we illustrate a sequence of three tasks \( \{\mathcal{T}_1,\,\mathcal{T}_2,\,\mathcal{T}_3\} \), each introducing new object classes (e.g.\ \( \{\textit{car, truck}\}, \{\textit{person, bus}\}, \{\textit{rider, bike, motor}\} \)) under different domain shifts (Night Sunny, Daytime Sunny, Daytime Foggy) respectively. A generic object detector \( \mathcal{M}_{\theta_0} \) is first fine-tuned on \( \mathcal{T}_1 \), yielding parameters \( \theta_{1} \), the parameters of the model at every Task are decomposed into shared (\( \theta_{s} \)) and task-specific (\( \theta_{\tau_1} \)) parameters. Subsequently, for each incremental Task (\( \mathcal{T}_t, t \geq 2\)), the shared parameters are merged via \textit{DuET Module} while the \textit{Incremental Head} concatenates the task-specific parameters from the old and current Task, \( \theta_{\tau_t} \) \& \( (\theta_{\tau_{t-1}})_{\text{incre}} \) respectively. During the \textit{Inference Phase}, the merged incremental weights \( \theta_1, (\theta_2)_{\text{incre}}\ \& \ (\theta_3)_{\text{incre}} \) are used to detect objects from the respective cumulative sets of classes \( \mathcal{C}_1,  \mathcal{C}_{1:2} \ \& \ \mathcal{C}_{1:3} \).
\emph{\textbf{Top-Right:}} We depict the decomposition of \( \theta_{s} \) \& \( \theta_{\tau} \) for a generic, backbone-agnostic object detector \( \mathcal{M}_{\theta} \), with the DuET framework validated across a variety of object detectors (e.g. YOLO11 \cite{YOLO11}, RT-DETR \cite{RT-DETR}). 
\emph{\textbf{Bottom-Right:}} We zoom in on the DuET Module, which merges the old and current \emph{shared task vectors} (\( \tau_{old}, \tau_{curr} \)) using \textit{layer-wise retention (\(\alpha_l\)) and adaptation (\(\beta_l\)) factors} as per equation \ref{merging_eqn} (zoomed-in for best view).
}
\vspace{-12mm}
\label{fig:main_arch}
\end{center}
\end{figure*}
To tackle DuIOD, we discuss the DuET approach in this and subsequent sections. As depicted in Figure~\ref{fig:main_arch}, DuET strategically decomposes model parameters into shared ($\theta_s$) and task-specific ($\theta_{\tau_t}$) components, thereby facilitating efficient knowledge transfer and modular adaptation. The DuET framework is built upon two core components:
\begin{itemize}
    \item \textbf{DuET Module:} This module fuses shared task vectors ($\tau_{\text{old}}$ and $\tau_{\text{curr}}$) through dynamic, layer-wise weighting factors—retention ($\alpha_l$) and adaptation ($\beta_l$)—as described in Section~\ref{duet_module}. This design not only preserves previously acquired knowledge but also incorporates new information in a balanced manner, effectively countering catastrophic forgetting.
    \item \textbf{Incremental Head:} This component consolidates task-specific parameters from both past and current tasks, thereby enhancing the model’s generalization across diverse detection scenarios. Additionally, \textbf{Directional Consistency Loss} ($\mathcal{L}_{\text{DC}}$, see Section~\ref{dc_loss}) enforces alignment in the direction of incremental updates and prevents detrimental sign conflicts.
\end{itemize}
\noindent Unlike prior approaches \cite{Purified_Distillation, LwF, ERD}, which are optimized for specific detection backbones, DuET exhibits broad generalization across a variety of object detectors \cite{DeformableDETR, YOLO11, RT-DETR} (see Table~\ref{all_detectors}). Moreover, DuET's training and TA algorithm are provided in \texttt{Algorithms 1 \& 2} (\texttt{suppl.}).
\subsection{Training Strategy}
\label{training_phase}
\paragraph{Base Task: \(\mathcal{T}_1\).}
Consider a generic, pre-trained object detector \(\mathcal{M}_{\theta_0}\) parameterized by \(\theta_0\) as shown in figure \ref{fig:main_arch}. Let Task~\(\mathcal{T}_1\) introduce new \(\mathcal{C}_1\) classes from \(\mathcal{D}_1\) domain. We first train this model on the dataset \(\{\mathcal{X}_1, \mathcal{Y}_1\}\) for Task ~\(\mathcal{T}_1\) using the standard detection loss, \(\mathcal{L}_{\text{Detector}}\), which is specific to the object detector being considered. After the first update step, the model’s parameters become:
\begin{equation}
    \theta_1 \;\; \leftarrow \;\; \theta_0 \;-\; \eta \,\nabla_{\theta}\,\mathcal{L}_{\text{Detector}}\bigl(\theta_0; \mathcal{X}_1, \mathcal{Y}_1\bigr),
\end{equation}
where \(\eta\) is the learning rate.  

At every incremental step, we decompose the model parameters into \textit{shared parameters} \(\theta_s\), which are common across all incremental tasks, and \textit{task-specific parameters} \(\theta_{\tau_t}\), which evolve with each incremental step. For instance, in YOLO11 \cite{YOLO11}, the backbone and neck components, responsible for multi-scale feature extraction and feature refinement, constitute \(\theta_s\), whereas the detection head, which is responsible for final task-specific object localization and classification, forms \(\theta_{\tau_t}\). Hence, for Task~\(\mathcal{T}_1\), we have:
\begin{equation}
    \theta_0 \;\rightarrow\; \bigl[\theta_{s_0}, \,\theta_{\tau_0}\bigr], 
    \quad
    \theta_1 \;\rightarrow\; \bigl[\theta_{s_1}, \,\theta_{\tau_1}\bigr].
\end{equation}
After Task~\(\mathcal{T}_1\) training, following the task vector definition from \cite{TA_Paper_0}, we compute the first \textbf{task vector} as:
\begin{equation}
    \tau_{s_1} \;=\; \theta_{s_1} \;-\; \theta_{s_0}, 
\end{equation}
Here \(\tau_{s_1}\) effectively captures how the shared parameters must shift to accommodate the first Task, while the remaining parameters, \(\theta_{\tau_1}\) constitute the task-specific weights that help in localization and classification of the newly introduced \(\mathcal{C}_1\) classes. As illustrated in Figure~\ref{fig:main_arch}, for inference on the base task \(\mathcal{T}_1\), we directly employ \(\mathcal{M}_{\theta_1}\). 
\vspace{-5.4mm}
\paragraph{Incremental Tasks \(\{\mathcal{T}_2, \mathcal{T}_3\dots, \mathcal{T}_T\}\).}
For each incremental task (\( \mathcal{T}_t, t \geq 2\)), the model encounters a new set of classes \(\mathcal{C}_t\) from domain \(\mathcal{D}_t\). We follow sequential fine-tuning and initialize \(\theta_t \gets \theta_{t-1}\), following \cite{MAGMAX}, but this time, we train the model using \(\mathcal{L}_{\text{Total}}\) (eq.~\ref{dc_loss}), which incorporates a modified Distillation Loss \(\mathcal{L}_{\text{Distill}}^{*}\) to help preserve knowledge from previously learned tasks and Directional Consistency Loss \(\mathcal{L}_{\text{DC}}\) to help resolve sign conflicts and encourage stable parameter updates, along with the object-detector loss \(\mathcal{L}_{\text{Detector}}\) (see section \ref{dc_loss}). Hence, after every incremental Task \(\mathcal{T}_t\), the model parameters are updated as:
\begin{equation}
\label{eq:theta_update}
    \theta_t \;\; \leftarrow \;\; \theta_{t-1} \;-\; \eta \,\nabla_{\theta}\,\mathcal{L}_{\text{Total}}\bigl(\theta_{t-1}; \mathcal{X}_t, \mathcal{Y}_t\bigr).
\end{equation}
As before, we then decompose the updated parameters into:
\begin{equation}
    \theta_{t-1} \;\rightarrow\; \bigl[\theta_{s_{t-1}}, \,\theta_{\tau_{t-1}}\bigr], 
    \quad
    \theta_{t} \;\rightarrow\; \bigl[\theta_{s_{t}}, \,\theta_{\tau_{t}}\bigr].
\end{equation}
Now, to incorporate the knowledge from current domain \(\mathcal{D}_t\) and current classes \(\mathcal{C}_t\) into the shared parameters \(\theta_s\) without discarding knowledge of older tasks \(\mathcal{T}_{1:t-1}\), we define two \textbf{shared task-vectors} as:
\begin{equation}
\label{eq:old_new_tau}
    \tau_{\text{old}} \;=\; \theta_{s_{t-1}} \;-\; \theta_{s_{0}}\footnotemark[1],
    \quad
    \tau_{\text{curr}} \;=\; \theta_{s_{t}} \;-\; \theta_{s_{0}}\footnotemark[1].
\end{equation}
\footnotetext[1]{Task vectors are computed assuming access to the pre-trained model for all tasks, a standard assumption in task vector-based methods \cite{TA_Paper_0, EMR-Merging}.}
The shared task vectors, \(\tau_{\text{old}}\) and \(\tau_{\text{curr}}\), play a crucial role in capturing the evolution of the shared parameters as the model transitions between tasks. Specifically, \(\tau_{\text{old}}\) represents the cumulative update from the initial pre-trained state up to the last incremental phase, effectively embedding the knowledge of all previous tasks. In contrast, \(\tau_{\text{curr}}\) captures the immediate update introduced by the current Task. This separation allows us to precisely balance the retention of previously learned features with the integration of new information, thereby mitigating catastrophic forgetting while enhancing the model’s generalization capability. By isolating these vectors, the proposed DuET framework benefits from clearly delineating what knowledge needs to be preserved versus what can be adapted.

\subsection{Fusing task vectors on shared weights}
\label{duet_module}
The core of DuET approach lies in the \textbf{DuET Module}, which performs the dynamic integration of \(\tau_{\text{old}}\) and \(\tau_{\text{curr}}\) to update the shared weights (Figure~\ref{fig:main_arch}). Although these retention-adaptation factors are empirically motivated, they can be viewed through a \emph{stability-plasticity} lens in incremental learning. Interpreting \(\tau_{\text{old}}^l\) and \(\tau_{\text{curr}}^l\) as “directions” in parameter space—one encoding the cumulative shifts for past tasks and the other the shift introduced by the new Task—our layer-wise \(\alpha_l\) and \(\beta_l\) define a linear combination that balances preserving old knowledge (\(\alpha_l \approx 1\)) versus integrating new information (\(\beta_l \approx 1\)). By comparing \(\|\tau_{\text{old}}^l\|\) and \(\|\tau_{\text{curr}}^l\|\) at each layer, we obtain a first-order approximation of “which update dominates,” adaptively favoring retention when \(\|\tau_{\text{old}}^l\|\) is large and plasticity when \(\|\tau_{\text{curr}}^l\|\) is large. Geometrically, this ensures the merged parameters lie neither too close to the old solution nor fully dominated by new updates, effectively mitigating catastrophic forgetting and enhancing generalization. For each layer \(l \in \{1, \dots, \mathcal{L}\}\), we define a p-factor:
\begin{equation}
    p_l = \frac{\|\tau_{\text{old}}^l\| - \|\tau_{\text{curr}}^l\|}{\|\tau_{\text{old}}^l + \tau_{\text{curr}}^l\| + \epsilon},
\end{equation}
where \(\|\cdot\|\) denotes the \(\ell_1\) norm and \(\epsilon\) ensures numerical stability. Mapping \(p_l\) via a hyperbolic tangent scaled by \(\gamma\) gives:
\begin{equation}
    \delta_l = \gamma\,\tanh(p_l),
\end{equation}
After clamping \(\delta_l\) to \([-\gamma, \gamma]\), and adding a base coefficient \(\alpha_{\text{base}}\), we define:
\begin{equation}
    \alpha_l = \alpha_{\text{base}} + \operatorname{clamp}(\delta_l, -\gamma, \gamma), 
    \quad 
    \beta_l = 1 - \alpha_l.
\end{equation}
Hence, unlike second-order methods (e.g., Fisher Merging \cite{Fisher-merging}) that require computing parameter importance matrices, the p-factor captures a simpler, local notion of importance. It forgoes expensive matrix approximations yet still encodes how far each layer’s weights have drifted from the original baseline. Given these per-layer coefficients, the updated shared weights for layer \(l\) are:
\begin{equation}
\label{merging_eqn}
    (\theta_{s_t}^l)_{\text{incre}} = \theta_{s_0}^l + \alpha_l\,\tau_{\text{old}}^l + \beta_l\,\tau_{\text{curr}}^l.
\end{equation}
Applying this across all layers yields \((\theta_{s_t})_{\text{incre}}\). 
Simultaneously, the \textbf{Incremental Head} concatenates the current task-specific parameters with the previous ones:
\begin{equation}
    (\theta_{\tau_t})_{\text{incre}} \gets \bigl[\theta_{\tau_t} \,;\, (\theta_{\tau_{t-1}})_{\text{incre}}\bigr]
\end{equation}
Finally, combining these updated shared and task-specific components produces the overall incremental weights:
\begin{equation}
    (\theta_t)_{\text{incre}} \gets \bigl[(\theta_{s_t})_{\text{incre}},\, (\theta_{\tau_t})_{\text{incre}}\bigr]
\end{equation}
which initialize the model for the subsequent Task, i.e., \(\theta_{t+1} \gets (\theta_t)_{\text{incre}}\).

As illustrated in Figure~\ref{fig:main_arch}, for inference on incremental tasks (\(t \geq 2\)), we employ the merged incremental models \(\mathcal{M}_{\theta_t}\), parameterized by the reconstructed weights \((\theta_t)_{\text{incre}}\), to detect \(\mathcal{C}_{1:t}\) classes across domains \(\mathcal{D}_{1:t}\) following the evaluation protocol detailed in Section~\ref{evaluation_protocol}.

\subsection{Loss objectives}
\label{loss_objectives}
For the base task (\(t=1\)), we optimize the object detector using the standard detection loss, \(\mathcal{L}_{\text{Detector}}\), which is specific to the base detector and is responsible for object localization and classification. For incremental tasks (\(t \geq 2\)), we augment \(\mathcal{L}_{\text{Detector}}\) with a modified Distillation Loss \(\mathcal{L}_{\text{Distill}}^{*}\) (discussed in \texttt{Appendix B}), along with our Directional Consistency Loss \(\mathcal{L}_{\text{DC}}\) (discussed below) scaled by scaling coefficients \( \lambda_{\text{Distill}} \) and \( \lambda_{\text{DC}} \) respectively.

Formally, the total loss for task \(t\) is:
\begin{equation}
    \scalebox{0.88}{$
    \mathcal{L}_{\text{Total}} = 
    \begin{cases} 
    \mathcal{L}_{\text{Detector}}, & \text{for } t = 1, \\[6pt]
    \mathcal{L}_{\text{Detector}} 
    + \lambda_{\text{Distill}}\,\mathcal{L}_{\text{Distill}}^{*}
    + \lambda_{\text{DC}}\,\mathcal{L}_{\text{DC}}, & \text{for } t \geq 2.
    \end{cases} $}
\end{equation}
\paragraph{Directional Consistency Loss \((\mathcal{L}_{\text{DC}}\)).}
\label{dc_loss}
Incremental learning suffers from conflicting weight updates during model-merging \cite{TIES_MERGING}, which can destabilize the merging of shared parameters. We introduce the \textit{Directional Consistency Loss} (DC Loss) to address this. This loss penalizes updates in the shared parameter space that diverge in direction relative to previous incremental changes, thus promoting a consistent evolution of the shared task vectors.

Specifically, for \(t \geq 2\), we define the DC Loss between consecutive tasks \(t\) and \(t-1\) as:
\begin{equation}
    \scalebox{0.9}{$
    \mathcal{L}_{\text{DC}} = \sum\limits_{i \in \theta_s} \text{ReLU}\Bigl[-\Bigl( \bigl(\tau_{s_t}^{(i)} - \tau_{s_{t-1}}^{(i)}\bigr) \cdot \bigl(\tau_{s_{t-1}}^{(i)} - \tau_{s_{t-2}}^{(i)}\bigr) \Bigr)\Bigr] 
    $}
\end{equation}
Here, the dot product quantifies the alignment between successive updates: a negative value indicates a sign conflict (i.e., the current update opposes the previous direction), which the ReLU penalizes. This mechanism enforces a more coherent update trajectory for the shared parameters, effectively mitigating catastrophic forgetting. Consequently, the DC Loss remains object-detector agnostic, as it depends solely on the evolution of shared parameter updates  

\section{Experimental Details}
\label{experiments}
\subsection{Datasets}
\label{datasets}
Following prior works \cite{LDB, kiran2022incremental, liu2020multi, DivAlign, Single-DGOD}, we evaluate DuIOD using two dataset series: \emph{Pascal Series} and \emph{Diverse Weather Series}. \emph{Pascal Series} includes four domains—Pascal VOC \cite{PASCAL_VOC}, Clipart, Watercolor, and Comic \cite{PASCAL_Series}. Pascal VOC and Clipart share 20 object categories, while Watercolor and Comic contain six, forming subsets of the former. Similarly, in \emph{Diverse Weather Series}, we consider three diverse weather conditions—Daytime Sunny, Night Sunny, and Daytime Foggy—with seven common object categories, sourced from BDD-100k \cite{BDD100k}, FoggyCityscapes \cite{FoggyCityscapes}, and Adverse-Weather \cite{AdverseWeather}. The training sequence followed for both dataset series is provided in \texttt{Appendix A- Table S1}, and dataset statistics are detailed in \texttt{Appendix G}.

\subsection{Proposed Evaluation Protocol and Metrics}
\label{evaluation_protocol}
To comprehensively evaluate different methods on the DuIOD task, we propose a new evaluation protocol (see \texttt{Appendix A- Table S1}). Existing metrics such as \( \mathcal{F}_{map} \) \cite{chen2019new}, RSD \& RPD \cite{continual_survey}, and SPmAP \cite{yang2022multi}, primarily assess catastrophic forgetting and do not capture the model’s ability to generalize to unseen classes, which is critical for DuIOD. Since DuIOD goes beyond preserving old knowledge (retention), it also demands evaluating how well the model can adapt to unseen categories (adaptability). We introduce the \textbf{Retention-Adaptability Index (RAI)} to address this requirement. RAI is defined as the mean of the Average Retention Index (Avg RI) and Average Generalization Index (Avg GI). 

\textbf{Avg RI} quantifies the model’s ability to retain past knowledge and is computed as:
\begin{equation}
    \scalebox{0.95}{$
    \text{Avg RI} = \frac{1}{T - 1} \sum_{i=1}^{T - 1} RI_{\mathcal{D}_i},
    \
    RI_{\mathcal{D}_i} = \frac{\text{mAP}_{\text{old}}^{\mathcal{T}_T}(\mathcal{D}_i[\mathcal{C}_i])}{\text{mAP}_{\text{new}}^{\mathcal{T}_i}(\mathcal{D}_i[\mathcal{C}_i])}. $}
\end{equation}
Here, \( \text{mAP}_{\text{old}}^{\mathcal{T}_T} \) represents the final-task performance on previously learned classes, while \( \text{mAP}_{\text{new}}^{\mathcal{T}_i} \) denotes the initial performance when the classes were first encountered. A higher Avg RI implies better retention of past knowledge while lower values indicate catastrophic forgetting.

Similarly, \textbf{Avg GI} quantifies the model’s generalization to unseen classes and is computed as: 
\begin{equation}
    \scalebox{0.8}{$
    \text{Avg GI} = \frac{1}{N} \sum_{(\mathcal{D}_i, \mathcal{T}_j)} GI_{\mathcal{D}_i, \mathcal{T}_j},
    \
    GI_{\mathcal{D}_i, \mathcal{T}_j} = \frac{\text{mAP}_{\text{unseen}}^{\mathcal{T}_j}(\mathcal{D}_i[\mathcal{C}_{\text{unseen}}])}{\text{mAP}_{\text{ref}}(\mathcal{D}_i[\mathcal{C}_{\text{unseen}}])}.$}
\end{equation}
Here, \( \text{mAP}_{\text{unseen}}^{\mathcal{T}_j} \) measures model performance on previously unseen classes, while \( \text{mAP}_{\text{ref}} \) is the reference mAP obtained by training solely on these unseen classes. A higher Avg GI indicates better adaptability to new categories. A detailed discussion, including metric formulations and rationale, is provided in \texttt{Appendix A}.  

Moreover, unlike prior works, we do not consider the upper-bound joint-training baseline, as Avg RI and Avg GI compute relative performance retention and generalization through direct ratio-based comparisons. This ensures that the overall metric (RAI) directly reflects catastrophic forgetting without biases introduced by architecture-specific joint-training performance.  
   
\begin{table*}[!ht]
    \centering
    \caption{Detailed results of various methods on Daytime Sunny [1:4] \(\rightarrow\) Night Sunny [5:7], where the expanded columns (\(\mathcal{T}_1\): Daytime Sunny [1:4], \(\mathcal{T}_2\): Night Sunny [5:7]) show \(\text{mAP@}0.5\%\), best among the columns in \textbf{bold}, second best \textit{\underline{underlined}}.}
    \label{E4_expanded}
    \vspace{-0.5em}
    \resizebox{\textwidth}{!}{
    \begin{tabular}{ccc||c|c|c|c|c||ccc}
    \toprule
        \multirow{4}{*}{\textbf{Method}} & \multirow{4}{*}{\textbf{Base Detector}} & \multirow{4}{*}{\makecell{\textbf{Trainable} \\ \textbf{Params (M)}}} &
        \multirow{4}{*}{\makecell{\textbf{T1} \\ \textbf{Daytime} \\ \textbf{Sunny} \\ \textbf{{[1:4]}}}} &
        \multicolumn{4}{c||}{\textbf{T2: Night Sunny [5:7]}} &
        \multirow{4}{*}{\makecell{\textbf{Avg RI} \\ \textbf{(\%)}}} & 
        \multirow{4}{*}{\makecell{\textbf{Avg GI} \\ \textbf{(\%)}}} & 
        \multirow{4}{*}{\makecell{\textbf{RAI} \\ \textbf{(\%)}}} \\
        \cline{5-8}
        & & & & \makecell{\textbf{Old}} &  
        \makecell{\textbf{New}} &
        \multicolumn{2}{c||}{\textbf{Unseen}} \\
        \cline{5-8}
        & & & & \multirow{2}{*}{\makecell{\textbf{Daytime} \\ \textbf{Sunny [1:4]}}} & 
        \multirow{2}{*}{\makecell{\textbf{Night} \\ \textbf{Sunny [5:7]}}} & 
        \multirow{2}{*}{\makecell{\textbf{Night} \\ \textbf{Sunny [1:4]}}} & 
        \multirow{2}{*}{\makecell{\textbf{Daytime} \\ \textbf{Sunny [5:7]}}} & & & \\
        & & & & & & & & & & \\
        \hline
        Sequential FT & YOLO11n & 2.58 & 49.40\scriptsize{$\pm$0.3} & 0.00\scriptsize{$\pm$0.0} & 62.20\scriptsize{$\pm$0.5} & 12.60\scriptsize{$\pm$0.4} & 35.90\scriptsize{$\pm$0.3} & 0.00\scriptsize{$\pm$0.0} & 45.88\scriptsize{$\pm$0.6} & 22.94\scriptsize{$\pm$0.3} \\
        \(\mathrm{LwF}_{\textcolor{MidnightBlue}{\text{ECCV'}16}}\) \cite{LwF} & YOLO11n & 2.58 & 49.40\scriptsize{$\pm$0.2} & 27.60\scriptsize{$\pm$0.4} & 0.34\scriptsize{$\pm$0.6} & 21.30\scriptsize{$\pm$0.3} & 0.67\scriptsize{$\pm$0.5} & 55.87\scriptsize{$\pm$0.3} & 21.88\scriptsize{$\pm$0.7} & 38.88\scriptsize{$\pm$0.6} \\
        \(\mathrm{ERD}_{\textcolor{MidnightBlue}{\text{CVPR'}22}}\) \cite{ERD} & YOLO11n & 2.58 & 49.40\scriptsize{$\pm$0.5} & 33.00\scriptsize{$\pm$0.4} & 34.00\scriptsize{$\pm$0.3} & 26.10\scriptsize{$\pm$0.6} & 29.10\scriptsize{$\pm$0.7} & \textit{\underline{66.80}}\scriptsize{$\pm$0.5} & 53.04\scriptsize{$\pm$0.3} & \textit{\underline{59.92}}\scriptsize{$\pm$0.4} \\
        \(\mathrm{LDB}_{\textcolor{MidnightBlue}{\text{AAAI'}24}}\footnotemark[2]\)  \cite{LDB} & VitDet & 110.52 & 45.30\scriptsize{$\pm$0.6} & 0.50\scriptsize{$\pm$0.3} & 15.10\scriptsize{$\pm$0.4} & 0.30\scriptsize{$\pm$0.5} & 16.90\scriptsize{$\pm$0.7} & 1.10\scriptsize{$\pm$0.2} & 22.41\scriptsize{$\pm$0.3} & 11.76\scriptsize{$\pm$0.6} \\
        \(\mathrm{CL-DETR}_{\textcolor{MidnightBlue}{\text{CVPR'}23}}\footnotemark[2]\) \cite{CL-DETR} & Deformable DETR & 39.85 & 46.29\scriptsize{$\pm$0.4} & 27.41\scriptsize{$\pm$0.5} & 31.94\scriptsize{$\pm$0.6} & 19.85\scriptsize{$\pm$0.3} & 32.55\scriptsize{$\pm$0.4} & 59.21\scriptsize{$\pm$0.2} & \textit{\underline{54.96}}\scriptsize{$\pm$0.5} & 57.09\scriptsize{$\pm$0.4} \\
        \rowcolor{lightblue} \textbf{DuET (Ours)} & YOLO11n & 2.58 & 49.40\scriptsize{$\pm$0.2} & 43.50\scriptsize{$\pm$0.1} & 22.20\scriptsize{$\pm$0.3} & 31.60\scriptsize{$\pm$0.2} & 27.40\scriptsize{$\pm$0.1} & \textbf{88.06}\scriptsize{$\pm$0.2} & \textbf{56.95}\scriptsize{$\pm$0.1} & \textbf{72.51}\scriptsize{$\pm$0.2} \\
    \bottomrule
    \end{tabular}
    }
\end{table*}
\begin{table*}[!ht]
    \centering
    \caption{Results of two-phase and multi-phase experiments on DuIOD task. Best among the columns in \textbf{bold}, second best \textit{\underline{underlined}}.}
    \label{E1_E3_E6_combined}
    \vspace{-0.5em}
    \renewcommand{\arraystretch}{1.2} 
    \resizebox{\textwidth}{!}{%
    \begin{tabular}{cc||>{\centering\arraybackslash}p{1.8cm} >{\centering\arraybackslash}p{1.8cm} >{\centering\arraybackslash}p{1.8cm}|>{\centering\arraybackslash}p{1.8cm} >{\centering\arraybackslash}p{1.8cm} >{\centering\arraybackslash}p{1.8cm}|>{\centering\arraybackslash}p{2cm} >{\centering\arraybackslash}p{2cm} >{\centering\arraybackslash}p{2cm}}
    \toprule
        \multirow{3}{*}{\textbf{Method}} & \multirow{3}{*}{\textbf{Base Detector}} & 
        \multicolumn{3}{c|}{\multirow{2}{*}{\textbf{VOC [1:10]} $\rightarrow$ \textbf{Clipart [11:20]}}} & 
        \multicolumn{3}{c|}{\multirow{2}{*}{\makecell{\textbf{Watercolor [1:3]} $\rightarrow$ \textbf{Comic [4:6]} \\ $\rightarrow$ \textbf{Clipart [7:13]} $\rightarrow$ \textbf{VOC [14:20]}}}} & 
        \multicolumn{3}{c}{\multirow{2}{*}{\makecell{\textbf{Night Sunny [1:2]} $\rightarrow$ \textbf{Daytime Sunny [3:4]} \\ $\rightarrow$ \textbf{Daytime Foggy [5:7]}}}} \\
        & & & & & & & & & & \\
        \cline{3-11}
        & & \textbf{Avg RI (\%)} & \textbf{Avg GI (\%)} & \textbf{RAI (\%)} & \textbf{Avg RI (\%)} & \textbf{Avg GI (\%)} & \textbf{RAI (\%)} & \textbf{Avg RI (\%)} & \textbf{Avg GI (\%)} & \textbf{RAI (\%)} \\
        \hline
        Sequential FT & YOLO11n & 0.75\scriptsize{$\pm$0.2} & 12.86\scriptsize{$\pm$0.4} & 6.81\scriptsize{$\pm$0.3} & 0.00\scriptsize{$\pm$0.0} & 11.05\scriptsize{$\pm$0.5} & 5.53\scriptsize{$\pm$0.3} & 0.00\scriptsize{$\pm$0.0} & 30.51\scriptsize{$\pm$0.6} & 15.26\scriptsize{$\pm$0.5} \\
        LwF \cite{LwF} & YOLO11n & \textit{\underline{72.64}}\scriptsize{$\pm$0.3} & 33.74\scriptsize{$\pm$0.5} & 53.19\scriptsize{$\pm$0.4} & 52.66\scriptsize{$\pm$0.6} & 17.01\scriptsize{$\pm$0.4} & 34.84\scriptsize{$\pm$0.3} & 27.94\scriptsize{$\pm$0.7} & 23.78\scriptsize{$\pm$0.5} & 25.86\scriptsize{$\pm$0.6} \\
        ERD \cite{ERD} & YOLO11n & 68.66\scriptsize{$\pm$0.4} & \textit{\underline{43.68}}\scriptsize{$\pm$0.3} & \textit{\underline{56.17}}\scriptsize{$\pm$0.6} & 54.76\scriptsize{$\pm$0.5} & \textit{\underline{41.13}}\scriptsize{$\pm$0.4} & 47.95\scriptsize{$\pm$0.7} & 44.60\scriptsize{$\pm$0.6} & 39.40\scriptsize{$\pm$0.5} & 42.00\scriptsize{$\pm$0.3} \\
        LDB\footnotemark[2] \cite{LDB} & ViTDet & 66.89\scriptsize{$\pm$0.5} & 18.76\scriptsize{$\pm$0.3} & 42.83\scriptsize{$\pm$0.4} & \textit{\underline{86.08}}\scriptsize{$\pm$0.6} & 19.57\scriptsize{$\pm$0.5} & 52.83\scriptsize{$\pm$0.4} & 50.50\scriptsize{$\pm$0.6} & 5.42\scriptsize{$\pm$0.7} & 27.96\scriptsize{$\pm$0.5} \\
        CL-DETR\footnotemark[2] \cite{CL-DETR} & Deformable DETR & 68.29\scriptsize{$\pm$0.3} & 40.72\scriptsize{$\pm$0.4} & 54.51\scriptsize{$\pm$0.3} & 71.73\scriptsize{$\pm$0.5} & 36.63\scriptsize{$\pm$0.6} & \textit{\underline{54.18}}\scriptsize{$\pm$0.4} & \textit{\underline{64.26}}\scriptsize{$\pm$0.3} & \textbf{43.46}\scriptsize{$\pm$0.5} & \textit{\underline{53.86}}\scriptsize{$\pm$0.6} \\
        \rowcolor{lightblue} \textbf{DuET (Ours)} & YOLO11n & \textbf{87.44}\scriptsize{$\pm$0.2} & \textbf{44.54}\scriptsize{$\pm$0.1} & \textbf{65.99}\scriptsize{$\pm$0.3} & \textbf{89.30}\scriptsize{$\pm$0.2} & \textbf{42.60}\scriptsize{$\pm$0.1} & \textbf{65.95}\scriptsize{$\pm$0.2} & \textbf{88.57}\scriptsize{$\pm$0.2} & \textit{\underline{41.92}}\scriptsize{$\pm$0.1} & \textbf{65.25}\scriptsize{$\pm$0.2} \\
    \bottomrule
    \end{tabular}
    }
    \vspace{-4mm}
\end{table*}
\begin{table}[!ht]
    \centering
    \caption{Results of proposed DuET framework on Daytime Sunny [1:4] $\rightarrow$ Night Sunny [5:7] experiment with different base detectors. Best among the columns in \textbf{bold}, second best \textit{\underline{underlined}}.}
    \label{all_detectors}
    \vspace{-0.5em}
    \resizebox{\columnwidth}{!}{%
    \begin{tabular}{cc|cc|ccc}
    \toprule
        \multirow{2}{*}{\makecell{\textbf{Method}}} &
        \multirow{2}{*}{\makecell{\textbf{Base Detector}}} &
        \multirow{2}{*}{\makecell{\textbf{Trainable} \\ \textbf{Params (M)}}} &
        \multirow{2}{*}{\makecell{\textbf{GFLOPs}}} &
        \multirow{2}{*}{\makecell{\textbf{Avg RI} \\ \textbf{(\%)}}} &
        \multirow{2}{*}{\makecell{\textbf{Avg GI} \\ \textbf{(\%)}}} &
        \multirow{2}{*}{\makecell{\textbf{RAI} \\ \textbf{(\%)}}} \\
        & & & & & \\
        \hline
        \multirow{5}{*}{\makecell{\textbf{DuET} \\ \textbf{(Ours)}}} 
        & ViTDet \cite{VitDet} & 110.52 & 1829.61 & 27.55\tiny{$\pm$0.3} & 28.22\tiny{$\pm$0.8} & 27.89\tiny{$\pm$0.5} \\
        & Deformable DETR \cite{DeformableDETR} & 39.85 & \textit{\underline{11.77}} &  84.45\tiny{$\pm$0.2} & 33.45\tiny{$\pm$0.1} & 58.95\tiny{$\pm$0.2} \\
        & RT-DETR-l \cite{RT-DETR} & \textit{\underline{32.00}} & 103.4 & 47.73\tiny{$\pm$0.2} & 21.00\tiny{$\pm$0.1} & 34.37\tiny{$\pm$0.2} \\
        & RT-DETR-x \cite{RT-DETR} & 65.49 & 222.5 & 56.39\tiny{$\pm$0.2} & 24.15\tiny{$\pm$0.1} & 40.27\tiny{$\pm$0.2} \\
        & YOLO11n \cite{YOLO11} & \textbf{2.58} & \textbf{6.3} &\textit{\underline{88.06}}\tiny{$\pm$0.2} & \textbf{56.95}\tiny{$\pm$0.1} & \textbf{72.51}\tiny{$\pm$0.2} \\
        & YOLO11x \cite{YOLO11} & 56.84 & 194.4 & \textbf{96.88}\tiny{$\pm$0.2} & \textit{\underline{42.41}}\tiny{$\pm$0.1} & \textit{\underline{69.18}}\tiny{$\pm$0.2} \\
    \bottomrule
    \end{tabular}
    }
    \vspace{-2mm}
\end{table}
\begin{table}[t!]
    \centering
    \caption{Ablations of DuET framework with different components and losses on VOC [1:10] \(\rightarrow\) Clipart [11:20] with YOLO11n as base detector.}
    \label{main_ablations_table}
    \vspace{-0.5em}
    \resizebox{\columnwidth}{!}{
    \begin{tabular}{ccccc|ccc}
    \toprule
        \multirow{2}{*}{\makecell{\textbf{Seq FT}}} &
        \multirow{2}{*}{\makecell{\textbf{Incremental} \\ \textbf{Head}}} &
        \multirow{2}{*}{\makecell{\textbf{DuET} \\ \textbf{Module}}} &
        \multirow{2}{*}{\makecell{$\boldsymbol{\mathcal{L}_{\text{\textbf{Distill}}}^{*}}$}} &
        \multirow{2}{*}{\makecell{$\boldsymbol{\mathcal{L}_{\text{\textbf{DC}}}}$}} &
        \multirow{2}{*}{\makecell{\textbf{Avg RI} \\ \textbf{(\%)}}} &
        \multirow{2}{*}{\makecell{\textbf{Avg GI} \\ \textbf{(\%)}}} &
        \multirow{2}{*}{\makecell{\textbf{RAI} \\ \textbf{(\%)}}} \\
        & & & & & & & \\
        \hline
        \xmark & \xmark & \xmark & \xmark & \xmark & 0.5 & 9.13 & 4.82 \\
        \cmark & \xmark & \xmark & \xmark & \xmark & 0.75 & 12.86 & 6.81 \\  
        \cmark & \cmark & \xmark & \xmark & \xmark & 24.75 & 33.36 & 29.06 \\ 
        \cmark & \cmark & \cmark & \xmark & \xmark & 75.00 & 37.26 & 56.13 \\
        \cmark & \cmark & \cmark & \cmark & \xmark & 87.06 & 37.75 & 62.41 \\ 
        \rowcolor{lightblue} \cmark & \cmark & \cmark & \cmark & \cmark & \textbf{87.44} & \textbf{44.54} & \textbf{65.99} \\ 
    \bottomrule
    \end{tabular}
    }
    \vspace{-6mm}
\end{table}
\begin{figure*}[ht]
\centerline{\includegraphics[width=\textwidth]{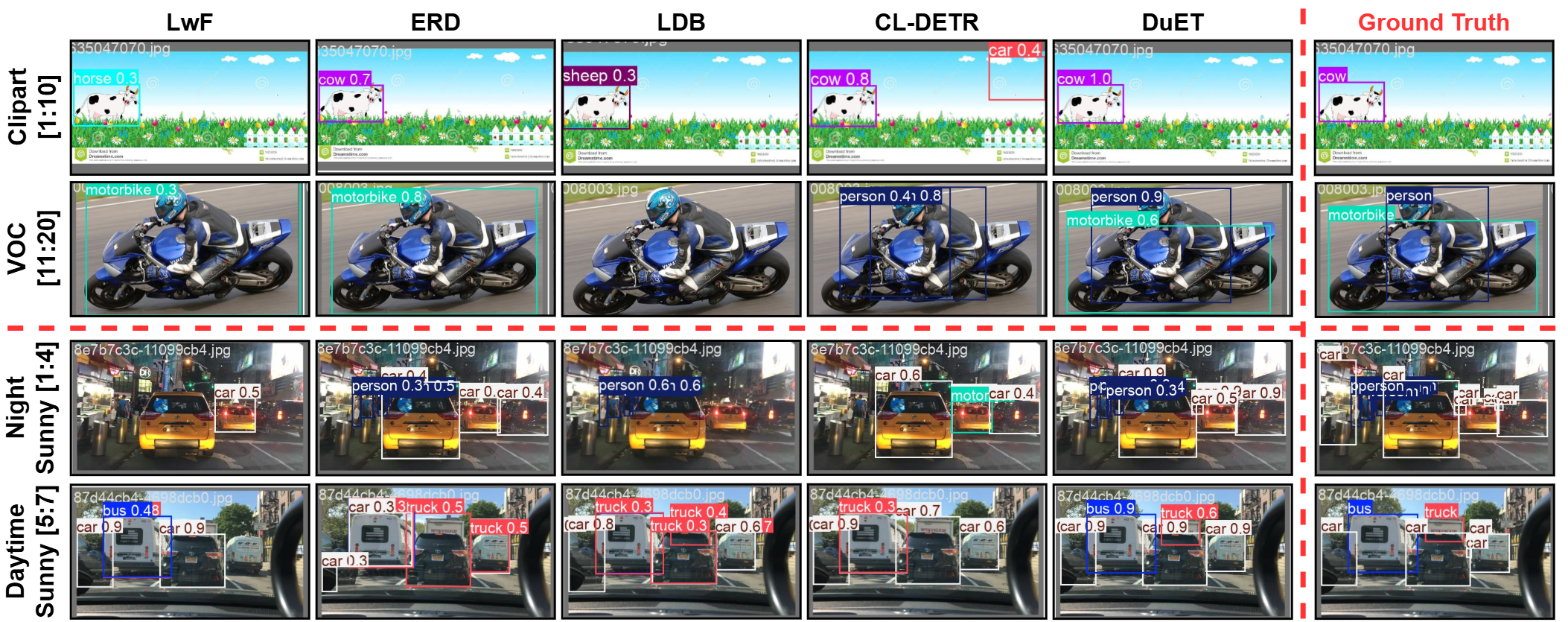}}
\vspace{-0.5em}
\caption{\textbf{Qualitative comparisons on two-phase experiments for different methods on the DuIOD task.} The \emph{\textbf{top two rows}} display detection results on unseen classes: Clipart [1:10] \& VOC [11:20], for: \textit{VOC [1:10] $\rightarrow$ Clipart [11:20]} experiment. Similarly, the \emph{\textbf{bottom two rows}} show detection results on unseen classes: Night Sunny[1:4] \& Daytime Sunny[5:7], for: \textit{Daytime Sunny[1:4] $\rightarrow$ Night Sunny[5:7]} experiment (zoomed-in for best view).}
\vspace{-5mm}
\label{fig:qual_results}
\end{figure*}
\begin{figure}[h]
\begin{center}
\centerline{\includegraphics[width=\columnwidth]{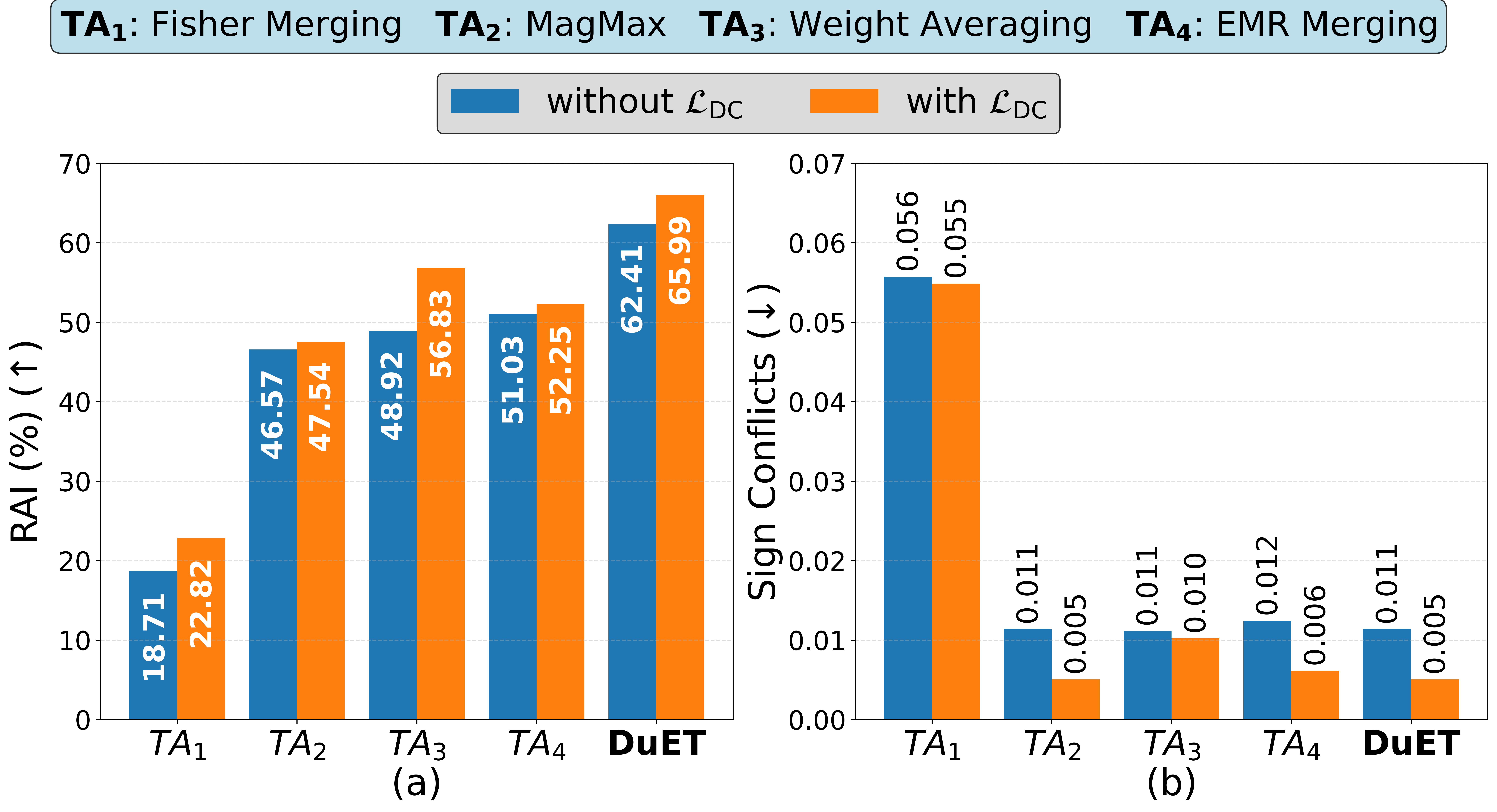}}
\vspace{-1em}
\caption{Impact of \(\mathcal{L}_{DC}\) on (a) improving RAI and (b) reducing sign conflicts across different TA approaches, evaluated on VOC [1:10] $\rightarrow$ Clipart [11:20] using YOLO11n with Incremental Head and sequential fine-tuning.}
\vspace{-12mm}
\label{fig:RAI_Sign_Conf}
\end{center}
\end{figure}
\vspace{-2mm}
\section{Results \& Discussion}
\subsection{Quantitative Results}
To evaluate the effectiveness of the DuET, we compare it against five baselines: Sequential Fine-Tuning, LwF \cite{LwF} and ERD \cite{ERD} (common IOD methods), LDB \cite{LDB} and CL-DETR \cite{CL-DETR} (\textbf{current open-source SOTA} in DIOD and CIOD). We re-implemented all under the proposed DuIOD setting, reporting the best results across seven experiments—five two-phase and two multi-phase. Results from two two-phase and two multi-phase experiments are in Table~\ref{E4_expanded} and Table~\ref{E1_E3_E6_combined}, while comprehensive results from all experiments are provided in \texttt{Appendix E}. Moreover, as detailed in \texttt{Appendix C3}, DuET maintains a consistent performance irrespective of the randomness in class-domain orders.

We begin with naïve \textbf{Sequential Fine-Tuning} using YOLO11n \cite{YOLO11} as the base detector, where we observe severe catastrophic forgetting, with Avg RI~\(< 1\%\) across all experiments. \textbf{LwF} and \textbf{ERD}, both implemented on YOLO11n by modifying the Ultralytics pipeline \cite{YOLO11} to incorporate distillation losses (LwF) and the Elastic Response Selection algorithm (ERD), show moderate RAI improvement but gives poor results in multi-phase experiments (Table~\ref{E1_E3_E6_combined}) compared to DuET. On the DIOD side, \textbf{LDB}\footnotemark[2] with VitDet \cite{VitDet} as the base detector suffers from severe catastrophic forgetting, particularly in cross-domain uneven class shifts (Table~\ref{E4_expanded}). On the CIOD side, \textbf{CL-DETR}\footnotemark[2] with Deformable DETR \cite{DeformableDETR} struggles under severe domain shifts despite having more trainable parameters. 
In contrast, DuET consistently outperforms all methods despite having significantly fewer trainable parameters (Table~\ref{E4_expanded}) while also demonstrating strong generalization across multiple object detection architectures (Table~\ref{all_detectors}). 

DuET achieves an RAI improvement of \textbf{+9.82\%} and \textbf{+13.12\%} while preserving \textbf{87.44\%} and \textbf{89.30\%} Avg RI on the \emph{Pascal Series} two-phase and multi-phase experiments, respectively. Similarly, on the Diverse Weather Series, DuET achieves an RAI improvement of \textbf{+12.59\%} and \textbf{+11.39\%} while preserving \textbf{88.06\%} and \textbf{88.57\%} Avg RI on the two-phase and multi-phase experiments, respectively. 
\footnotetext[2]{CL-DETR~\cite{CL-DETR} relies on a DETR-based framework with bipartite matching and cannot be trivially adapted to YOLO11. LDB~\cite{LDB}, designed for learning domain biases, was unable to handle simultaneous class-domain shifts of DuIOD with YOLO11 (often yielding 0 mAP and at least a 40\% drop in RAI compared to its original backbone), so we compare each method on its own backbone.}

\subsection{Qualitative Analysis}
In Figure~\ref{fig:qual_results}, we present qualitative comparisons of different methods on two-phase experiments, while additional qualitative results on multi-phase experiments are provided in \texttt{Appendix E2}. In \textit{VOC [1:10] $\rightarrow$ Clipart [11:20]}, LwF and LDB misclassify the \texttt{cow} class (introduced in \(\mathcal{T}_1\): VOC [1:10]) as \texttt{horse} or \texttt{sheep} (both introduced in \(\mathcal{T}_2\): Clipart [11:20]). While ERD and CL-DETR correctly classify it, they exhibit significantly lower confidence scores compared to DuET. A similar pattern is observed for the \texttt{person} and \texttt{motorbike} classes, both introduced in \(\mathcal{T}_2\), where only DuET correctly detects both objects in the unseen VOC [11:20], while competing methods mostly fail.

Similarly, in \textit{Daytime Sunny [1:4] $\rightarrow$ Night Sunny [5:7]}, the \texttt{car} class (introduced in \(\mathcal{T}_1\): Daytime Sunny [1:4]) undergoes a severe domain shift when evaluated in unseen Night Sunny [1:4]. As a result, most methods fail to detect a majority of \texttt{car} instances, instead misclassifying them as background, because of the background shift challenge in DuIOD. (see \texttt{Appendix F} for more details). However, DuET effectively overcomes this challenge by correctly detecting most of the \texttt{car} instances (improving $\text{mAP}_{\text{car}}^{\mathcal{T}_2}$ from \textbf{49.4} to \textbf{71.8}). Similar misclassifications are observed for the \texttt{truck} class, introduced in \(\mathcal{T}_2\): Night Sunny [5:7], where most methods struggle to detect it in unseen Daytime Sunny [5:7], while DuET provides consistent detections. These qualitative results highlight the effectiveness of DuET in seamlessly adapting to unseen classes in new domains.
\vspace{-2mm}
\section{Main Ablation Studies}
\label{ablations}
This section covers the main ablation studies, while extended ablations, including the \textbf{impact of loss components}, \textbf{hyperparameter sensitivity analysis}, the \textbf{influence of random class-domain order}, and \textbf{complexity analysis}, are provided in \texttt{Appendix C}. 
\vspace{-4mm}
\paragraph{Role of Sequential Fine-tuning.}
While DuIOD inherently involves sequential learning, we compare independent (first row, Table \ref{main_ablations_table}) vs. sequential fine-tuning (second row) in DuET framework, observing a significant improvement in Avg GI and, hence, RAI with the latter, aligning with the findings in \cite{MAGMAX}, highlighting its importance in DuIOD.
\vspace{-4mm}
\paragraph{Effectiveness of DuET Module and Incremental Head.}
Next, we evaluate the effectiveness of the DuET Module and Incremental Head in the DuET framework. Incremental Head (third row) brings a substantial improvement in both Avg RI and Avg GI over Seq FT. This highlights the importance of concatenating task-specific parameters from previous and current tasks, thereby enhancing the model’s generalization capability. Incorporating the DuET Module further leads to a significant improvement in both Avg RI and Avg GI (fourth row), effectively mitigating catastrophic forgetting while enhancing generalization.
\vspace{-4mm}
\paragraph{Impact of $\mathcal{L}_{\text{Distill}}^{*}$ and $\mathcal{L}_{\text{DC}}$.}
We further investigate the impact of $\mathcal{L}_{\text{Distill}}^{*}$ and $\mathcal{L}_{\text{DC}}$ in the proposed DuET approach. As shown in Table \ref{main_ablations_table} (rows five and six), both losses contribute significantly. While $\mathcal{L}_{\text{Distill}}^{*}$ brings a substantial improvement in Avg RI by helping to retain knowledge from previous tasks, $\mathcal{L}_{\text{DC}}$ further enhances Avg GI, thereby improving the model’s generalization capability. The combination of both losses yields the best performance, achieving the highest Avg RI and Avg GI, highlighting their critical role in the DuET approach.
\vspace{-4mm}
\paragraph{Comparison with other model-merging algorithms.}
In Figure \ref{fig:RAI_Sign_Conf}, we compare the performance of related TA approaches with the proposed approach. The results indicate that $\mathcal{L}_{\text{DC}}$ improves RAI by an average of \textbf{9.67\%} while significantly reducing sign conflicts by an average of \textbf{34\%}, averaged across all TA approaches, with DuET achieving the best overall performance.

\section{Conclusions}
In this paper, we introduced DuET, a task-arithmetic-driven solution for robust IOD, enabling models to learn new object categories while generalizing across evolving domains incrementally. Unlike previous works that focus on either class or domain shifts, DuET integrates both, making it practical for real-world applications like autonomous driving and surveillance. We demonstrated its effectiveness across diverse object detectors like YOLO11 \& RT-DETR through extensive experiments on the \emph{Pascal Series} and \emph{Diverse Weather Series} datasets. Additionally, we proposed Directional Consistency Loss to mitigate sign conflicts and a Retention-Adaptability Index to comprehensively evaluate retention and adaptation. DuET opens a new research direction in IOD, providing a promising framework for continual learning and domain adaptation.
{
    \clearpage
    \newpage
    \small
    \bibliographystyle{ieeenat_fullname}
    \bibliography{main}

\begin{thebibliography}{53}
\providecommand{\natexlab}[1]{#1}
\providecommand{\url}[1]{\texttt{#1}}
\expandafter\ifx\csname urlstyle\endcsname\relax
  \providecommand{\doi}[1]{doi: #1}\else
  \providecommand{\doi}{doi: \begingroup \urlstyle{rm}\Url}\fi

\bibitem[Acharya et~al.(2020)Acharya, Hayes, and Kanan]{Rodeo}
Manoj Acharya, Tyler~L Hayes, and Christopher Kanan.
\newblock Rodeo: Replay for online object detection.
\newblock \emph{arXiv preprint arXiv:2008.06439}, 2020.

\bibitem[Cermelli et~al.(2022)Cermelli, Geraci, Fontanel, and Caputo]{cermelli2022modeling}
Fabio Cermelli, Antonino Geraci, Dario Fontanel, and Barbara Caputo.
\newblock Modeling missing annotations for incremental learning in object detection.
\newblock In \emph{Proceedings of the IEEE/CVF Conference on Computer Vision and Pattern Recognition}, pages 3700--3710, 2022.

\bibitem[Chen et~al.(2020)Chen, Wang, Chen, Cai, and Qian]{chen2020incremental}
Jingzhou Chen, Shihao Wang, Ling Chen, Haibin Cai, and Yuntao Qian.
\newblock Incremental detection of remote sensing objects with feature pyramid and knowledge distillation.
\newblock \emph{IEEE Transactions on Geoscience and Remote Sensing}, 60:\penalty0 1--13, 2020.

\bibitem[Chen et~al.(2019)Chen, Yu, and Chen]{chen2019new}
Li Chen, Chunyan Yu, and Lvcai Chen.
\newblock A new knowledge distillation for incremental object detection.
\newblock In \emph{2019 International Joint Conference on Neural Networks (IJCNN)}, pages 1--7. IEEE, 2019.

\bibitem[Cordts et~al.(2016)Cordts, Omran, Ramos, Rehfeld, Enzweiler, Benenson, Franke, Roth, and Schiele]{FoggyCityscapes}
Marius Cordts, Mohamed Omran, Sebastian Ramos, Timo Rehfeld, Markus Enzweiler, Rodrigo Benenson, Uwe Franke, Stefan Roth, and Bernt Schiele.
\newblock The cityscapes dataset for semantic urban scene understanding.
\newblock In \emph{Proceedings of the IEEE conference on computer vision and pattern recognition}, pages 3213--3223, 2016.

\bibitem[Danish et~al.(2024)Danish, Khan, Munir, Sarfraz, and Ali]{DivAlign}
Muhammad~Sohail Danish, Muhammad~Haris Khan, Muhammad~Akhtar Munir, M~Saquib Sarfraz, and Mohsen Ali.
\newblock Improving single domain-generalized object detection: A focus on diversification and alignment.
\newblock In \emph{Proceedings of the IEEE/CVF Conference on Computer Vision and Pattern Recognition}, pages 17732--17742, 2024.

\bibitem[Everingham et~al.(2010)Everingham, Van~Gool, Williams, Winn, and Zisserman]{PASCAL_VOC}
Mark Everingham, Luc Van~Gool, Christopher~KI Williams, John Winn, and Andrew Zisserman.
\newblock The pascal visual object classes (voc) challenge.
\newblock \emph{International journal of computer vision}, 88:\penalty0 303--338, 2010.

\bibitem[Feng et~al.(2022)Feng, Wang, and Yuan]{ERD}
Tao Feng, Mang Wang, and Hangjie Yuan.
\newblock Overcoming catastrophic forgetting in incremental object detection via elastic response distillation.
\newblock In \emph{Proceedings of the IEEE/CVF conference on computer vision and pattern recognition}, pages 9427--9436, 2022.

\bibitem[Gupta et~al.(2022)Gupta, Narayan, Joseph, Khan, Khan, and Shah]{OW-DETR}
Akshita Gupta, Sanath Narayan, KJ Joseph, Salman Khan, Fahad~Shahbaz Khan, and Mubarak Shah.
\newblock Ow-detr: Open-world detection transformer.
\newblock In \emph{Proceedings of the IEEE/CVF conference on computer vision and pattern recognition}, pages 9235--9244, 2022.

\bibitem[Hao et~al.(2019)Hao, Fu, Jiang, and Tian]{hao2019end}
Yu Hao, Yanwei Fu, Yu-Gang Jiang, and Qi Tian.
\newblock An end-to-end architecture for class-incremental object detection with knowledge distillation.
\newblock In \emph{2019 IEEE International Conference on Multimedia and Expo (ICME)}, pages 1--6. IEEE, 2019.

\bibitem[Hassaballah et~al.(2020)Hassaballah, Kenk, Muhammad, and Minaee]{AdverseWeather}
Mahmoud Hassaballah, Mourad~A Kenk, Khan Muhammad, and Shervin Minaee.
\newblock Vehicle detection and tracking in adverse weather using a deep learning framework.
\newblock \emph{IEEE transactions on intelligent transportation systems}, 22\penalty0 (7):\penalty0 4230--4242, 2020.

\bibitem[Huang et~al.(2024)Huang, Ye, Chen, He, Yue, and Ouyang]{EMR-Merging}
Chenyu Huang, Peng Ye, Tao Chen, Tong He, Xiangyu Yue, and Wanli Ouyang.
\newblock Emr-merging: Tuning-free high-performance model merging.
\newblock \emph{arXiv preprint arXiv:2405.17461}, 2024.

\bibitem[Ilharco et~al.(2022)Ilharco, Ribeiro, Wortsman, Gururangan, Schmidt, Hajishirzi, and Farhadi]{TA_Paper_0}
Gabriel Ilharco, Marco~Tulio Ribeiro, Mitchell Wortsman, Suchin Gururangan, Ludwig Schmidt, Hannaneh Hajishirzi, and Ali Farhadi.
\newblock Editing models with task arithmetic.
\newblock \emph{arXiv preprint arXiv:2212.04089}, 2022.

\bibitem[Inoue et~al.(2018)Inoue, Furuta, Yamasaki, and Aizawa]{PASCAL_Series}
Naoto Inoue, Ryosuke Furuta, Toshihiko Yamasaki, and Kiyoharu Aizawa.
\newblock Cross-domain weakly-supervised object detection through progressive domain adaptation.
\newblock In \emph{Proceedings of the IEEE conference on computer vision and pattern recognition}, pages 5001--5009, 2018.

\bibitem[Jia et~al.(2024)Jia, Wu, Fang, Zeng, Zhang, and Li]{Purified_Distillation}
Shilong Jia, Tingting Wu, Yingying Fang, Tieyong Zeng, Guixu Zhang, and Zhi Li.
\newblock Purified distillation: Bridging domain shift and category gap in incremental object detection.
\newblock In \emph{Proceedings of the 32nd ACM International Conference on Multimedia}, pages 1197--1205, 2024.

\bibitem[Jin et~al.(2022)Jin, Ren, Preotiuc-Pietro, and Cheng]{RegMean}
Xisen Jin, Xiang Ren, Daniel Preotiuc-Pietro, and Pengxiang Cheng.
\newblock Dataless knowledge fusion by merging weights of language models.
\newblock \emph{arXiv preprint arXiv:2212.09849}, 2022.

\bibitem[Jocher and Qiu(2024)]{YOLO11}
Glenn Jocher and Jing Qiu.
\newblock Ultralytics yolo11, 2024.

\bibitem[Kang et~al.(2023)Kang, Zhang, Zhang, Wang, Chen, Ma, and Huang]{kang2023alleviating}
Mengxue Kang, Jinpeng Zhang, Jinming Zhang, Xiashuang Wang, Yang Chen, Zhe Ma, and Xuhui Huang.
\newblock Alleviating catastrophic forgetting of incremental object detection via within-class and between-class knowledge distillation.
\newblock In \emph{Proceedings of the IEEE/CVF International Conference on Computer Vision}, pages 18894--18904, 2023.

\bibitem[Kim et~al.(2024{\natexlab{a}})Kim, Cho, Kim, Tiruneh, and Baek]{SDDGR}
Junsu Kim, Hoseong Cho, Jihyeon Kim, Yihalem~Yimolal Tiruneh, and Seungryul Baek.
\newblock Sddgr: Stable diffusion-based deep generative replay for class incremental object detection.
\newblock In \emph{Proceedings of the IEEE/CVF Conference on Computer Vision and Pattern Recognition}, pages 28772--28781, 2024{\natexlab{a}}.

\bibitem[Kim et~al.(2024{\natexlab{b}})Kim, Ku, Kim, Cha, and Baek]{VLM-PL}
Junsu Kim, Yunhoe Ku, Jihyeon Kim, Junuk Cha, and Seungryul Baek.
\newblock Vlm-pl: Advanced pseudo labeling approach for class incremental object detection via vision-language model.
\newblock In \emph{Proceedings of the IEEE/CVF Conference on Computer Vision and Pattern Recognition}, pages 4170--4181, 2024{\natexlab{b}}.

\bibitem[Kiran et~al.(2022)Kiran, Pedersoli, Dolz, Blais-Morin, Granger, et~al.]{kiran2022incremental}
Madhu Kiran, Marco Pedersoli, Jose Dolz, Louis-Antoine Blais-Morin, Eric Granger, et~al.
\newblock Incremental multi-target domain adaptation for object detection with efficient domain transfer.
\newblock \emph{Pattern Recognition}, 129:\penalty0 108771, 2022.

\bibitem[Li et~al.(2019)Li, Tasci, Ghosh, Zhu, Zhang, and Heck]{RILOD}
Dawei Li, Serafettin Tasci, Shalini Ghosh, Jingwen Zhu, Junting Zhang, and Larry Heck.
\newblock Rilod: Near real-time incremental learning for object detection at the edge.
\newblock In \emph{Proceedings of the 4th ACM/IEEE Symposium on Edge Computing}, pages 113--126, 2019.

\bibitem[Li et~al.(2018)Li, Wu, Xu, and Shang]{li2018incremental}
Wei Li, Qingbo Wu, Linfeng Xu, and Chao Shang.
\newblock Incremental learning of single-stage detectors with mining memory neurons.
\newblock In \emph{2018 IEEE 4th International Conference on Computer and Communications (ICCC)}, pages 1981--1985. IEEE, 2018.

\bibitem[Li et~al.(2022)Li, Mao, Girshick, and He]{VitDet}
Yanghao Li, Hanzi Mao, Ross Girshick, and Kaiming He.
\newblock Exploring plain vision transformer backbones for object detection.
\newblock In \emph{European conference on computer vision}, pages 280--296. Springer, 2022.

\bibitem[Li and Hoiem(2017)]{LwF}
Zhizhong Li and Derek Hoiem.
\newblock Learning without forgetting.
\newblock \emph{IEEE transactions on pattern analysis and machine intelligence}, 40\penalty0 (12):\penalty0 2935--2947, 2017.

\bibitem[Liu et~al.(2020)Liu, Yang, Ravichandran, Bhotika, and Soatto]{liu2020multi}
Xialei Liu, Hao Yang, Avinash Ravichandran, Rahul Bhotika, and Stefano Soatto.
\newblock Multi-task incremental learning for object detection.
\newblock \emph{arXiv preprint arXiv:2002.05347}, 2020.

\bibitem[Liu et~al.(2023{\natexlab{a}})Liu, Cong, Goswami, Liu, and van~de Weijer]{ABR_IOD}
Yuyang Liu, Yang Cong, Dipam Goswami, Xialei Liu, and Joost van~de Weijer.
\newblock Augmented box replay: Overcoming foreground shift for incremental object detection.
\newblock In \emph{Proceedings of the IEEE/CVF International Conference on Computer Vision (ICCV)}, pages 11367--11377, 2023{\natexlab{a}}.

\bibitem[Liu et~al.(2023{\natexlab{b}})Liu, Schiele, Vedaldi, and Rupprecht]{CL-DETR}
Yaoyao Liu, Bernt Schiele, Andrea Vedaldi, and Christian Rupprecht.
\newblock Continual detection transformer for incremental object detection.
\newblock In \emph{Proceedings of the IEEE/CVF Conference on Computer Vision and Pattern Recognition}, pages 23799--23808, 2023{\natexlab{b}}.

\bibitem[Loshchilov et~al.(2017)Loshchilov, Hutter, et~al.]{AdamW}
Ilya Loshchilov, Frank Hutter, et~al.
\newblock Fixing weight decay regularization in adam.
\newblock \emph{arXiv preprint arXiv:1711.05101}, 5:\penalty0 5, 2017.

\bibitem[Lv et~al.(2023)Lv, Xu, Zhao, Wang, Wei, Cui, Du, Dang, and Liu]{RT-DETR}
Wenyu Lv, Shangliang Xu, Yian Zhao, Guanzhong Wang, Jinman Wei, Cheng Cui, Yuning Du, Qingqing Dang, and Yi Liu.
\newblock Detrs beat yolos on real-time object detection, 2023.

\bibitem[Marczak et~al.(2025)Marczak, Twardowski, Trzci{\'n}ski, and Cygert]{MAGMAX}
Daniel Marczak, Bart{\l}omiej Twardowski, Tomasz Trzci{\'n}ski, and Sebastian Cygert.
\newblock Magmax: Leveraging model merging for seamless continual learning.
\newblock In \emph{European Conference on Computer Vision}, pages 379--395. Springer, 2025.

\bibitem[Matena and Raffel(2022)]{Fisher-merging}
Michael~S Matena and Colin~A Raffel.
\newblock Merging models with fisher-weighted averaging.
\newblock \emph{Advances in Neural Information Processing Systems}, 35:\penalty0 17703--17716, 2022.

\bibitem[Menezes et~al.(2023)Menezes, de~Moura, Alves, and de~Carvalho]{continual_survey}
Angelo~G Menezes, Gustavo de Moura, C{\'e}zanne Alves, and Andr{\'e}~CPLF de Carvalho.
\newblock Continual object detection: a review of definitions, strategies, and challenges.
\newblock \emph{Neural networks}, 161:\penalty0 476--493, 2023.

\bibitem[Mirza et~al.(2022)Mirza, Masana, Possegger, and Bischof]{DISC}
M~Jehanzeb Mirza, Marc Masana, Horst Possegger, and Horst Bischof.
\newblock An efficient domain-incremental learning approach to drive in all weather conditions.
\newblock In \emph{Proceedings of the IEEE/CVF conference on computer vision and pattern recognition}, pages 3001--3011, 2022.

\bibitem[Pang et~al.(2019)Pang, Chen, Shi, Feng, Ouyang, and Lin]{pang2019libra}
Jiangmiao Pang, Kai Chen, Jianping Shi, Huajun Feng, Wanli Ouyang, and Dahua Lin.
\newblock Libra r-cnn: Towards balanced learning for object detection.
\newblock In \emph{Proceedings of the IEEE/CVF conference on computer vision and pattern recognition}, pages 821--830, 2019.

\bibitem[Park et~al.(2025)Park, Lee, and Park]{VIL}
Min-Yeong Park, Jae-Ho Lee, and Gyeong-Moon Park.
\newblock Versatile incremental learning: Towards class and domain-agnostic incremental learning.
\newblock In \emph{European Conference on Computer Vision}, pages 271--288. Springer, 2025.

\bibitem[Peng et~al.(2020)Peng, Zhao, and Lovell]{Faster_ILOD}
Can Peng, Kun Zhao, and Brian~C Lovell.
\newblock Faster ilod: Incremental learning for object detectors based on faster rcnn.
\newblock \emph{Pattern recognition letters}, 140:\penalty0 109--115, 2020.

\bibitem[Peng et~al.(2021)Peng, Zhao, Maksoud, Li, and Lovell]{SID}
Can Peng, Kun Zhao, Sam Maksoud, Meng Li, and Brian~C Lovell.
\newblock Sid: incremental learning for anchor-free object detection via selective and inter-related distillation.
\newblock \emph{Computer vision and image understanding}, 210:\penalty0 103229, 2021.

\bibitem[Robins(1995)]{cat-forget1}
Anthony Robins.
\newblock Catastrophic forgetting, rehearsal and pseudorehearsal.
\newblock \emph{Connection Science}, 7\penalty0 (2):\penalty0 123--146, 1995.

\bibitem[Shieh et~al.(2020)Shieh, Haq, Haq, Karam, Chondro, Gao, and Ruan]{shieh2020continual}
Jeng-Lun Shieh, Qazi Mazhar~ul Haq, Muhamad~Amirul Haq, Said Karam, Peter Chondro, De-Qin Gao, and Shanq-Jang Ruan.
\newblock Continual learning strategy in one-stage object detection framework based on experience replay for autonomous driving vehicle.
\newblock \emph{Sensors}, 20\penalty0 (23):\penalty0 6777, 2020.

\bibitem[Shmelkov et~al.(2017)Shmelkov, Schmid, and Alahari]{shmelkov2017incremental}
Konstantin Shmelkov, Cordelia Schmid, and Karteek Alahari.
\newblock Incremental learning of object detectors without catastrophic forgetting.
\newblock In \emph{Proceedings of the IEEE international conference on computer vision}, pages 3400--3409, 2017.

\bibitem[Song et~al.(2024)Song, He, Dong, and Gong]{LDB}
Xiang Song, Yuhang He, Songlin Dong, and Yihong Gong.
\newblock Non-exemplar domain incremental object detection via learning domain bias.
\newblock In \emph{Proceedings of the AAAI Conference on Artificial Intelligence}, pages 15056--15065, 2024.

\bibitem[Wang et~al.(2025)Wang, He, Dong, Gao, Wang, and Gong]{PINA}
Qiang Wang, Yuhang He, Songlin Dong, Xinyuan Gao, Shaokun Wang, and Yihong Gong.
\newblock Non-exemplar domain incremental learning via cross-domain concept integration.
\newblock In \emph{European Conference on Computer Vision}, pages 144--162. Springer, 2025.

\bibitem[Wang et~al.(2022)Wang, Huang, and Hong]{S-Prompts}
Yabin Wang, Zhiwu Huang, and Xiaopeng Hong.
\newblock S-prompts learning with pre-trained transformers: An occam’s razor for domain incremental learning.
\newblock \emph{Advances in Neural Information Processing Systems}, 35:\penalty0 5682--5695, 2022.

\bibitem[Wu and Deng(2022)]{Single-DGOD}
Aming Wu and Cheng Deng.
\newblock Single-domain generalized object detection in urban scene via cyclic-disentangled self-distillation.
\newblock In \emph{Proceedings of the IEEE/CVF Conference on computer vision and pattern recognition}, pages 847--856, 2022.

\bibitem[Yadav et~al.(2024)Yadav, Tam, Choshen, Raffel, and Bansal]{TIES_MERGING}
Prateek Yadav, Derek Tam, Leshem Choshen, Colin~A Raffel, and Mohit Bansal.
\newblock Ties-merging: Resolving interference when merging models.
\newblock \emph{Advances in Neural Information Processing Systems}, 36, 2024.

\bibitem[Yang et~al.(2022{\natexlab{a}})Yang, Zhou, Shi, Wu, and Wang]{RD-IOD}
Dongbao Yang, Yu Zhou, Wei Shi, Dayan Wu, and Weiping Wang.
\newblock Rd-iod: Two-level residual-distillation-based triple-network for incremental object detection.
\newblock \emph{ACM Transactions on Multimedia Computing, Communications, and Applications (TOMM)}, 18\penalty0 (1):\penalty0 1--23, 2022{\natexlab{a}}.

\bibitem[Yang et~al.(2022{\natexlab{b}})Yang, Zhou, Zhang, Sun, Wu, Wang, and Ye]{yang2022multi}
Dongbao Yang, Yu Zhou, Aoting Zhang, Xurui Sun, Dayan Wu, Weiping Wang, and Qixiang Ye.
\newblock Multi-view correlation distillation for incremental object detection.
\newblock \emph{Pattern Recognition}, 131:\penalty0 108863, 2022{\natexlab{b}}.

\bibitem[Yu et~al.(2020)Yu, Chen, Wang, Xian, Chen, Liu, Madhavan, and Darrell]{BDD100k}
Fisher Yu, Haofeng Chen, Xin Wang, Wenqi Xian, Yingying Chen, Fangchen Liu, Vashisht Madhavan, and Trevor Darrell.
\newblock Bdd100k: A diverse driving dataset for heterogeneous multitask learning.
\newblock In \emph{Proceedings of the IEEE/CVF conference on computer vision and pattern recognition}, pages 2636--2645, 2020.

\bibitem[Yu et~al.(2024)Yu, Yu, Yu, Huang, and Li]{DARE}
Le Yu, Bowen Yu, Haiyang Yu, Fei Huang, and Yongbin Li.
\newblock Language models are super mario: Absorbing abilities from homologous models as a free lunch.
\newblock In \emph{Forty-first International Conference on Machine Learning}, 2024.

\bibitem[Zhang et~al.(2021)Zhang, Sun, Zhang, and Xiao]{zhang2021incremental}
Nan Zhang, Zhigang Sun, Kai Zhang, and Li Xiao.
\newblock Incremental learning of object detection with output merging of compact expert detectors.
\newblock In \emph{2021 4th international conference on intelligent autonomous systems (ICoIAS)}, pages 1--7. IEEE, 2021.

\bibitem[Zhong et~al.(2025)Zhong, Jiao, and Bao]{zhong2025replay}
Jian Zhong, Yifan Jiao, and Bing-Kun Bao.
\newblock Replay-based incremental object detection with local response exploration.
\newblock \emph{IEEE Transactions on Multimedia}, 2025.

\bibitem[Zhu et~al.(2020)Zhu, Su, Lu, Li, Wang, and Dai]{DeformableDETR}
Xizhou Zhu, Weijie Su, Lewei Lu, Bin Li, Xiaogang Wang, and Jifeng Dai.
\newblock Deformable detr: Deformable transformers for end-to-end object detection.
\newblock \emph{arXiv preprint arXiv:2010.04159}, 2020.

\end{thebibliography}
}
\clearpage
\renewcommand{\thesection}{\Alph{section}}
\renewcommand{\thetable}{S\arabic{table}}
\renewcommand{\thefigure}{S\arabic{figure}}
\setcounter{section}{0}
\setcounter{figure}{0}
\setcounter{table}{0}

\maketitlesupplementary

\section{Detailed Evaluation Protocol and Metrics}
\label{sup_eval_protocol}
\begin{table*}[!ht]
    \centering
    \caption{Training Sequence \& Evaluation Protocol for different DuIOD experiments.}
    \label{tab:eval_protocol}
    \vspace{-2mm}
    \renewcommand{\arraystretch}{1.2}
    \resizebox{\textwidth}{!}{
    \begin{tabular}{c||c|c|c|c|c}
    \toprule
        \multirow{2}{*}{\makecell{\textbf{DuIOD} \\\textbf{Experiment}}} & 
        \multicolumn{2}{c|}{\textbf{Training Sequence}} &
        \multicolumn{3}{c}{\textbf{Evaluation Protocol}} \\
        \cline{2-6}
        & \textbf{Task} & \textbf{Class IDs} & \textbf{New Classes} & \textbf{Old Classes} & \textbf{Unseen Classes} \\
    \hline
    \rowcolor{lightblue} \multicolumn{6}{c}{\textbf{\textit{Pascal Series Datasets}}} \\
    \hline
         \multirow{3}{*}{\makecell{\textbf{Two Phase} \\ VOC [1:10] $\rightarrow$ \\ Clipart [11:20]}} 
        & $\mathcal{T}_1$ & 1-10 from VOC & $\text{mAP}_{\text{new}}^{\mathcal{T}_1}(\text{VOC}[1:10])$ & --- & --- \\
        \cline{2-6}
        & $\mathcal{T}_2$ & 11-20 from Clipart & $\text{mAP}_{\text{new}}^{\mathcal{T}_2}(\text{Clipart}[11:20])$ & $\text{mAP}_{\text{old}}^{\mathcal{T}_2}(\text{VOC}[1:10])$ & \makecell{$\text{mAP}_{\text{unseen}}^{\mathcal{T}_2}(\text{VOC}[11:20])$ \\  $\text{mAP}_{\text{unseen}}^{\mathcal{T}_2}(\text{Clipart}[1:10])$} \\
    \hline
    \hline
        \multirow{9}{*}{\makecell{\textbf{Multi Phase} \\ Watercolor [1:3] \\ $\rightarrow$ Comic [4:6] \\ $\rightarrow$ Clipart [7:13] \\ $\rightarrow$ VOC [14:20]}} 
        & $\mathcal{T}_1$ & 1-3 from Watercolor & $\text{mAP}_{\text{new}}^{\mathcal{T}_1}(\text{Watercolor}[1:3])$ & --- & --- \\
        \cline{2-6}
        & $\mathcal{T}_2$ & 4-6 from Comic & $\text{mAP}_{\text{new}}^{\mathcal{T}_2}(\text{Comic}[4:6])$ & $\text{mAP}_{\text{old}}^{\mathcal{T}_2}(\text{Watercolor}[1:3])$ & \makecell{$\text{mAP}_{\text{unseen}}^{\mathcal{T}_2}(\text{Watercolor}[4:6])$ \\ $\text{mAP}_{\text{unseen}}^{\mathcal{T}_2}(\text{Comic}[1:3])$} \\ 
        \cline{2-6}
        & $\mathcal{T}_3$ & 7-13 from Clipart & $\text{mAP}_{\text{new}}^{\mathcal{T}_3}(\text{Clipart}[7:13])$ & \makecell{$\text{mAP}_{\text{old}}^{\mathcal{T}_3}(\text{Watercolor}[1:3])$ \\ $\text{mAP}_{\text{old}}^{\mathcal{T}_3}(\text{Comic}[4:6])$} & \makecell{$\text{mAP}_{\text{unseen}}^{\mathcal{T}_3}(\text{Watercolor}[4:6])$ \\ $\text{mAP}_{\text{unseen}}^{\mathcal{T}_3}(\text{Comic}[1:3])$ \\ $\text{mAP}_{\text{unseen}}^{\mathcal{T}_3}(\text{Clipart}[1:6])$} \\
        \cline{2-6}
        & $\mathcal{T}_4$ & 14-20 from VOC & $\text{mAP}_{\text{new}}^{\mathcal{T}_4}(\text{VOC}[14:20])$ & \makecell{$\text{mAP}_{\text{old}}^{\mathcal{T}_4}(\text{Watercolor}[1:3])$ \\ $\text{mAP}_{\text{old}}^{\mathcal{T}_4}(\text{Comic}[4:6])$ \\  $\text{mAP}_{\text{old}}^{\mathcal{T}_4}(\text{Clipart}[7:13])$} & \makecell{$\text{mAP}_{\text{unseen}}^{\mathcal{T}_4}(\text{Watercolor}[4:6])$ \\ $\text{mAP}_{\text{unseen}}^{\mathcal{T}_4}(\text{Comic}[1:3])$ \\ $\text{mAP}_{\text{unseen}}^{\mathcal{T}_4}(\text{Clipart}[1:6])$ \\ $\text{mAP}_{\text{unseen}}^{\mathcal{T}_4}(\text{VOC}[1:13])$} \\
    \hline
    \rowcolor{lightblue} \multicolumn{6}{c}{\textbf{\textit{Diverse Weather Series Datasets}}} \\
    \hline
    \multirow{3}{*}{\makecell{\textbf{Two Phase} \\ Daytime Sunny [1:4]  \\ $\rightarrow$ Night Sunny [5:7]}} 
        & $\mathcal{T}_1$ & 1-4 from Daytime Sunny &$\text{mAP}_{\text{new}}^{\mathcal{T}_1}(\text{Daytime Sunny}[1:4])$ & --- & --- \\ 
    \cline{2-6}
        & $\mathcal{T}_2$ & 5-7 from Night Sunny & $\text{mAP}_{\text{new}}^{\mathcal{T}_2}(\text{Night Sunny}[5:7])$ & $\text{mAP}_{\text{old}}^{\mathcal{T}_2}(\text{Daytime Sunny}[1:4])$ & \makecell{$\text{mAP}_{\text{unseen}}^{\mathcal{T}_2}(\text{Daytime Sunny}[5:7])$ \\  $\text{mAP}_{\text{unseen}}^{\mathcal{T}_2}(\text{Night Sunny}[1:4])$} \\
    \hline
    \hline
    \multirow{6}{*}{\makecell{\textbf{Multi Phase} \\ Night Sunny [1:2] \\ $\rightarrow$ Daytime Sunny [3:4] \\ $\rightarrow$ Daytime Foggy [5:7]}} 
        & $\mathcal{T}_1$ & 1-2 from Night Sunny & $\text{mAP}_{\text{new}}^{\mathcal{T}_1}(\text{Night Sunny}[1:2])$ & --- & --- \\ 
    \cline{2-6}
        & $\mathcal{T}_2$ & 3-4 from Daytime Sunny & $\text{mAP}_{\text{new}}^{\mathcal{T}_2}(\text{Daytime Sunny}[3:4])$ & $\text{mAP}_{\text{old}}^{\mathcal{T}_2}(\text{Night Sunny}[1:2])$ & \makecell{$\text{mAP}_{\text{unseen}}^{\mathcal{T}_2}(\text{Night Sunny}[3:4])$ \\  $\text{mAP}_{\text{unseen}}^{\mathcal{T}_2}(\text{Daytime Sunny}[1:2])$} \\
    \cline{2-6}
        & $\mathcal{T}_3$ & 5-7 from Daytime Foggy & $\text{mAP}_{\text{new}}^{\mathcal{T}_3}(\text{Daytime Foggy}[5:7])$ & \makecell{$\text{mAP}_{\text{old}}^{\mathcal{T}_3}(\text{Night Sunny}[1:2])$ \\  $\text{mAP}_{\text{old}}^{\mathcal{T}_3}(\text{Daytime Sunny}[3:4])$} & \makecell{$\text{mAP}_{\text{unseen}}^{\mathcal{T}_3}(\text{Night Sunny}[3:4])$ \\ $\text{mAP}_{\text{unseen}}^{\mathcal{T}_3}(\text{Daytime Sunny}[1:2])$ \\  $\text{mAP}_{\text{unseen}}^{\mathcal{T}_3}(\text{Daytime Foggy}[1:4])$} \\
    \bottomrule
    \end{tabular}
    }
\end{table*}
This section provides a detailed discussion of the proposed evaluation protocol and metrics for the DuIOD task. Table~\ref{tab:eval_protocol} outlines the training sequence that is followed for four different DuIOD experiments, along with the detailed evaluation protocol that is used to comprehensively evaluate the performance of different object detectors on the respective DuIOD setting. Unlike existing metrics \cite{chen2019new, continual_survey, yang2022multi} that focus only on catastrophic forgetting, we used the \textbf{Retention-Adaptability Index (RAI)}, which balances both knowledge retention and generalisation to unseen categories across evolving domains. We define RAI as the mean of the Average Retention Index (Avg RI) and Average Generalisation Index (Avg GI), which are discussed in the sections below.
\begin{equation}
    RAI = \frac{\text{Avg RI} + \text{Avg GI}}{2}
\end{equation}
\subsection{Average Retention Index}
For each domain \( \mathcal{D}_i \) corresponding to task \( \mathcal{T}_i \) where \( i \in \{1, \dots, T - 1\} \), we define the Retention Index \( RI_{\mathcal{D}_i} \) as:
\begin{equation}
    RI_{\mathcal{D}_i} = \frac{\text{mAP}_{\text{old}}^{\mathcal{T}_T}(\mathcal{D}_i[\mathcal{C}_i])}{\text{mAP}_{\text{new}}^{\mathcal{T}_i}(\mathcal{D}_i[\mathcal{C}_i])}
\end{equation}
Here, \( \text{mAP}_{\text{old}}^{\mathcal{T}_T}(\mathcal{D}_i[\mathcal{C}_i]) \) denotes the mean Average Precision (mAP) at IoU threshold = 0.5 of the object detector at the final task \( \mathcal{T}_T \) on the classes \( \mathcal{C}_i \) which were learned from domain \( \mathcal{D}_i \), and \( \text{mAP}_{\text{new}}^{\mathcal{T}_i}(\mathcal{D}_i[\mathcal{C}_i]) \) is the mAP when classes \( \mathcal{C}_i \) from domain \( \mathcal{D}_i \) were first encountered and learned, at task \( \mathcal{T}_i \). The Avg RI is then calculated as:
\begin{equation}
    \text{Avg RI} = \frac{1}{T - 1} \sum_{i=1}^{T - 1} RI_{\mathcal{D}_i}.
\end{equation}
To illustrate this, consider the multi-phase experiment (Table~\ref{tab:eval_protocol}) with training sequence: \textit{Night Sunny [1:2] $\rightarrow$ Daytime Sunny [3:4] $\rightarrow$ Daytime Foggy [5:7]}. The Avg RI is computed as the mean of the Retention Index values for Night Sunny (NS) and Daytime Sunny (DS) domains at the final task \( \mathcal{T}_3 \) as follows:
\begin{equation}
    \scalebox{0.99}{ $
    RI_{\text{NS}} = \frac{\text{mAP}_{\text{old}}^{\mathcal{T}_3}(\text{NS}[1:2])}{\text{mAP}_{\text{new}}^{\mathcal{T}_1}(\text{NS}[1:2])}
    \quad
    RI_{\text{DS}} = \frac{\text{mAP}_{\text{old}}^{\mathcal{T}_3}(\text{DS}[3:4])}{\text{mAP}_{\text{new}}^{\mathcal{T}_2}(\text{DS}[3:4])}
    $ }
\end{equation}
\begin{equation}
    \text{Avg RI} = \frac{RI_{\text{NS}} + RI_{\text{DS}}}{2}
\end{equation}
Hence, in this case, a higher Avg RI indicates how effectively the object detector has retained past knowledge from old Night Sunny and Daytime Sunny domains in the final task \(\mathcal{T}_3\). Conversely, a lower value indicates significant catastrophic forgetting.

\subsection{Average Generalization Index}
The Generalisation Index (\( GI_{\mathcal{D}_i, \mathcal{T}_j} \)) quantifies how well the model detects unseen classes from domain \( \mathcal{D}_i \) at task \( \mathcal{T}_j \). These classes were not part of the training set for task \( \mathcal{T}_j \), meaning the model is required to generalise beyond its explicitly trained classes (see Table~\ref{tab:eval_protocol}). For a given domain \( \mathcal{D}_i \) at task \( \mathcal{T}_j \), the Generalization Index is computed as:
\begin{equation}
    GI_{\mathcal{D}_i, \mathcal{T}_j} = \frac{\text{mAP}_{\text{unseen}}^{\mathcal{T}_j}(\mathcal{D}_i[\mathcal{C}_{\text{unseen}}])}{\text{mAP}_{\text{ref}}(\mathcal{D}_i[\mathcal{C}_{\text{unseen}}])}
\end{equation}
Here, \( \text{mAP}_{\text{unseen}}^{\mathcal{T}_j}(\mathcal{D}_i[\mathcal{C}_{\text{unseen}}]) \) is the mAP of the model at task \( \mathcal{T}_j \) on the unseen classes \( \mathcal{C}_{\text{unseen}} \) from domain \( \mathcal{D}_i \), and \( \text{mAP}_{\text{ref}}(\mathcal{D}_i[\mathcal{C}_{\text{unseen}}]) \) is the reference mAP obtained by training the object detector solely on these unseen classes on domain \( \mathcal{D}_i \). The Average Generalisation Index (Avg GI) over all relevant domain-task pairs is then computed as:
\begin{equation}
    \text{Avg GI} = \frac{1}{\mathcal{N}} \sum_{(\mathcal{D}_i, \mathcal{T}_j)} GI_{\mathcal{D}_i, \mathcal{T}_j}
\end{equation}
where \( \mathcal{N} \) is the total number of unseen-class domain-task pairs considered in the evaluation. 

Continuing with the same example, the Avg GI is computed as the mean of the Generalization Index values for unseen classes across the Night Sunny (NS), Daytime Sunny (DS), and Daytime Foggy (DF) domains for a total of five domain-task pairs- two from task \( \mathcal{T}_2 \) and three from \( \mathcal{T}_3 \):
\begin{equation}
    \scalebox{0.9}{ $
        GI_{\text{NS}, \mathcal{T}_2} = \frac{\text{mAP}_{\text{unseen}}^{\mathcal{T}_2}(\text{NS}[3:4])}{\text{mAP}_{\text{ref}}(\text{NS}[3:4])}
        \quad
        GI_{\text{DS}, \mathcal{T}_2} = \frac{\text{mAP}_{\text{unseen}}^{\mathcal{T}_2}(\text{DS}[1:2])}{\text{mAP}_{\text{ref}}(\text{DS}[1:2])}
    $ }
\end{equation}
\begin{equation*}
    \scalebox{0.95}{ $
        GI_{\text{NS}, \mathcal{T}_3} = \frac{\text{mAP}_{\text{unseen}}^{\mathcal{T}_3}(\text{NS}[3:4])}{\text{mAP}_{\text{ref}}(\text{NS}[3:4])} 
        \quad GI_{\text{DS}, \mathcal{T}_3} = \frac{\text{mAP}_{\text{unseen}}^{\mathcal{T}_3}(\text{DS}[1:2])}{\text{mAP}_{\text{ref}}(\text{DS}[1:2])}
    $ }
\end{equation*}
\begin{equation}
    GI_{\text{DF}, \mathcal{T}_3} = \frac{\text{mAP}_{\text{unseen}}^{\mathcal{T}_3}(\text{DF}[1:4])}{\text{mAP}_{\text{ref}}(\text{DF}[1:4])} 
\end{equation}
\begin{equation}
\scalebox{0.99}{ $
\text{Avg GI} = \frac{GI_{\text{NS}, T_2} + GI_{\text{DS}, T_2} + GI_{\text{NS}, T_3} + GI_{\text{DS}, T_3} + GI_{\text{DF}, T_3}}{5}
$ }
\end{equation}
Hence, in this case, a higher Avg GI indicates better zero-shot generalisation to unseen categories: NS [3:4], DS [1:2], and DF [1:4] across incremental training. Conversely, a lower value suggests that the model is overfitting to seen classes and fails to generalise.
\section{Loss Function Formulation}
To ensure effective incremental learning in case of DuIOD, we employ a combination of standard detector loss \( \mathcal{L}_{\text{Detector}} \), a modified distillation loss \( \mathcal{L}_{\text{Distill}}^{*} \) (discussed below), and the Directional Consistency Loss \( \mathcal{L}_{\text{DC}} \) (discussed in main paper \texttt{Section 3.5}). This section details the formulation of \( \mathcal{L}_{\text{total}} \) using these loss components.

Knowledge distillation plays a crucial role in mitigating catastrophic forgetting during incremental learning. In our approach, we extend the standard distillation loss (\( \mathcal{L}_{\text{Distill}} \)) for incremental learning \cite{LwF} by incorporating a dynamic thresholding mechanism that filters low-confidence classification outputs and high-variance bounding box predictions from the old (previous task) model.

Let \( \mathcal{M}_{\theta_{t-1}} \) represent the previous task model, and \( \mathcal{M}_{\theta_t} \) be the current model being trained on \( \mathcal{T}_t \). Given input data for the current task \( \mathcal{X}_t \), the classification outputs and predicted bounding boxes from both models will be:
\begin{equation}
\mathbf{z}_{\text{curr}} = \mathcal{M}_{\theta_t}(x), \quad \mathbf{z}_{\text{old}} = \mathcal{M}_{\theta_{t-1}}(x)
\end{equation}
where \( \mathbf{z} = (\mathbf{c}, \mathbf{b}) \), with \( \mathbf{c} \) being classification logits and \( \mathbf{b} \) the predicted bounding box coordinates.

The classification distillation loss is computed as:
\begin{equation}
\mathcal{L}_{\text{Distill}_{\text{cls}}}^{*} = \frac{1}{| \mathcal{M}_{\text{cls}}^{*} |} \sum_{i \in \mathcal{M}^*} \left\| \mathbf{c}_{\text{curr}}^{(i)} - \mathbf{c}_{\text{old}}^{(i)} \right\|^2
\end{equation}
where \( \mathcal{M}_{\text{cls}}^{*} \) is the dynamically selected mask that excludes predictions with low confidence scores in \( \mathbf{c}_{\text{old}} \):
\begin{equation}
\mathcal{M}_{\text{cls}}^{*} = \{ i \mid \max(\mathbf{c}_{\text{old}}^{(i)}) \geq \tau_{\text{cls}} \}
\end{equation}
where \( \tau_{\text{cls}} \) is an adaptive threshold computed as the 75th percentile of \( \max(\mathbf{c}_{\text{old}}) \) values.

Similarly, the bounding box regression distillation loss is computed by computing the KL divergence between the softmax of bounding box outputs from the current and old models:
\begin{equation}
\scalebox{0.78}{ $
\mathcal{L}_{\text{Distill}_{\text{bbox}}}^{*} = \frac{1}{| \mathcal{M}_{\text{bbox}}^{*} |} \sum\limits_{j \in \mathcal{M}_{\text{bbox}}^{*}} \mathcal{D}_{\text{KL}} \Big( \text{Softmax}(\mathbf{b}_{\text{curr}}^{(j)}) \Big|\Big| \text{Softmax}(\mathbf{b}_{\text{old}}^{(j)}) \Big)
$}
\end{equation}
where \( \mathcal{M}_{\text{bbox}}^{*} \) filters out bounding boxes with high variance in \( \mathbf{b}_{\text{old}} \):
\begin{equation}
\mathcal{M}_{\text{bbox}}^{*} = \{ j \mid \text{Var}(\mathbf{b}_{\text{old}}^{(j)}) \leq \tau_{\text{bbox}} \}
\end{equation}
where \( \tau_{\text{bbox}} \) is an adaptive threshold computed as the 75th percentile of bounding box variance values.

The final modified distillation loss becomes:
\begin{equation}
\mathcal{L}_{\text{Distill}}^{*} = \mathcal{L}_{\text{Distill}_{\text{cls}}}^{*} + \mathcal{L}_{\text{Distill}_{\text{bbox}}}^{*}
\end{equation}

\( \mathcal{L}_{\text{Detector}} \) depends on the object detector used. In our framework, we augment the detector losses of YOLO11 \cite{YOLO11} and RT-DETR \cite{RT-DETR} object detectors with \(\mathcal{L}_{\text{Distill}}^{*}\) and \( \mathcal{L}_{\text{DC}} \) on the Ultralytics \cite{YOLO11} pipeline. In the case of YOLO11, the detection loss consists of classification loss, bounding box regression loss, and Distribution Focal Loss \cite{YOLO11}, while in the case of RT-DETR, the detection loss follows a Hungarian matching strategy, consisting of classification, bounding box, and Generalized IoU (GIoU) losses \cite{RT-DETR}.

Hence, the total loss for incremental tasks (\( t \geq 2 \)) is computed as:
\begin{equation}
\mathcal{L}_{\text{Total}} = \mathcal{L}_{\text{Detector}} + \lambda_{\text{Distill}} \mathcal{L}_{\text{Distill}}^{*} + \lambda_{\text{DC}} \mathcal{L}_{\text{DC}}
\end{equation}
where \( \lambda_{\text{Distill}} \) and \( \lambda_{\text{DC}} \) are scaling coefficients that control the impact of distillation and directional consistency losses, respectively.
\section{Extended Ablation Studies}
\subsection{Impact of Loss Components}
\begin{table}[!ht]
    \centering
    \caption{Performance comparison of different model-merging algorithms on VOC [1:10] \(\rightarrow\) Clipart [11:20] depicting the impact of $\mathcal{L}_{\text{DC}}$, with YOLO11n \cite{YOLO11} as the base detector. Among columns, best in \textbf{bold}, second best \textit{\underline{underlined}}.}
    \label{table:ta_ablation_detailed}
    \vspace{-2mm}
    \resizebox{\columnwidth}{!}{%
    \begin{tabular}{cc||ccc}
    \toprule
        \multirow{2}{*}{\makecell{\textbf{Model-merging} \\ \textbf{algorithm}}} &
        \multirow{2}{*}{\makecell{\textbf{$\mathcal{L}_{\text{DC}}$}}} &
        \multirow{2}{*}{\makecell{\textbf{Avg RI} \\ \textbf{(\%)}}} &
        \multirow{2}{*}{\makecell{\textbf{Avg GI} \\ \textbf{(\%)}}} &
        \multirow{2}{*}{\makecell{\textbf{RAI} \\ \textbf{(\%)}}} \\
        & & & & \\
        \hline
        \hline
        Fisher-Merging \cite{Fisher-merging} & \xmark & 20.27 & 17.15 & 18.71 \\ 
        Fisher-Merging \cite{Fisher-merging} & \cmark & 21.64 & 24 & 22.82 \textcolor{red}{(+ 4.11)} \\ 
        \hline
        MagMax \cite{MAGMAX} & \xmark & 65.05 & 28.09 & 46.57  \\ 
        MagMax \cite{MAGMAX} & \cmark & 66.79 & 28.28 & 47.54 \textcolor{red}{(+ 0.97)} \\ 
        \hline
        Weight-Averaging \cite{TA_Paper_0} & \xmark & 66.42 & 31.42 & 48.92 \\ 
        Weight-Averaging \cite{TA_Paper_0} & \cmark & 76.12 & 37.53 & 56.83 \textcolor{red}{(+ 7.91)} \\ 
        \hline
        EMR-Merging \cite{EMR-Merging} & \xmark & 67.66 & 34.4 & 51.03 \\ 
        EMR-Merging \cite{EMR-Merging} & \cmark & 68.03 & 36.46 & 52.25 \textcolor{red}{(+ 1.22)} \\
        \hline
        \rowcolor{lightblue} \textbf{DuET (Ours)} & \xmark & \underline{87.06} & \textit{\underline{37.75}} & \textit{\underline{62.41}} \\ 
        \rowcolor{lightblue} \textbf{DuET (Ours)} & \cmark & \textbf{87.44} & \textbf{44.54} & \textbf{65.99} \textcolor{red}{(+ 3.58)} \\ 
    \bottomrule
    \end{tabular}
    }
\end{table}

\begin{figure}[ht]
\begin{center}
\centerline{\includegraphics[width=\columnwidth]{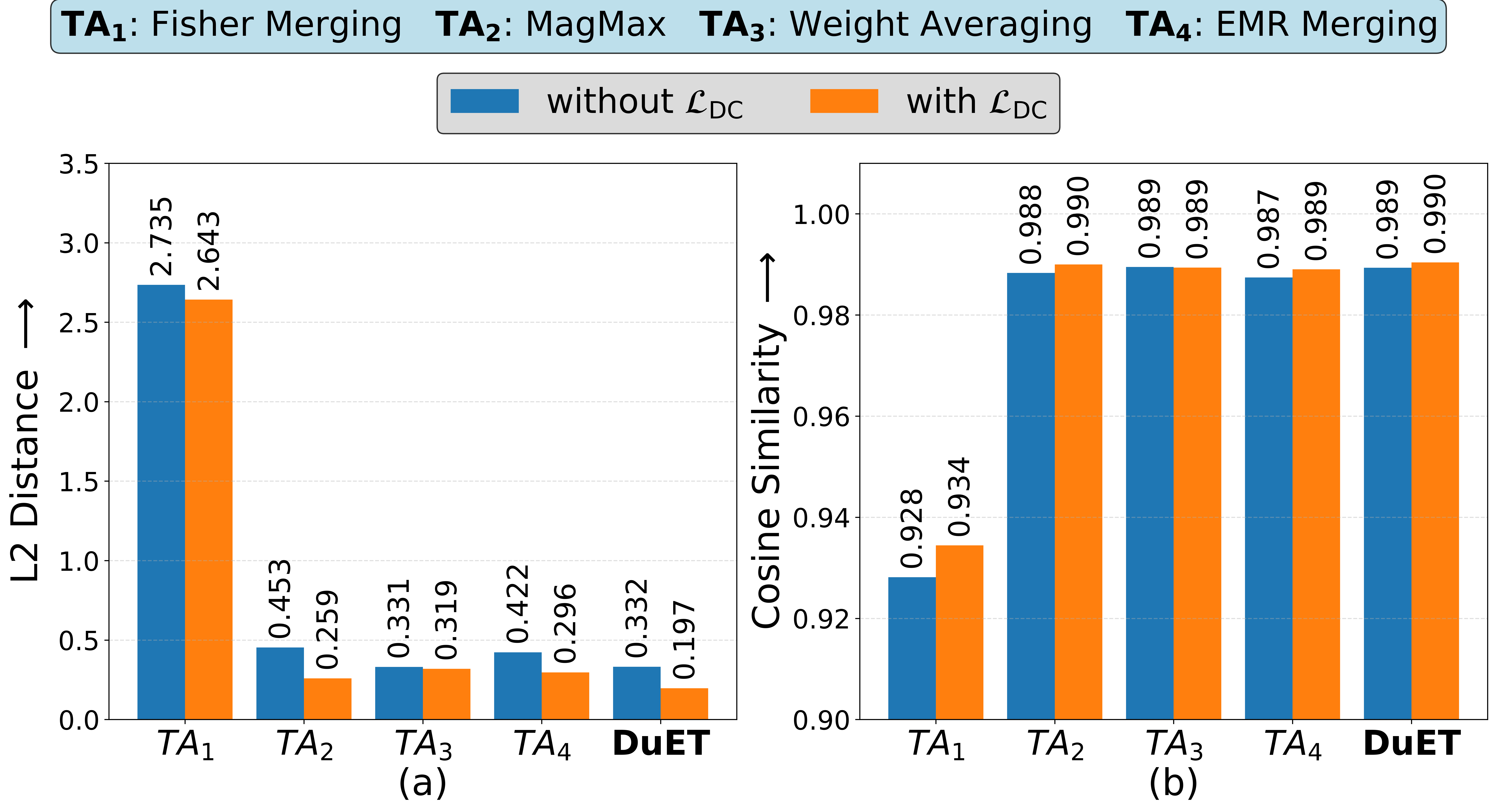}}
\vspace{-2mm}
\caption{\textbf{Impact of \(\mathcal{L}_{DC}\) in (a) reducing L2 Distance and (b) improving cosine similarity.} These results are obtained on the VOC [1:10] → Clipart [11:20] experiment using YOLO11n \cite{YOLO11} as the base detector with Incremental Head and Sequential Fine-tuning.}
\label{fig:l2_cosine}
\vspace{-10mm}
\end{center}
\end{figure}
\paragraph{Impact of \(\mathcal{L}_{DC}\).}
Continuing the ablations from the main paper (Section 6), in this section, we further investigate the role of \(\mathcal{L}_{DC}\). In Table~\ref{table:ta_ablation_detailed}, we compare the performance of different model-merging algorithms, with and without \(\mathcal{L}_{DC}\). The results show that \(\mathcal{L}_{DC}\) consistently improves the RAI across all methods, with an average RAI improvement of \textbf{+3.56\%} among all merging methods, with DuET achieving the best performance. Moreover, the bar charts in Figure~\ref{fig:l2_cosine} compare the L2 distance and cosine similarity between the merged model weights with both old and current model weights across different model-merging algorithms. The results show that \(\mathcal{L}_{DC}\) significantly reduces L2 distance by \textbf{43.46\%} averaged across all methods, with DuET achieving the lowest values. Lower L2 distance suggests that after incorporating \(\mathcal{L}_{DC}\), the merged model lies closer to the original models, ensuring effective knowledge integration from both. Similarly, incorporation of \(\mathcal{L}_{DC}\) consistently improves cosine similarity across all methods by \textbf{0.23\%} average, with DuET achieving the highest values. Higher cosine similarity suggests that \(\mathcal{L}_{DC}\) helps the merged model better align with the original models. 
\begin{table}[!ht]
    \centering
    \caption{Ablation studies of different loss components augmented with detector loss (\(\mathcal{L}_{\text{Detector}}\)).}
    \label{table:loss_ablations}
    \vspace{-2mm}
    \resizebox{\columnwidth}{!}{%
    \begin{tabular}{l|ccc}
    \toprule
        \textbf{Loss Component} &
        \textbf{Avg RI (\%)} &
        \textbf{Avg GI (\%)} &
        \textbf{RAI (\%)} \\
        \hline
        \(\mathcal{L}_{Detector} + \mathcal{L}_{\text{Distill}}\) & 72.64 & 33.74 & 53.19 \\
        \(\mathcal{L}_{Detector} + \mathcal{L}_{\text{Distill}}^{*}\) & 87.06 & 37.75 & 62.41 \\
        \(\mathcal{L}_{Detector} + \mathcal{L}_{\text{Distill}}^{*} + \mathcal{L}_{\text{DC}}\) & \textbf{87.44} & \textbf{44.54} & \textbf{65.99} \\
    \bottomrule
    \end{tabular}
    }
\end{table}
\vspace{-4mm}
\paragraph{Impact of \(\mathcal{L}_{Distill}^{*}\).}
Table~\ref{table:loss_ablations} presents the ablation studies of different loss components augmented with detector loss (\(\mathcal{L}_{\text{Detector}}\)). We observe that the inclusion of \(\mathcal{L}_{\text{Distill}}^{*}\) instead of \(\mathcal{L}_{\text{Distill}}\) significantly improves all metrics, with a \textbf{+14.42\%} increase in Avg RI, \textbf{+4.01\%} increase in Avg GI, and \textbf{+9.22\%} increase in RAI. The addition of \(\mathcal{L}_{\text{DC}}\) brings in additional improvements, leading to the best performance across all metrics.
\subsection{Sensitivity Analysis for key hyper-parameters}
\begin{figure*}[ht]
    \centering
    \begin{subfigure}[b]{0.48\textwidth}
        \centering
        \includegraphics[width=\linewidth]{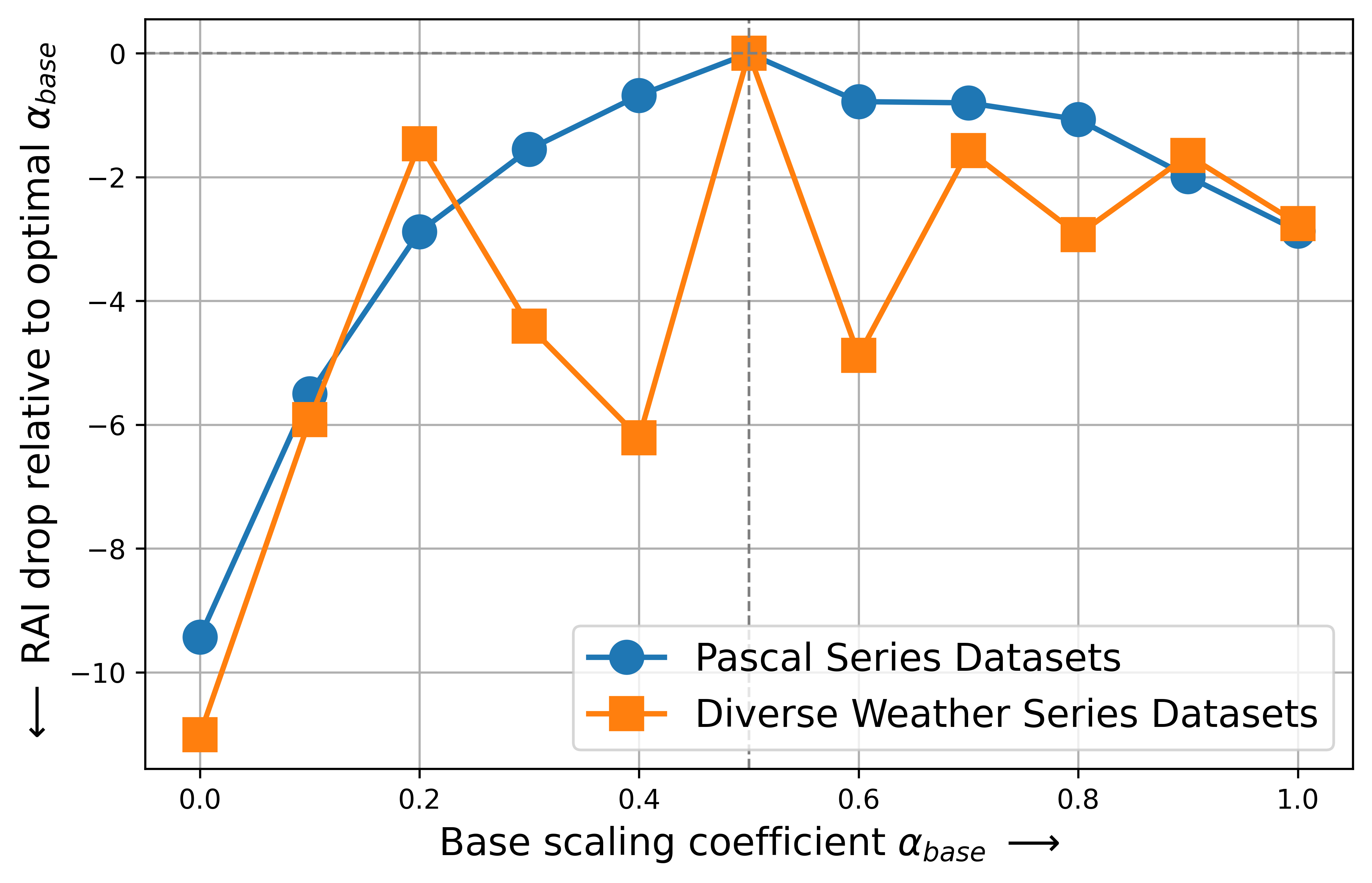}
        \caption{}
        \label{fig:alpha_base}
    \end{subfigure}
    \hfill
    \begin{subfigure}[b]{0.48\textwidth}
        \centering
        \includegraphics[width=\linewidth]{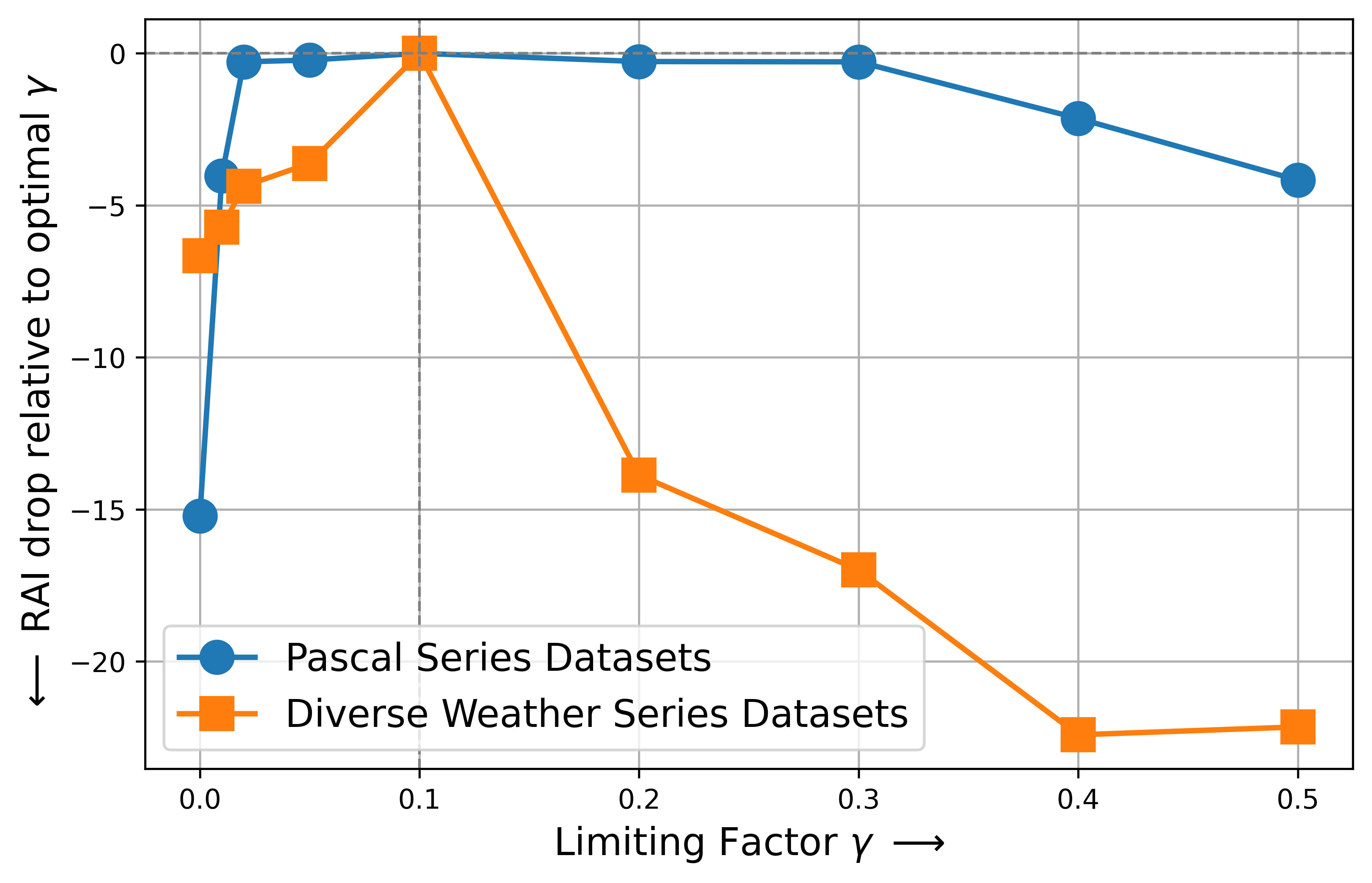}
        \caption{}
        \label{fig:limiting_factor}
    \end{subfigure}
    \\
    \begin{subfigure}[b]{0.48\textwidth}
        \centering
        \includegraphics[width=\linewidth]{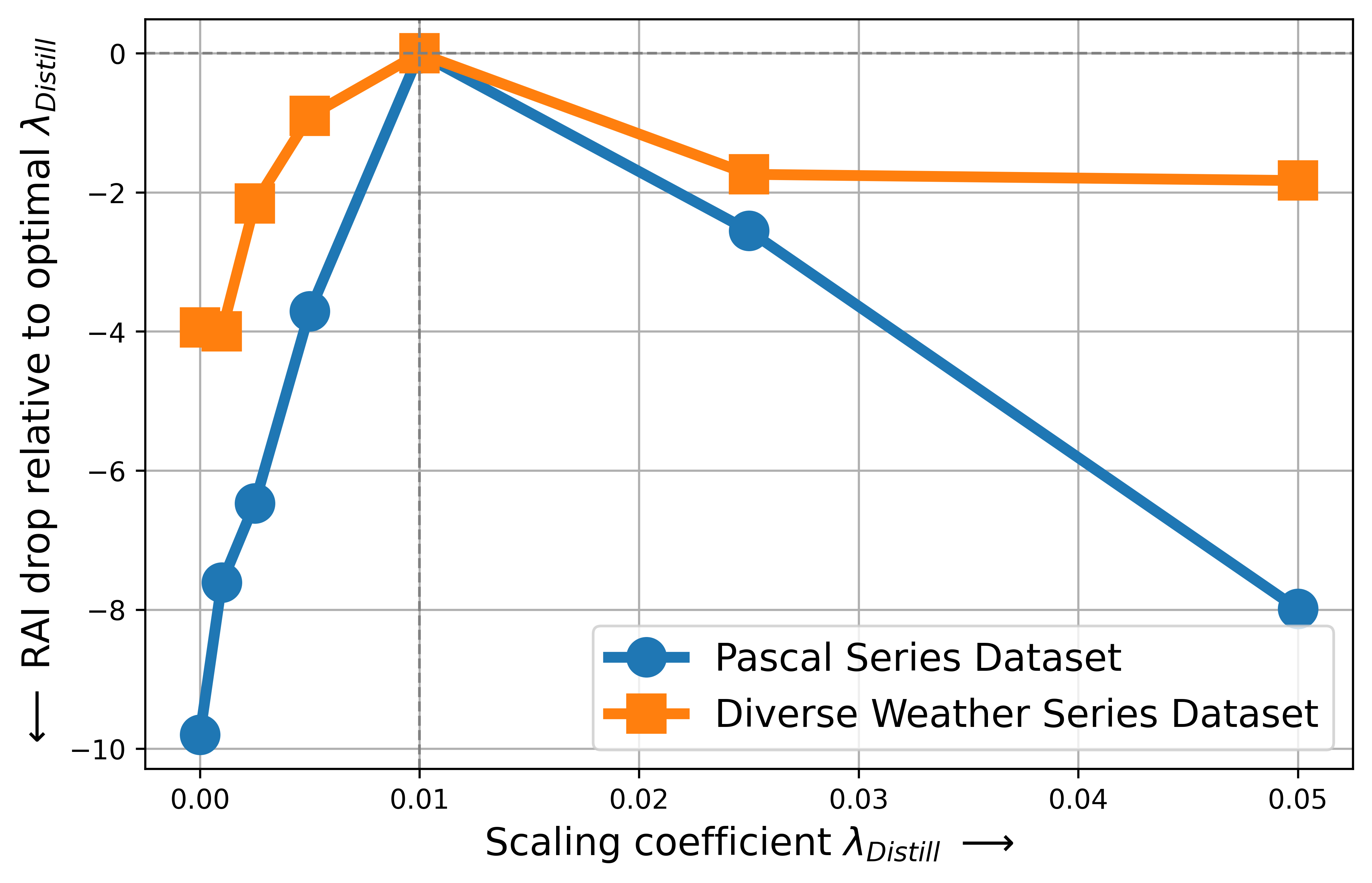}
        \caption{}
        \label{fig:lambda_distill}
    \end{subfigure}
    \hfill
    \begin{subfigure}[b]{0.48\textwidth}
        \centering
        \includegraphics[width=\linewidth]{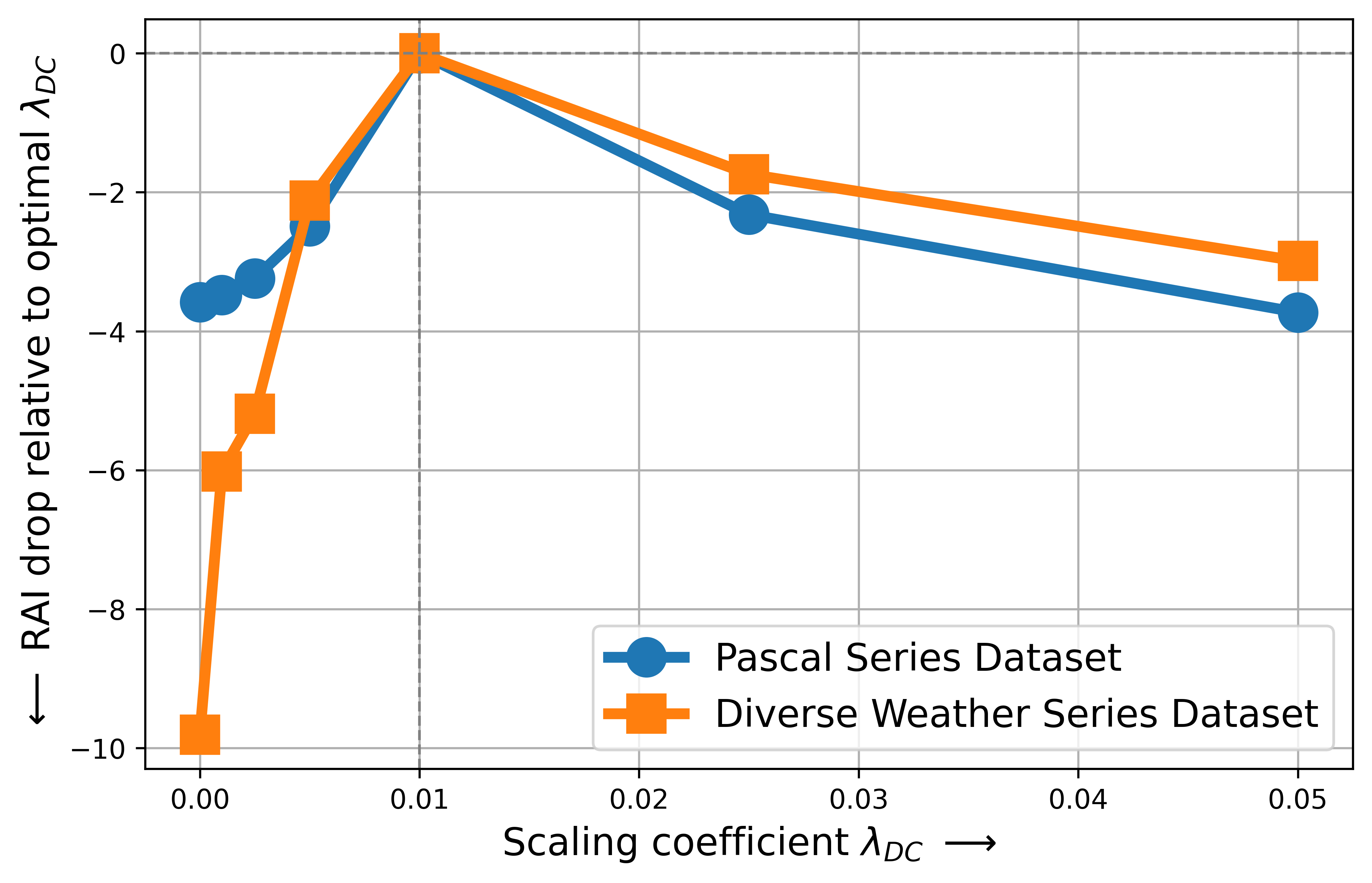}
        \caption{}
        \label{fig:lambda_dc}
    \end{subfigure}
    \caption{\textbf{Sensitivity analysis of DuET approach to key hyperparameters.} 
    \subref{fig:alpha_base} shows the effect of varying base scaling coefficient \(\alpha_{base}\). 
    \subref{fig:limiting_factor} illustrates the impact of the limiting factor \(\gamma\) on RAI.
    \subref{fig:lambda_distill} and \subref{fig:lambda_dc} depicts the effect of varying scaling coefficients \(\lambda_{Distill}\) and \(\lambda_{DC}\) on RAI respectively.}
    \label{fig:hyperparam_sensitivity}
\end{figure*}
Figure~\ref{fig:hyperparam_sensitivity} shows the sensitivity analysis for key hyperparameters used in the DuET approach. Base scaling coefficient \(\alpha_{base}\) (Figure~\ref{fig:alpha_base}) effectively controls the contributions from the old (prior task) model and current model; hence, a value of \textbf{0.5} ensures a balanced trade-off between past knowledge retention and new adaptation, while extreme values (\(\alpha_{base} < 0.3\) or \(\alpha_{base} > 0.7\)) significantly degrade RAI. The limiting factor \(\gamma\) (Figure~\ref{fig:limiting_factor}) impacts task-merging, with \(\gamma = 0.1\) giving optimal results for both \emph{Pascal Series} and \emph{Diverse Weather Series} datasets. The scaling coefficients \(\lambda_{Distill}\) and \(\lambda_{DC}\) (Figures~\ref{fig:lambda_distill} and \ref{fig:lambda_dc}) control the impact of Distillation and Directional Consistency losses, respectively. We observe that, for both of them, a value of \textbf{0.01} gives the best results and effectively helps in mitigating catastrophic forgetting by improving retention while preventing sign conflicts; deviations from these values lead to reduced adaptability and degraded performance on both series of datasets. 
\begin{table}[!ht]
    \centering
    \caption{Influence of random domain and class permutations across three incremental tasks on the \textit{Diverse Weather Series} dataset.}
    \label{tab:random_order}
    \vspace{-2mm}
    \renewcommand{\arraystretch}{1.2}
    \resizebox{\columnwidth}{!}{%
    \begin{tabular}{ccc|ccc}    
        \toprule
        \textbf{$\mathcal{T}_1$} & \textbf{$\mathcal{T}_2$} & \textbf{$\mathcal{T}_3$} & \textbf{Avg RI} & \textbf{Avg GI} & \textbf{RAI} \\
        \hline
        Night Sunny [5:7] & Daytime Sunny [1:2] & Daytime Foggy [3:4] & 83.49 & 51.01 & 67.25 \\
        Night Sunny [3:4] & Daytime Sunny [5:7] & Daytime Foggy [1:2] & 80.39 & 51.76 & 66.08 \\
        Night Sunny [1:2] & Daytime Sunny [3:4] & Daytime Foggy [5:7] & 88.57 & 41.92 & 65.25 \\
        Daytime Foggy [1:2] & Night Sunny [3:4] & Daytime Sunny [5:7] & 78.34 & 50.54 & 64.44 \\
        Daytime Sunny [1:2] & Daytime Foggy [3:4] & Night Sunny [5:7] & 88.33 & 35.97 & 62.15 \\
        \hline
        \multicolumn{3}{c|}{\textbf{Standard Deviation}} & \textbf{4.12} & \textbf{6.26} & \textbf{1.72} \\
        \bottomrule
    \end{tabular}
    }
    \vspace{-2mm}
\end{table}

\subsection{Influence of random class-domain order}
In real-world incremental learning scenarios, the sequence in which new classes and domains are introduced can influence knowledge retention and generalisation. To check the sensitivity of the proposed DuET approach to such variations, we conducted experiments with shuffled class orders while keeping the same domain progression (top three rows in Table~\ref{tab:random_order}) and shuffled domain orders while keeping the same class sets (bottom three rows). We observe that Avg RI remains consistently high across all permutations (\( > 78\%\)), and there are minor variations in RAI and Avg RI with standard deviations of \textbf{1.72} and \textbf{4.12}, respectively. However, the slight variation in Avg GI, with a standard deviation of \textbf{6.26}, stems from domain shifts affecting generalisation. This indicates that the proposed DuET approach maintains a consistent performance irrespective of the randomness in class-domain orders.
\subsection{Complexity Analysis}
Table~\ref{tab:complexity_analysis} presents a detailed complexity analysis of various methods evaluated in the multi-phase experiment: \textit{Watercolour [1:3] $\rightarrow$ Comic [4:6] $\rightarrow$ Clipart [7:13] $\rightarrow$ VOC [14:20]}. Table~\ref{tab:complexity_analysis} compares computational complexity in terms of GFLOPs, trainable parameters (in millions), average inference speed (in milliseconds), and average memory footprint (in gigabytes) across all incremental tasks. While training time (in hours) as evaluated on a single NVIDIA A100-PCIE-40GB on \textit{Daytime Sunny [1:4] $\rightarrow$ Night Rainy [5:7]} experiment is reported in Table~\ref{E7_exps_table}. The results demonstrate that DuET retains the real-time detection capabilities of YOLO11n \cite{YOLO11}, effectively transforming it into a robust real-time incremental object detector with only a minimal increase in memory footprint (\textbf{0.244 GB}) compared to its unaltered counterpart, Sequential FT (\textbf{0.235 GB}). Furthermore, since the proposed DuET approach does not modify the base detector architecture, it preserves the same GFLOPs and the number of trainable parameters.

\begin{figure}[h]
\begin{center}
\centerline{\includegraphics[width=1.05\columnwidth, height=4.8cm]{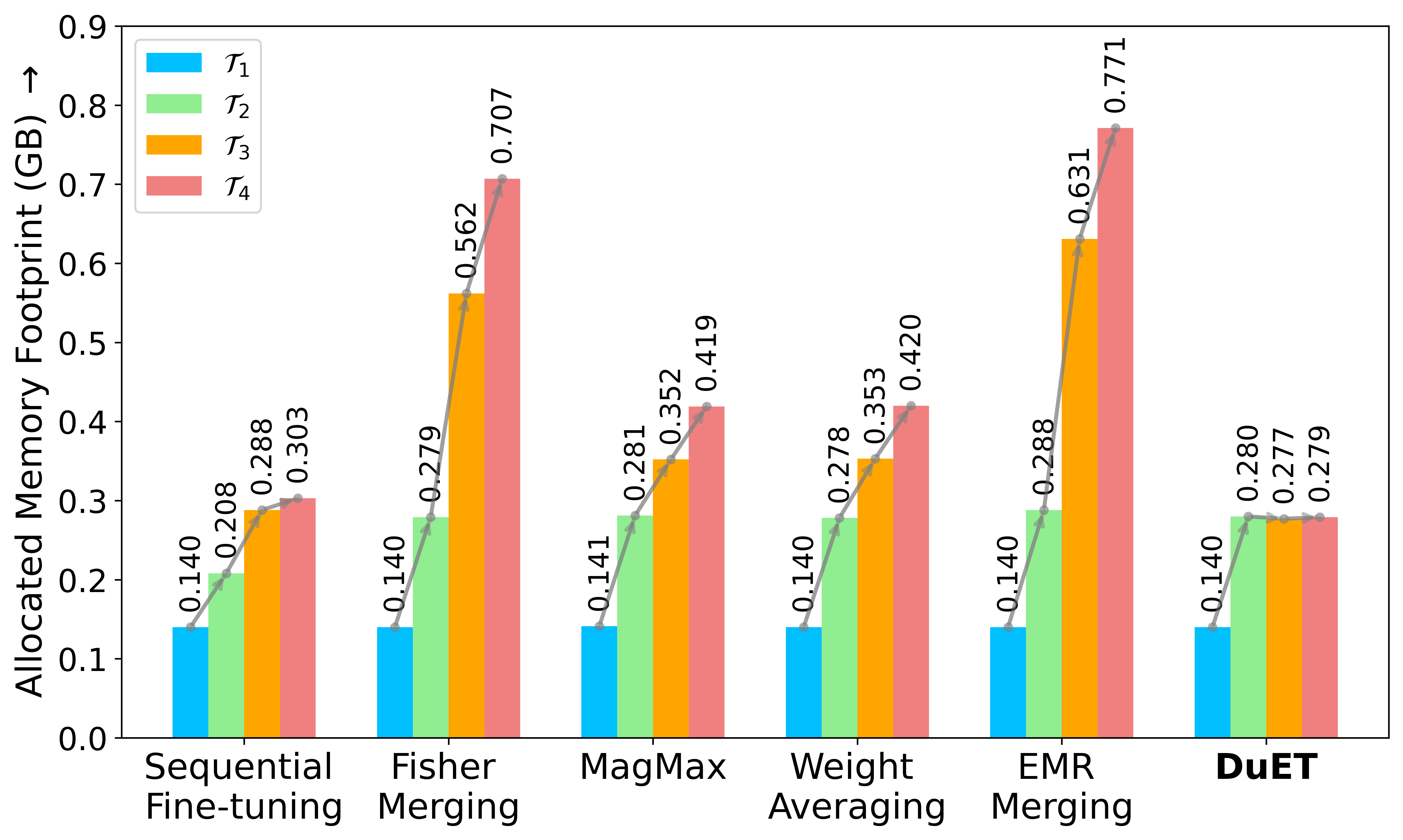}}
\vspace{-0.5em}
\caption{Comparison of allocated memory footprint (in GB) for various model-merging approaches on the multi-phase experiment with four tasks: Watercolor [1:3] ($\mathcal{T}_1$) $\rightarrow$ Comic [4:6] ($\mathcal{T}_2$) $\rightarrow$ Clipart [7:13] ($\mathcal{T}_3$) $\rightarrow$ VOC [14:20] ($\mathcal{T}_4)$.}
\label{fig:TA_mem_foo}
\vspace{-8mm}
\end{center}
\end{figure}

In contrast to other model-merging algorithms \cite{EMR-Merging, TA_Paper_0, MAGMAX, Fisher-merging}, which require storing task vectors—and consequently, model weights—for every task, our approach is designed to be more efficient and scalable. DuET maintains only two shared task vectors at any given task: $\tau_{\text{old}}$ (derived from the previous phase's model weights) and $\tau_{\text{curr}}$ (derived from the current phase's model weights), along with the pre-trained model weights. This design utilizes the fact that knowledge from earlier tasks, {$\mathcal{T}_{1}, \mathcal{T}_{2}, \dots, \mathcal{T}_{t-2}$}, is already encapsulated within the previous phase's weights, $\theta{s_{t-1}}$. Consequently, DuET avoids the overhead of maintaining a complete history of task vectors, resulting in a consistent memory footprint across all incremental tasks ($\mathcal{T}_{t}, t \geq 2$), while in case of other TA approaches, the memory footprint grows linearly with the number of tasks (Figure~\ref{fig:TA_mem_foo}).
\section{Implementation Details}
\label{implementation_details}
Our implementation is primarily based on the Ultralytics  framework\footnotemark (v8.3.9), with YOLO11n \cite{YOLO11} primarily serving as the base detector, also extending to other variants of YOLO11 and RT-DETR \cite{RT-DETR}. Following the default configuration provided by Ultralytics, we used AdamW \cite{AdamW} optimiser with auto lr find and OneCycleLR scheduler, keeping a batch size of 64. For every task, we trained the detector for 100 epochs, keeping five warm-up epochs with a higher initial learning rate by a factor of 10. For the base task (\(t = 1\)), we use the default weight decay of 0.0005, while for incremental tasks (\(t \geq 2\)), we slightly increase it to 0.001 to prevent overfitting to new tasks and help prevent catastrophic forgetting. The same protocol is used while preparing other baselines for a fair comparison. Moreover, unlike LDB \cite{LDB} and CL-DETR \cite{CL-DETR}, we keep all layers trainable during incremental training to ensure that shared task vectors effectively capture the shift in shared knowledge across incremental tasks.
\footnotetext{https://github.com/ultralytics/ultralytics}
\section{Comprehensive Results}
\subsection{Detailed analysis of Quantitative Results:}
Tables~\ref{E1_exps_table} to \ref{E7_exps_table} present the comprehensive results of various methods on different DuIOD experiments across multiple base detectors. We conducted a total of seven DuIOD experiments—five two-phase and two multi-phase experiments—three from the \emph{Pascal Series} and the remaining four from the \emph{Diverse Weather Series} datasets.

We provide detailed results for the five two-phase experiments: \textit{VOC[1:10] $\rightarrow$ Clipart[11:20]} (Table~\ref{E1_exps_table}), \textit{Clipart [1:10] $\rightarrow$ VOC [11:20]} (Table~\ref{E2_exps_table}), \textit{Daytime Sunny [1:4] $\rightarrow$ Night Sunny [5:7]} (Table~\ref{E3_exps_table}), \textit{Night Sunny [1:4] $\rightarrow$ Daytime Sunny [5:7]} (Table~\ref{E4_exps_table}) and \textit{Daytime Sunny [1:4] $\rightarrow$ Night Rainy [5:7]} (Table~\ref{E7_exps_table}). Meanwhile, the results for the two multi-phase experiments—\textit{Watercolor [1:3] $\rightarrow$ Comic [4:6] $\rightarrow$ Clipart [7:13] $\rightarrow$ VOC [14:20]} and \textit{Night Sunny [1:2] $\rightarrow$ Daytime Sunny [3:4] $\rightarrow$ Daytime Foggy [5:7]}—are presented in Tables~\ref{E5_exps_table} and \ref{E6_exps_table}, respectively. 

We evaluate DuET across all DuIOD experiments using five detection backbones: DeformableDETR \cite{DeformableDETR}, YOLO11n \& YOLO11x \cite{YOLO11}, and RTDETR-l \& RTDETR-x \cite{RT-DETR}. Our results show that DuET consistently outperforms the baselines in nearly all DuIOD experiments, demonstrating its effectiveness in addressing the DuIOD task.
Notably, DuET outperforms both CL-DETR \cite{CL-DETR} \& LDB \cite{LDB} on their respective backbones (Deformable DETR \cite{DeformableDETR} \& ViTDet \cite{VitDet}) with a \textbf{+5.97\%} and \textbf{+11.98\%} RAI gain, preserving \textbf{80.3\%} vs. \textbf{66.85\%} and \textbf{50.99\%} vs. \textbf{41.74\%} Avg RI respectively (refer Tables \ref{E1_exps_table} to \ref{E7_exps_table}). This indicates that \textbf{gains are method-specific and not backbone dependent}.
\subsection{Qualitative Visualizations}
\begin{figure*}[ht]
\begin{center}
\centerline{\includegraphics[width=\textwidth]{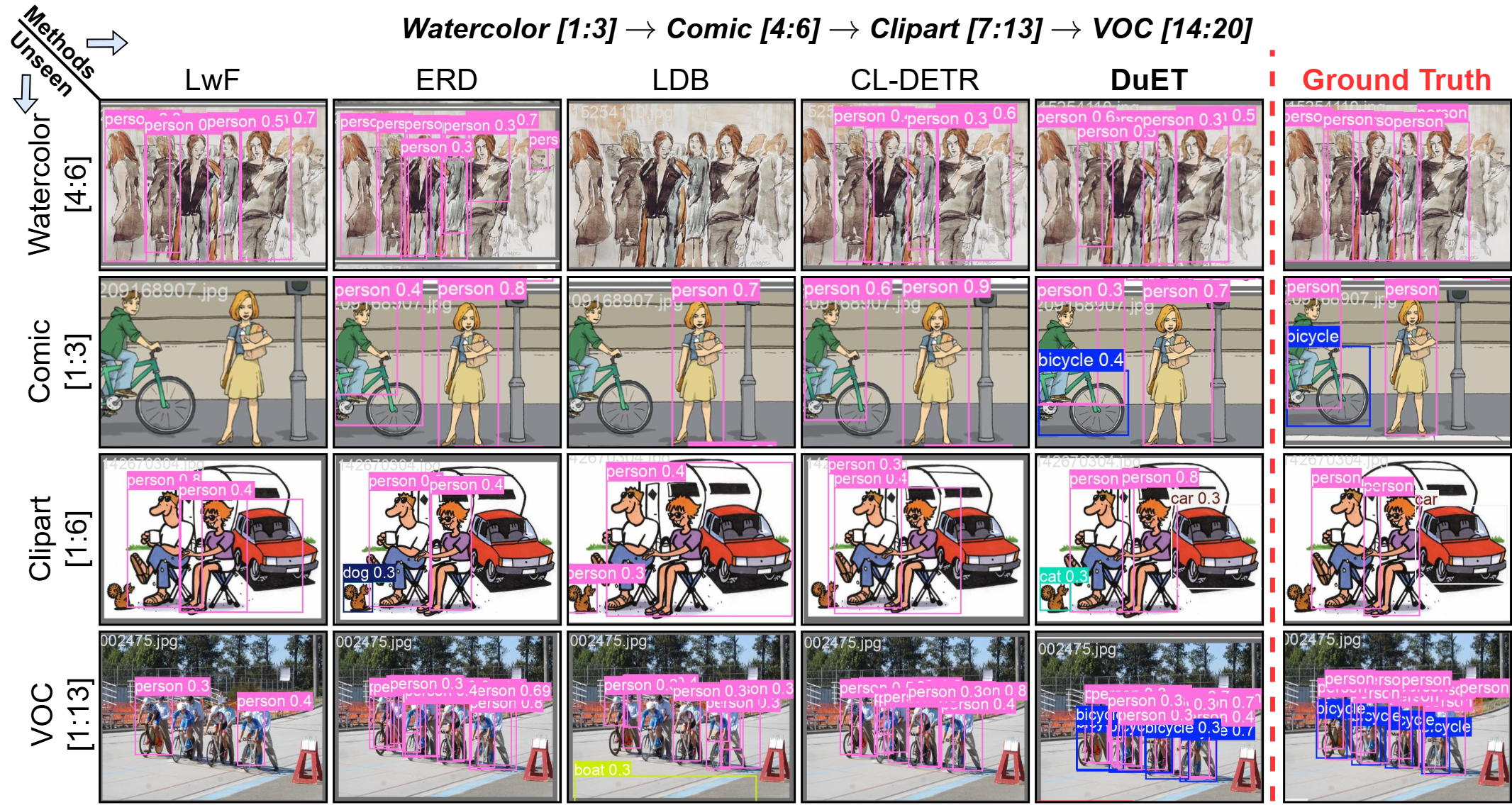}}
\vspace{-0.5em}
\caption{
Qualitative comparisons on \emph{Pascal Series} multi-phase experiment: Watercolour [1:3] $\rightarrow$ Comic [4:6] $\rightarrow$ Clipart [7:13] $\rightarrow$ VOC [14:20] for different methods on the DuIOD task. The four rows display detection results on unseen classes: Watercolour [4:6], Comic [1:3], Clipart [1:6], and VOC [1:13] on the final task \(\mathcal{T}_4\). (zoomed in for best view).}
\label{fig:qual_results_E3}
\vspace{-6mm}
\end{center}
\end{figure*}

\begin{figure*}[]
\begin{center}
\centerline{\includegraphics[width=\textwidth]{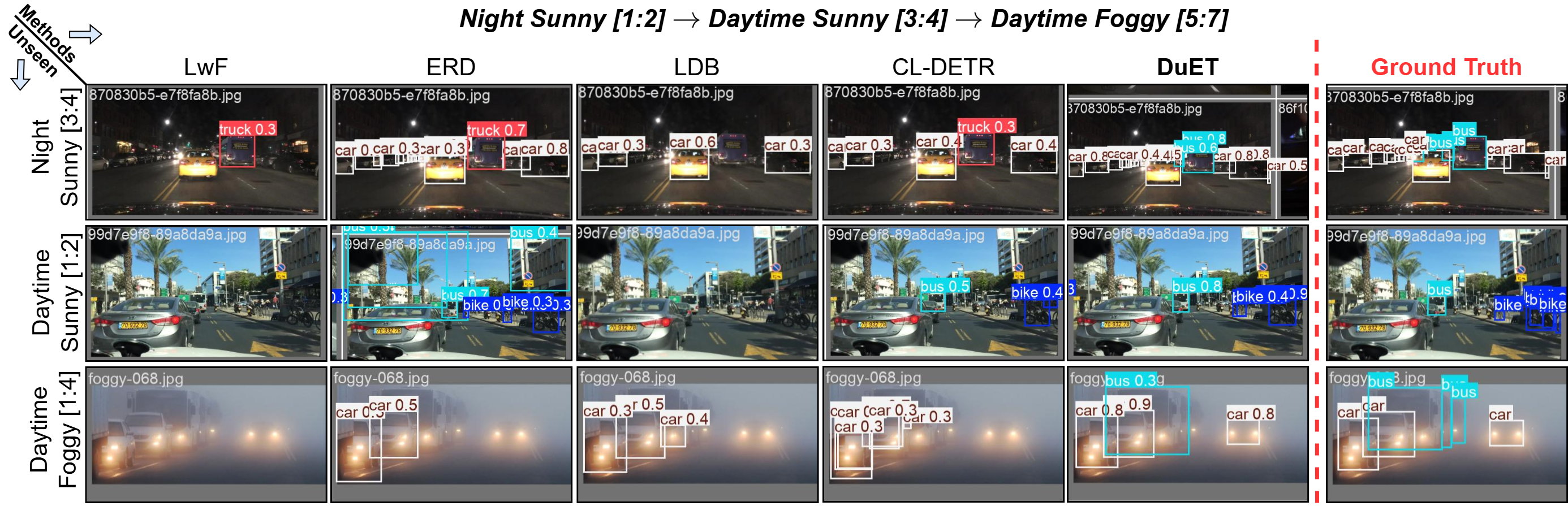}}
\vspace{-0.5em}
\caption{
Qualitative comparisons on \emph{Diverse Series} multi-phase experiment: Night Sunny [1:2] $\rightarrow$ Daytime Sunny [3:4] $\rightarrow$ Daytime Foggy [5:7] for different methods on the DuIOD task. The four rows display detection results on unseen classes: Night Sunny [3:4], Daytime Sunny [1:2], and Daytime Foggy [1:4] on the final task \(\mathcal{T}_3\). (zoomed-in for best view).}
\label{fig:qual_results_E6}
\vspace{-12mm}
\end{center}
\end{figure*}

In Figure~\ref{fig:qual_results_E3}, we present qualitative visualisations for various methods on the task sequence: \textit{Watercolour [1:3] $\rightarrow$ Comic [4:6] $\rightarrow$ Clipart [7:13] $\rightarrow$ VOC [14:20]}. The detection results are shown for unseen classes: Watercolour [4:6], Comic [1:3], Clipart [1:6], and VOC [1:13] in the final task, \(\mathcal{T}_4\). Our observations indicate that DuET consistently outperforms other methods by accurately detecting most objects across different domains and classes. Notably, in the second row (Comic [1:3]), the \texttt{bicycle} class, which was learned in \(\mathcal{T}_1\) (Watercolor [1:3]), and the \texttt{person} class, introduced in \(\mathcal{T}_2\) (Comic [4:6]), are examined. We observe that only DuET successfully retains the knowledge from \(\mathcal{T}_1\) and correctly detects the \texttt{bicycle} class in the unseen Comic [1:3] domain. A similar trend is observed in the fourth row (VOC [1:13]). Additionally, in the third row, the \texttt{car} class, introduced in \(\mathcal{T}_1\), is not detected by other methods in the unseen Clipart [1:6] domain, whereas DuET consistently identifies it. These qualitative results further emphasise DuET's effectiveness in adapting to unseen classes across different domains in the DuIOD task. A similar trend is observed in the \emph{Diverse Weather Series} (Figure~\ref{fig:qual_results_E6}), where DuET consistently outperforms other methods by accurately detecting most objects across various domains and classes.
\begin{table}[h]
    \centering
    \caption{\textbf{Dataset Statistics:} Class-wise distribution across different domains in \emph{Pascal Series} datasets.}
    \label{tab:ds_pascal_series}
    \vspace{-2mm}
    \renewcommand{\arraystretch}{1.2}
    \resizebox{\columnwidth}{!}{%
    \begin{tabular}{c c | c c c c}
    \toprule
        \textbf{Class ID} & \textbf{Class Name} & \textbf{Watercolor \cite{PASCAL_Series}} & \textbf{Comic \cite{PASCAL_Series}} & \textbf{Clipart \cite{PASCAL_Series}} & \textbf{VOC \cite{PASCAL_VOC}} \\
        \hline
        1  & bicycle      & \cmark & \cmark & \cmark & \cmark \\
        2  & bird        & \cmark & \cmark & \cmark & \cmark \\
        3  & car         & \cmark & \cmark & \cmark & \cmark \\
        4  & cat         & \cmark & \cmark & \cmark & \cmark \\
        5  & dog         & \cmark & \cmark & \cmark & \cmark \\
        6  & person      & \cmark & \cmark & \cmark & \cmark \\
        7  & aeroplane   &        &        & \cmark & \cmark \\
        8  & boat        &        &        & \cmark & \cmark \\
        9  & bottle      &        &        & \cmark & \cmark \\
        10 & bus         &        &        & \cmark & \cmark \\
        11 & chair       &        &        & \cmark & \cmark \\
        12 & cow         &        &        & \cmark & \cmark \\
        13 & diningtable &        &        & \cmark & \cmark \\
        14 & horse       &        &        & \cmark & \cmark \\
        15 & motorbike   &        &        & \cmark & \cmark \\
        16 & pottedplant &        &        & \cmark & \cmark \\
        17 & sheep       &        &        & \cmark & \cmark \\
        18 & sofa        &        &        & \cmark & \cmark \\
        19 & train       &        &        & \cmark & \cmark \\
        20 & tvmonitor   &        &        & \cmark & \cmark \\
        \hline
        \multicolumn{2}{c|}{\textbf{Total Classes}} & 6 & 6 & 20 & 20 \\
        \hline
        \multicolumn{2}{c|}{\textbf{Train Images}} & 1000 & 1000 & 500 & 16551 \\
        \multicolumn{2}{c|}{\textbf{Val Images}} & 1000 & 1000 & 500 & 4952 \\
    \bottomrule
    \end{tabular}
    }
    \vspace{-2mm}
\end{table}
\begin{table}[h]
    \centering
    \caption{\textbf{Dataset Statistics:} Class-wise distribution across different weather conditions in \emph{Diverse Weather Series} datasets.}
    \label{tab:ds_div_weather_series}
    \vspace{-2mm}
    \renewcommand{\arraystretch}{1.2}
    \resizebox{\columnwidth}{!}{%
    \begin{tabular}{c c | c c c}
    \toprule
        \textbf{Class ID} & \textbf{Class Name} & \textbf{Daytime Sunny \cite{BDD100k}} & \textbf{Night Sunny \cite{BDD100k}} & \textbf{Daytime Foggy \cite{FoggyCityscapes, AdverseWeather}} \\
        \hline
        1  & bike       & \cmark & \cmark & \cmark \\
        2  & bus        & \cmark & \cmark & \cmark \\
        3  & car        & \cmark & \cmark & \cmark \\
        4  & motor      & \cmark & \cmark & \cmark \\
        5  & person     & \cmark & \cmark & \cmark \\
        6  & rider      & \cmark & \cmark & \cmark \\
        7  & truck      & \cmark & \cmark & \cmark \\
        \hline
        \multicolumn{2}{c|}{\textbf{Total Classes}} & 7 & 7 & 7 \\
        \hline
        \multicolumn{2}{c|}{\textbf{Train Images}} & 19317 & 25868 & 1829 \\
        \multicolumn{2}{c|}{\textbf{Val Images}} & 8289 & 7756 & 688 \\
    \bottomrule
    \end{tabular}
    }
    \vspace{-6mm}
\end{table}

\section{Background Shift}
\label{Background_Shift}
Background shift is a major issue in IOD scenarios \cite{ABR_IOD, continual_survey}, where previously learned object categories, if unannotated in subsequent tasks, are treated as background. DuET mitigates this shift by explicitly decomposing model parameters into shared and task-specific components, and then merging these through TA with dynamic, layer-wise retention and adaptation weights (see Section 3.4). This strategy ensures that parameters crucial for previously learned object representations remain stable, thus preventing catastrophic forgetting and minimising the likelihood of previously learned objects being erroneously classified as background when they become unlabeled in subsequent tasks.

\section{Dataset Statistics}
Following prior works \cite{LDB, kiran2022incremental, DivAlign, Single-DGOD}, we evaluate the DuET approach on two dataset series: the \emph{Pascal Series} and the \emph{Diverse Weather Series}, which cover diverse environmental conditions and domain variations, respectively. Table~\ref{tab:ds_pascal_series} presents the class-wise distribution, capturing cross-domain variations across four different domains: Watercolour, Comic, Clipart, and VOC. Following \cite{kiran2022incremental, liu2020multi}, we combined the PASCAL VOC 2007 and 2012 \cite{PASCAL_VOC} datasets to form the VOC domain, while the Watercolour, Comic, and Clipart domains were taken from \cite{PASCAL_Series}. Watercolour and Comic domains consist of six object categories, forming a subset of the 20 object categories present in Clipart and VOC. We used class splits of \(10+10\) and \(3+3+7+7\) with Class IDs as mentioned in Table~\ref{tab:ds_pascal_series} to conduct two-phase and multi-phase DuIOD experiments on the \emph{Pascal Series} datasets, respectively. Similarly, Table~\ref{tab:ds_div_weather_series} presents the class-wise distribution across three different weather conditions: Daytime Sunny, Night Sunny, and Daytime Foggy. Following \cite{kiran2022incremental, LDB}, the datasets are taken from BDD100k \cite{BDD100k}, Foggy Cityscapes \cite{FoggyCityscapes}, and Adverse Weather \cite{AdverseWeather}. Since each domain contains a common set of seven classes, we used class splits of \(4+3\) and \(2+2+3\), with Class IDs sorted in alphabetical order (as shown in Table~\ref{tab:ds_div_weather_series}), to perform two-phase and multi-phase DuIOD experiments on the \emph{Diverse Weather Series} datasets, respectively.
\section{Limitations and future works}
Since DuET is a task vector-based model-merging approach, it inherits the limitations of existing task vector-based methods, and hence it cannot be generalised to models trained from scratch and requires access to pre-trained object detectors to calculate shared task vectors at each incremental task. This is a common limitation of task vector-based methods.
Moreover, DuET merges shared task vectors through a weighted linear interpolation mechanism, which may be suboptimal for highly heterogeneous class shifts or extreme domain variations across incremental tasks. Future work could explore more sophisticated non-linear merging approaches to better capture the shared knowledge across tasks.
\clearpage
\begin{table*}[!ht]
    \centering
    \caption{Results of various methods on VOC [1:10] $\rightarrow$ Clipart [11:20] with different base detectors.
    Among columns, best in \textbf{bold}, second best \textit{\underline{underlined}}.}
    \label{E1_exps_table}
    \vspace{-2mm}
    \resizebox{\textwidth}{!}{%
    \setlength{\tabcolsep}{1.1pt}
    \renewcommand{\arraystretch}{1.08}
    \begin{tabular}{cc||c|c|c|c|c||ccc}
    \toprule
        \multirow{3}{*}{\textbf{Method}} & \multirow{3}{*}{\textbf{Base Detector}} & 
        \multirow{3}{*}{\makecell{\textbf{T1} \\ \textbf{VOC } \\ \textbf{[1:10]}}} &
        \multicolumn{4}{c||}{\textbf{T2: Clipart [11:20]}} &
        \multirow{3}{*}{\makecell{\textbf{Avg RI} \\ \textbf{(\%)}}} & 
        \multirow{3}{*}{\makecell{\textbf{Avg GI} \\ \textbf{(\%)}}} & 
        \multirow{3}{*}{\makecell{\textbf{RAI} \\ \textbf{(\%)}}} \\
        \cline{4-7}
        & & & \makecell{\textbf{Old}} &  
        \makecell{\textbf{New}} &
        \multicolumn{2}{c||}{\textbf{Unseen}} \\
        \cline{4-7}
        & & & \makecell{\textbf{VOC [1:10]}} & 
        \makecell{\textbf{Clipart [11:20]}} & 
        \makecell{\textbf{Clipart [1:10]}} & 
        \makecell{\textbf{VOC [11:20]}} & & & \\
        \hline
        \hline
        LDB \cite{LDB} & ViTDet & 74.9\tiny{$\pm$0.7} & 50.10\tiny{$\pm$0.5} & 22.30\tiny{$\pm$0.8} & 8.90\tiny{$\pm$0.4} & 9.60\tiny{$\pm$0.6} & 66.89\tiny{$\pm$0.5} & 18.76\tiny{$\pm$0.3} & 42.83\tiny{$\pm$0.4} \\
        \rowcolor{lightblue} \textbf{DuET (Ours)} & ViTDet & 74.9\tiny{$\pm$0.2} & 54.20\tiny{$\pm$0.4} & 17.60\tiny{$\pm$0.2} & 17.90\tiny{$\pm$0.3} & 11.80\tiny{$\pm$0.4} & 72.36\tiny{$\pm$0.2} & 32.63\tiny{$\pm$0.3} & 52.50\tiny{$\pm$0.2} \\
        \hline
        CL-DETR \cite{CL-DETR} & Deformable DETR & 56.1\tiny{$\pm$0.6} & 38.29\tiny{$\pm$0.7} & 9.04\tiny{$\pm$0.5} & 9.22\tiny{$\pm$0.8} & 3.02\tiny{$\pm$0.4} & 68.29\tiny{$\pm$0.3} & 40.72\tiny{$\pm$0.4} & 54.51\tiny{$\pm$0.3} \\
        \rowcolor{lightblue} \textbf{DuET (Ours)} & Deformable DETR & 56.1\tiny{$\pm$0.6} & 42.32\tiny{$\pm$0.3} & 4.10\tiny{$\pm$0.2} & 15.63\tiny{$\pm$0.4} & 1.68\tiny{$\pm$0.4} & 75.48\tiny{$\pm$0.6} & 72.37\tiny{$\pm$0.3} & \textit{\underline{73.93}\tiny{$\pm$0.5}} \\
        \hline
        Sequential FT & RTDETR-l & 87.1\tiny{$\pm$0.6} & 0.00\tiny{$\pm$0.0} & 55.00\tiny{$\pm$0.7} & 0.00\tiny{$\pm$0.0} & 32.20\tiny{$\pm$0.5} & 0.00\tiny{$\pm$0.0} & 18.85\tiny{$\pm$0.6} & 9.43\tiny{$\pm$0.4} \\
        LwF \cite{LwF} & RTDETR-l & 87.1\tiny{$\pm$0.6} & 3.13\tiny{$\pm$0.2} & 25.90\tiny{$\pm$0.8} & 1.30\tiny{$\pm$0.3} & 18.50\tiny{$\pm$0.4} & 3.59\tiny{$\pm$0.7} & 12.30\tiny{$\pm$0.5} & 7.95\tiny{$\pm$0.6} \\
        ERD \cite{ERD} & RTDETR-l & 87.1\tiny{$\pm$0.6} & 1.74\tiny{$\pm$0.3} & 56.00\tiny{$\pm$0.8} & 1.42\tiny{$\pm$0.2} & 37.80\tiny{$\pm$0.5} & 2.00\tiny{$\pm$0.1} & 23.74\tiny{$\pm$0.6} & 12.87\tiny{$\pm$0.8} \\
        \rowcolor{lightblue} \textbf{DuET (Ours)} & RTDETR-l & 87.1\tiny{$\pm$0.6} & 46.10\tiny{$\pm$0.1} & 68.00\tiny{$\pm$0.2} & 28.10\tiny{$\pm$0.4} & 62.20\tiny{$\pm$0.4} & 52.93\tiny{$\pm$0.7} & 68.20\tiny{$\pm$0.5} & 60.57\tiny{$\pm$0.6} \\
        \hline
        Sequential FT & RTDETR-x & 89.2\tiny{$\pm$0.5} & 0.00\tiny{$\pm$0.0} & 56.52\tiny{$\pm$0.7} & 0.33\tiny{$\pm$0.2} & 30.43\tiny{$\pm$0.6} & 0.00\tiny{$\pm$0.0} & 17.66\tiny{$\pm$0.4} & 8.83\tiny{$\pm$0.3} \\
        LwF \cite{LwF} & RTDETR-x & 89.2\tiny{$\pm$0.5} & 20.40\tiny{$\pm$0.5} & 28.80\tiny{$\pm$0.3} & 17.00\tiny{$\pm$0.7} & 17.90\tiny{$\pm$0.4} & 22.87\tiny{$\pm$0.6} & 28.27\tiny{$\pm$0.5} & 25.57\tiny{$\pm$0.2} \\
        ERD \cite{ERD} & RTDETR-x & 89.2\tiny{$\pm$0.5} & 22.60\tiny{$\pm$0.7} & 55.20\tiny{$\pm$0.8} & 3.83\tiny{$\pm$0.3} & 32.80\tiny{$\pm$0.4} & 25.34\tiny{$\pm$0.5} & 22.73\tiny{$\pm$0.1} & 24.04\tiny{$\pm$0.8} \\
        \rowcolor{lightblue} \textbf{DuET (Ours)} & RTDETR-x & 89.2\tiny{$\pm$0.5} & 60.50\tiny{$\pm$0.5} & 26.80\tiny{$\pm$0.4} & 49.00\tiny{$\pm$0.3} & 22.50\tiny{$\pm$0.2} & 67.83\tiny{$\pm$0.4} & 64.93\tiny{$\pm$0.6} & 66.38\tiny{$\pm$0.7} \\
        \hline
        Sequential FT & YOLO11n & 80.4\tiny{$\pm$0.3} & 0.60\tiny{$\pm$0.1} & 36.70\tiny{$\pm$0.8} & 1.02\tiny{$\pm$0.3} & 17.40\tiny{$\pm$0.4} & 0.75\tiny{$\pm$0.2} & 12.86\tiny{$\pm$0.4} & 6.81\tiny{$\pm$0.3} \\
        LwF \cite{LwF} & YOLO11n & 80.4\tiny{$\pm$0.3} & 58.40\tiny{$\pm$0.8} & 3.96\tiny{$\pm$0.3} & 28.60\tiny{$\pm$0.7} & 5.00\tiny{$\pm$0.2} & 72.64\tiny{$\pm$0.3} & 33.74\tiny{$\pm$0.5} & 53.19\tiny{$\pm$0.4} \\
        ERD \cite{ERD} & YOLO11n & 80.4\tiny{$\pm$0.3} & 55.20\tiny{$\pm$0.4} & 20.60\tiny{$\pm$0.7} & 30.50\tiny{$\pm$0.5} & 16.70\tiny{$\pm$0.8} & 68.66\tiny{$\pm$0.4} & 43.68\tiny{$\pm$0.3} & 56.17\tiny{$\pm$0.6} \\
        \rowcolor{lightblue} \textbf{DuET (Ours)} & YOLO11n & 80.4\tiny{$\pm$0.3} & 70.30\tiny{$\pm$0.3} & 8.45\tiny{$\pm$0.3} & 33.80\tiny{$\pm$0.3} & 12.80\tiny{$\pm$0.3} & \textbf{87.44\tiny{$\pm$0.2}} & 44.54\tiny{$\pm$0.1} & 65.99\tiny{$\pm$0.3} \\
        \hline
        Sequential FT & YOLO11x & 88.4\tiny{$\pm$0.5} & 0.00\tiny{$\pm$0.0} & 43.50\tiny{$\pm$0.8} & 0.00\tiny{$\pm$0.0} & 16.10\tiny{$\pm$0.6} & 0.00\tiny{$\pm$0.0} & 10.13\tiny{$\pm$0.7} & 5.07\tiny{$\pm$0.4} \\
        LwF \cite{LwF} & YOLO11x & 88.4\tiny{$\pm$0.5} & 57.30\tiny{$\pm$0.5} & 38.00\tiny{$\pm$0.3} & 40.30\tiny{$\pm$0.7} & 30.30\tiny{$\pm$0.4} & 64.82\tiny{$\pm$0.6} & \textit{\underline{74.57}\tiny{$\pm$0.2}} & 69.70\tiny{$\pm$0.3} \\
        ERD \cite{ERD} & YOLO11x & 88.4\tiny{$\pm$0.5} & 23.70\tiny{$\pm$0.7} & 46.80\tiny{$\pm$0.8} & 26.00\tiny{$\pm$0.5} & 23.60\tiny{$\pm$0.3} & 26.81\tiny{$\pm$0.4} & 50.66\tiny{$\pm$0.2} & 38.74\tiny{$\pm$0.8} \\
        \rowcolor{lightblue} \textbf{DuET (Ours)} & YOLO11x & 88.4\tiny{$\pm$0.5} & 74.30\tiny{$\pm$0.3} & 52.40\tiny{$\pm$0.2} & 44.50\tiny{$\pm$0.1} & 46.80\tiny{$\pm$0.1} & \textit{\underline{84.05}\tiny{$\pm$0.5}} & \textbf{90.73\tiny{$\pm$0.3}} & \textbf{87.39\tiny{$\pm$0.6}} \\
    \bottomrule
    \end{tabular}
    }
\end{table*}
\begin{table*}[!ht]
    \centering
    \caption{Results of various methods on Clipart [1:10] $\rightarrow$ VOC [11:20] with different base detectors. Among columns, best in \textbf{bold}, second best \textit{\underline{underlined}}.}
    \label{E2_exps_table}
    \vspace{-2mm}
    \resizebox{\textwidth}{!}{%
    \setlength{\tabcolsep}{0.7pt}
    \renewcommand{\arraystretch}{1.08}
    \begin{tabular}{cc||c|c|c|c|c||ccc}
    \toprule
        \multirow{3}{*}{\textbf{Method}} & \multirow{3}{*}{\textbf{Base Detector}} & 
        \multirow{3}{*}{\makecell{\textbf{T1} \\ \textbf{Clipart } \\ \textbf{[1:10]}}} &
        \multicolumn{4}{c||}{\textbf{T2: VOC [11:20]}} &
        \multirow{3}{*}{\makecell{\textbf{Avg RI} \\ \textbf{(\%)}}} & 
        \multirow{3}{*}{\makecell{\textbf{Avg GI} \\ \textbf{(\%)}}} & 
        \multirow{3}{*}{\makecell{\textbf{RAI} \\ \textbf{(\%)}}} \\
        \cline{4-7}
        & & & \makecell{\textbf{Old}} &  
        \makecell{\textbf{New}} &
        \multicolumn{2}{c||}{\textbf{Unseen}} \\
        \cline{4-7}
        & & & \makecell{\textbf{Clipart [1:10]}} & 
        \makecell{\textbf{VOC [11:20]}} & 
        \makecell{\textbf{VOC [1:10]}} & 
        \makecell{\textbf{Clipart [11:20]}} & & & \\
        \hline
        \hline
        LDB \cite{LDB} & ViTDet & 36.4\tiny{$\pm$0.4} & 16.30\tiny{$\pm$0.3} & 23.80\tiny{$\pm$0.5} & 7.10\tiny{$\pm$0.2} & 9.10\tiny{$\pm$0.6} & 44.78\tiny{$\pm$0.7} & 15.81\tiny{$\pm$0.4} & 30.30\tiny{$\pm$0.8} \\
        \rowcolor{lightblue} \textbf{DuET (Ours)} & ViTDet & 36.4\tiny{$\pm$0.2} & 31.20\tiny{$\pm$0.3} & 34.50\tiny{$\pm$0.3} & 24.40\tiny{$\pm$0.3} & 1.60\tiny{$\pm$0.1} & 85.71\tiny{$\pm$0.1} & 18.23\tiny{$\pm$0.2} & 51.97\tiny{$\pm$0.3} \\
        \hline
        CL-DETR \cite{CL-DETR} & Deformable DETR & 10.5\tiny{$\pm$0.6} & 8.88\tiny{$\pm$0.7} & 27.02\tiny{$\pm$0.4} & 3.88\tiny{$\pm$0.3} & 10.33\tiny{$\pm$0.8} & 84.57\tiny{$\pm$0.5} & \textbf{54.35}\tiny{$\pm$0.2} & \textit{\underline{69.46}}\tiny{$\pm$0.7} \\
        \rowcolor{lightblue} \textbf{DuET (Ours)} & Deformable DETR & 10.5\tiny{$\pm$0.2} & 9.54\tiny{$\pm$0.1} & 20.08\tiny{$\pm$0.1} & 3.17\tiny{$\pm$0.2} & 10.23\tiny{$\pm$0.2} & \textbf{90.86}\tiny{$\pm$0.1} & \textit{\underline{53.22}}\tiny{$\pm$0.2} & \textbf{72.04}\tiny{$\pm$0.2} \\
        \hline
        Sequential FT & RTDETR-l & 44.2\tiny{$\pm$0.5} & 0.00\tiny{$\pm$0.0} & 81.50\tiny{$\pm$0.7} & 0.00\tiny{$\pm$0.0} & 30.80\tiny{$\pm$0.6} & 0.00\tiny{$\pm$0.0} & 29.79\tiny{$\pm$0.8} & 14.90\tiny{$\pm$0.4} \\
        LwF \cite{LwF} & RTDETR-l & 44.2\tiny{$\pm$0.4} & 2.81\tiny{$\pm$0.2} & 66.00\tiny{$\pm$0.7} & 0.73\tiny{$\pm$0.3} & 37.20\tiny{$\pm$0.5} & 6.36\tiny{$\pm$0.6} & 36.39\tiny{$\pm$0.4} & 21.38\tiny{$\pm$0.8} \\
        ERD \cite{ERD} & RTDETR-l & 44.2\tiny{$\pm$0.5} & 0.37\tiny{$\pm$0.3} & 81.20\tiny{$\pm$0.6} & 2.81\tiny{$\pm$0.4} & 27.50\tiny{$\pm$0.7} & 0.84\tiny{$\pm$0.2} & 28.21\tiny{$\pm$0.8} & 14.53\tiny{$\pm$0.5} \\
        \rowcolor{lightblue} \textbf{DuET (Ours)} & RTDETR-l & 44.2\tiny{$\pm$0.1} & 37.80\tiny{$\pm$0.2} & 8.17\tiny{$\pm$0.1} & 29.40\tiny{$\pm$0.2} & 13.20\tiny{$\pm$0.2} & 85.52\tiny{$\pm$0.1} & 29.64\tiny{$\pm$0.2} & \textit{\underline{57.58}\tiny{$\pm$0.1}} \\
        \hline
        Sequential FT & RTDETR-x & 47.0\tiny{$\pm$0.6} & 0.00\tiny{$\pm$0.0} & 81.60\tiny{$\pm$0.7} & 0.00\tiny{$\pm$0.0} & 35.70\tiny{$\pm$0.5} & 0.00\tiny{$\pm$0.0} & 37.27\tiny{$\pm$0.8} & 18.64\tiny{$\pm$0.3} \\
        LwF \cite{LwF} & RTDETR-x & 47.0\tiny{$\pm$0.5} & 2.42\tiny{$\pm$0.3} & 64.30\tiny{$\pm$0.6} & 1.25\tiny{$\pm$0.2} & 35.70\tiny{$\pm$0.8} & 5.15\tiny{$\pm$0.4} & 37.97\tiny{$\pm$0.7} & 21.56\tiny{$\pm$0.5} \\
        ERD \cite{ERD} & RTDETR-x & 47.0\tiny{$\pm$0.4} & 0.67\tiny{$\pm$0.2} & 82.00\tiny{$\pm$0.7} & 0.93\tiny{$\pm$0.3} & 34.40\tiny{$\pm$0.6} & 1.43\tiny{$\pm$0.5} & 36.43\tiny{$\pm$0.8} & 18.93\tiny{$\pm$0.4} \\
        \rowcolor{lightblue} \textbf{DuET (Ours)} & RTDETR-x & 47.0\tiny{$\pm$0.2} & 41.10\tiny{$\pm$0.2} & 4.64\tiny{$\pm$0.2} & 21.30\tiny{$\pm$0.1} & 5.27\tiny{$\pm$0.2} & \textit{\underline{87.45}}\tiny{$\pm$0.2} & 17.44\tiny{$\pm$0.2} & 52.45\tiny{$\pm$0.2} \\
        \hline
        Sequential FT & YOLO11n & 47.1\tiny{$\pm$0.7} & 0.00\tiny{$\pm$0.0} & 73.60\tiny{$\pm$0.6} & 0.00\tiny{$\pm$0.0} & 29.10\tiny{$\pm$0.5} & 0.00\tiny{$\pm$0.0} & 30.12\tiny{$\pm$0.8} & 15.06\tiny{$\pm$0.4} \\
        LwF \cite{LwF} & YOLO11n & 47.1\tiny{$\pm$0.6} & 31.40\tiny{$\pm$0.7} & 4.00\tiny{$\pm$0.5} & 20.30\tiny{$\pm$0.3} & 5.36\tiny{$\pm$0.8} & 66.67\tiny{$\pm$0.4} & 18.17\tiny{$\pm$0.2} & 42.42\tiny{$\pm$0.7} \\
        ERD \cite{ERD} & YOLO11n & 47.1\tiny{$\pm$0.5} & 33.20\tiny{$\pm$0.3} & 0.72\tiny{$\pm$0.6} & 20.70\tiny{$\pm$0.4} & 0.63\tiny{$\pm$0.7} & 70.49\tiny{$\pm$0.2} & 13.53\tiny{$\pm$0.8} & 42.01\tiny{$\pm$0.5} \\
        \rowcolor{lightblue} \textbf{DuET (Ours)} & YOLO11n & 47.1\tiny{$\pm$0.2} & 32.70\tiny{$\pm$0.1} & 44.00\tiny{$\pm$0.2} & 21.70\tiny{$\pm$0.2} & 26.10\tiny{$\pm$0.1} & 69.43\tiny{$\pm$0.2} & 40.51\tiny{$\pm$0.2} & 54.97\tiny{$\pm$0.1} \\
        \hline
        Sequential FT & YOLO11x & 36.3\tiny{$\pm$0.5} & 0.00\tiny{$\pm$0.0} & 77.50\tiny{$\pm$0.6} & 0.00\tiny{$\pm$0.0} & 33.50\tiny{$\pm$0.8} & 0.00\tiny{$\pm$0.0} & 38.15\tiny{$\pm$0.7} & 19.08\tiny{$\pm$0.4} \\
        LwF \cite{LwF} & YOLO11x & 36.3\tiny{$\pm$0.4} & 25.10\tiny{$\pm$0.3} & 0.96\tiny{$\pm$0.8} & 13.00\tiny{$\pm$0.5} & 1.59\tiny{$\pm$0.6} & 69.15\tiny{$\pm$0.4} & 9.16\tiny{$\pm$0.7} & 39.16\tiny{$\pm$0.3} \\
        ERD \cite{ERD} & YOLO11x & 36.3\tiny{$\pm$0.5} & 29.40\tiny{$\pm$0.7} & 0.52\tiny{$\pm$0.3} & 12.90\tiny{$\pm$0.6} & 0.68\tiny{$\pm$0.2} & 80.99\tiny{$\pm$0.8} & 8.07\tiny{$\pm$0.4} & 44.53\tiny{$\pm$0.5} \\
        \rowcolor{lightblue} \textbf{DuET (Ours)} & YOLO11x & 36.3\tiny{$\pm$0.2} & 19.30\tiny{$\pm$0.1} & 1.25\tiny{$\pm$0.2} & 6.06\tiny{$\pm$0.2} & 3.03\tiny{$\pm$0.1} & 53.17\tiny{$\pm$0.2} & 6.88\tiny{$\pm$0.1} & 30.03\tiny{$\pm$0.2} \\
    \bottomrule
    \end{tabular}
    }
\end{table*}

\clearpage
\begin{table*}[!ht]
    \centering
    \caption{Results of various methods on Daytime Sunny [1:4] $\rightarrow$ Night Sunny [5:7] with different base detectors. Among columns, best in \textbf{bold}, second best \textit{\underline{underlined}}.}
    \label{E4_exps_table}
    \vspace{-2mm}
    \resizebox{\textwidth}{!}{
    \setlength{\tabcolsep}{1.8pt}
    \begin{tabular}{cc||c|c|c|c|c||ccc}
    \toprule
        \multirow{4}{*}{\textbf{Method}} & \multirow{4}{*}{\textbf{Base Detector}} & 
        \multirow{4}{*}{\makecell{\textbf{T1} \\ \textbf{Daytime} \\ \textbf{Sunny} \\ \textbf{{[1:4]}}}} &
        \multicolumn{4}{c||}{\textbf{T2: Night Sunny [5:7]}} &
        \multirow{4}{*}{\makecell{\textbf{Avg RI} \\ \textbf{(\%)}}} & 
        \multirow{4}{*}{\makecell{\textbf{Avg GI} \\ \textbf{(\%)}}} & 
        \multirow{4}{*}{\makecell{\textbf{RAI} \\ \textbf{(\%)}}} \\
        \cline{4-7}
        & & & \makecell{\textbf{Old}} &  
        \makecell{\textbf{New}} &
        \multicolumn{2}{c||}{\textbf{Unseen}} \\
        \cline{4-7}
        & & & \multirow{2}{*}{\makecell{\textbf{Daytime} \\ \textbf{Sunny [1:4]}}} & 
        \multirow{2}{*}{\makecell{\textbf{Night} \\ \textbf{Sunny [5:7]}}} & 
        \multirow{2}{*}{\makecell{\textbf{Night} \\ \textbf{Sunny [1:4]}}} & 
        \multirow{2}{*}{\makecell{\textbf{Daytime} \\ \textbf{Sunny [5:7]}}} & & & \\
        & & & & & & & & & \\
        \hline
        \hline
        LDB \cite{LDB} & VitDet & 45.3\tiny{$\pm$0.6} & 0.50\tiny{$\pm$0.3} & 15.10\tiny{$\pm$0.4} & 0.30\tiny{$\pm$0.5} & 16.90\tiny{$\pm$0.7} & 1.10\tiny{$\pm$0.2} & 22.41\tiny{$\pm$0.3} & 11.76\tiny{$\pm$0.6} \\
        \rowcolor{lightblue} \textbf{DuET (Ours)} & VitDet & 45.3\tiny{$\pm$0.2} & 12.48\tiny{$\pm$0.3} & 11.60\tiny{$\pm$0.3} & 4.33\tiny{$\pm$0.2} & 9.60\tiny{$\pm$0.2} & 27.55\tiny{$\pm$0.2} & 28.22\tiny{$\pm$0.1} & 27.89\tiny{$\pm$0.2} \\
        \hline
        CL-DETR \cite{CL-DETR} & Deformable DETR & 46.3\tiny{$\pm$0.4} & 27.41\tiny{$\pm$0.5} & 31.94\tiny{$\pm$0.6} & 19.85\tiny{$\pm$0.3} & 32.55\tiny{$\pm$0.4} & 59.20\tiny{$\pm$0.2} & \textit{\underline{54.96}}\tiny{$\pm$0.5} & 57.08\tiny{$\pm$0.4} \\
        \rowcolor{lightblue} \textbf{DuET (Ours)} & Deformable DETR & 46.3\tiny{$\pm$0.2} & 39.1\tiny{$\pm$0.1} & 15.06\tiny{$\pm$0.2} & 28.17\tiny{$\pm$0.2} & 4.33\tiny{$\pm$0.1} & 84.45\tiny{$\pm$0.2} & 33.45\tiny{$\pm$0.1} & 58.95\tiny{$\pm$0.2} \\
        \hline
        Sequential FT & RTDETR-l & 57.2\tiny{$\pm$0.5} & 0.00\tiny{$\pm$0.0} & 77.40\tiny{$\pm$0.7} & 2.52\tiny{$\pm$0.3} & 39.80\tiny{$\pm$0.6} & 0.00\tiny{$\pm$0.0} & 35.36\tiny{$\pm$0.8} & 17.68\tiny{$\pm$0.4} \\
        LwF \cite{LwF} & RTDETR-l & 57.2\tiny{$\pm$0.4} & 0.15\tiny{$\pm$0.2} & 76.40\tiny{$\pm$0.7} & 0.03\tiny{$\pm$0.1} & 41.50\tiny{$\pm$0.5} & 0.26\tiny{$\pm$0.2} & 35.01\tiny{$\pm$0.8} & 17.64\tiny{$\pm$0.4} \\
        ERD \cite{ERD} & RTDETR-l & 57.2\tiny{$\pm$0.5} & 0.09\tiny{$\pm$0.1} & 80.50\tiny{$\pm$0.7} & 0.04\tiny{$\pm$0.1} & 39.80\tiny{$\pm$0.6} & 0.16\tiny{$\pm$0.2} & 33.59\tiny{$\pm$0.8} & 16.88\tiny{$\pm$0.4} \\
        \rowcolor{lightblue} \textbf{DuET (Ours)} & RTDETR-l & 57.2\tiny{$\pm$0.2} & 27.30\tiny{$\pm$0.1} & 8.63\tiny{$\pm$0.2} & 20.10\tiny{$\pm$0.2} & 7.88\tiny{$\pm$0.1} & 47.73\tiny{$\pm$0.2} & 21.00\tiny{$\pm$0.1} & 34.37\tiny{$\pm$0.2} \\
        \hline
        Sequential FT & RTDETR-x & 61.0\tiny{$\pm$0.6} & 0.00\tiny{$\pm$0.0} & 84.80\tiny{$\pm$0.7} & 0.00\tiny{$\pm$0.0} & 40.80\tiny{$\pm$0.5} & 0.00\tiny{$\pm$0.0} & 33.77\tiny{$\pm$0.8} & 16.89\tiny{$\pm$0.3} \\
        LwF \cite{LwF} & RTDETR-x & 61.0\tiny{$\pm$0.5} & 0.57\tiny{$\pm$0.2} & 79.10\tiny{$\pm$0.7} & 0.61\tiny{$\pm$0.3} & 40.60\tiny{$\pm$0.6} & 0.93\tiny{$\pm$0.2} & 34.03\tiny{$\pm$0.8} & 17.48\tiny{$\pm$0.4} \\
        ERD \cite{ERD} & RTDETR-x & 61.0\tiny{$\pm$0.4} & 0.81\tiny{$\pm$0.2} & 84.80\tiny{$\pm$0.7} & 0.98\tiny{$\pm$0.3} & 38.90\tiny{$\pm$0.6} & 1.33\tiny{$\pm$0.2} & 32.87\tiny{$\pm$0.8} & 17.10\tiny{$\pm$0.4} \\
        \rowcolor{lightblue} \textbf{DuET (Ours)} & RTDETR-x & 61.0\tiny{$\pm$0.2} & 34.40\tiny{$\pm$0.1} & 6.51\tiny{$\pm$0.2} & 28.10\tiny{$\pm$0.2} & 6.02\tiny{$\pm$0.1} & 56.39\tiny{$\pm$0.2} & 24.15\tiny{$\pm$0.1} & 40.27\tiny{$\pm$0.2} \\
        \hline
        Sequential FT & YOLO11n & 49.4\tiny{$\pm$0.3} & 0.00\tiny{$\pm$0.0} & 62.20\tiny{$\pm$0.5} & 12.60\tiny{$\pm$0.4} & 35.90\tiny{$\pm$0.3} & 0.00\tiny{$\pm$0.0} & 45.88\tiny{$\pm$0.6} & 22.94\tiny{$\pm$0.3} \\
        LwF \cite{LwF} & YOLO11n & 49.4\tiny{$\pm$0.2} & 27.60\tiny{$\pm$0.4} & 0.34\tiny{$\pm$0.6} & 21.30\tiny{$\pm$0.3} & 0.67\tiny{$\pm$0.5} & 55.87\tiny{$\pm$0.3} & 21.88\tiny{$\pm$0.7} & 38.88\tiny{$\pm$0.6} \\
        ERD \cite{ERD} & YOLO11n & 49.4\tiny{$\pm$0.5} & 33.00\tiny{$\pm$0.4} & 34.00\tiny{$\pm$0.3} & 26.10\tiny{$\pm$0.6} & 29.10\tiny{$\pm$0.7} & 66.80\tiny{$\pm$0.5} & 53.04\tiny{$\pm$0.3} & 59.92\tiny{$\pm$0.4} \\
        \rowcolor{lightblue} \textbf{DuET (Ours)} & YOLO11n & 49.4\tiny{$\pm$0.2} & 43.50\tiny{$\pm$0.1} & 22.20\tiny{$\pm$0.3} & 31.60\tiny{$\pm$0.2} & 27.40\tiny{$\pm$0.1} & 88.06\tiny{$\pm$0.2} & \textbf{56.95}\tiny{$\pm$0.1} & \textbf{72.51}\tiny{$\pm$0.2} \\
        \hline
        Sequential FT & YOLO11x & 64.2\tiny{$\pm$0.6} & 12.50\tiny{$\pm$0.4} & 68.60\tiny{$\pm$0.7} & 18.80\tiny{$\pm$0.3} & 46.00\tiny{$\pm$0.8} & 19.47\tiny{$\pm$0.2} & 47.60\tiny{$\pm$0.5} & 33.54\tiny{$\pm$0.4} \\
        LwF \cite{LwF} & YOLO11x & 64.2\tiny{$\pm$0.7} & 62.10\tiny{$\pm$0.6} & 0.04\tiny{$\pm$0.2} & 42.40\tiny{$\pm$0.8} & 0.01\tiny{$\pm$0.1} & \textit{\underline{96.73}}\tiny{$\pm$0.5} & 27.79\tiny{$\pm$0.4} & 62.26\tiny{$\pm$0.8} \\
        ERD \cite{ERD} & YOLO11x & 64.2\tiny{$\pm$0.6} & 62.20\tiny{$\pm$0.7} & 0.06\tiny{$\pm$0.2} & 42.70\tiny{$\pm$0.8} & 0.07\tiny{$\pm$0.1} & 95.95\tiny{$\pm$0.5} & 28.04\tiny{$\pm$0.4} & 62.46\tiny{$\pm$0.8} \\
        \rowcolor{lightblue} \textbf{DuET (Ours)} & YOLO11x & 64.2\tiny{$\pm$0.2} & 61.60\tiny{$\pm$0.1} & 12.90\tiny{$\pm$0.2} & 44.70\tiny{$\pm$0.2} & 17.10\tiny{$\pm$0.1} & \textbf{96.88}\tiny{$\pm$0.2} & 42.41\tiny{$\pm$0.1} & \textit{\underline{69.18}}\tiny{$\pm$0.2} \\
    \bottomrule
    \end{tabular}
    }
\end{table*}
\begin{table*}[!ht]
    \centering
    \caption{Results of various methods on Night Sunny [1:4] $\rightarrow$ Daytime Sunny [5:7] with different base detectors. Among columns, best in \textbf{bold}, second best \textit{\underline{underlined}}.}
    \label{E5_exps_table}
    \vspace{-2mm}
    \resizebox{\textwidth}{!}{%
    \setlength{\tabcolsep}{2.2pt}
    \begin{tabular}{cc||c|c|c|c|c||ccc}
    \toprule
        \multirow{4}{*}{\textbf{Method}} & \multirow{4}{*}{\textbf{Base Detector}} & 
        \multirow{4}{*}{\makecell{\textbf{T1} \\ \textbf{Night} \\ \textbf{Sunny} \\ \textbf{{[1:4]}}}} &
        \multicolumn{4}{c||}{\textbf{T2: Daytime Sunny [5:7]}} &
        \multirow{4}{*}{\makecell{\textbf{Avg RI} \\ \textbf{(\%)}}} & 
        \multirow{4}{*}{\makecell{\textbf{Avg GI} \\ \textbf{(\%)}}} & 
        \multirow{4}{*}{\makecell{\textbf{RAI} \\ \textbf{(\%)}}} \\
        \cline{4-7}
        & & & \makecell{\textbf{Old}} &  
        \makecell{\textbf{New}} &
        \multicolumn{2}{c||}{\textbf{Unseen}} \\
        \cline{4-7}
        & & &  \multirow{2}{*}{\makecell{\textbf{Night} \\ \textbf{Sunny [1:4]}}} & 
        \multirow{2}{*}{\makecell{\textbf{Daytime} \\ \textbf{Sunny [5:7]}}} & 
        \multirow{2}{*}{\makecell{\textbf{Daytime} \\ \textbf{Sunny [1:4]}}} & 
        \multirow{2}{*}{\makecell{\textbf{Night} \\ \textbf{Sunny [5:7]}}} & & & \\
        & & & & & & & & & \\
    \hline
    \hline
        LDB \cite{LDB} & ViTDet & 37.0\tiny{$\pm$0.6} & 0.40\tiny{$\pm$0.4} & 18.30\tiny{$\pm$0.7} & 0.10\tiny{$\pm$0.2} & 14.30\tiny{$\pm$0.5} & 1.08\tiny{$\pm$0.3} & 20.66\tiny{$\pm$0.7} & 10.87\tiny{$\pm$0.8} \\
        \rowcolor{lightblue} \textbf{DuET (Ours)} & VitDet & 37.0\tiny{$\pm$0.2} & 5.50\tiny{$\pm$0.1} & 18.33\tiny{$\pm$0.1} & 3.33\tiny{$\pm$0.5} & 14.70\tiny{$\pm$0.2} & 14.86\tiny{$\pm$0.1} & 24.80\tiny{$\pm$0.2} & 19.83\tiny{$\pm$0.1} \\
        \hline
        CL-DETR \cite{CL-DETR} & Deformable DETR & 48.8\tiny{$\pm$0.7} & 25.90\tiny{$\pm$0.8} & 45.70\tiny{$\pm$0.5} & 19.88\tiny{$\pm$0.6} & 41.33\tiny{$\pm$0.2} & 53.04\tiny{$\pm$0.7} & \textit{\underline{55.64}}\tiny{$\pm$0.3} & 54.34\tiny{$\pm$0.8} \\
        \rowcolor{lightblue} \textbf{DuET (Ours)} & Deformable DETR & 48.8\tiny{$\pm$0.2} & 38.81\tiny{$\pm$0.1} & 4.24\tiny{$\pm$0.2} & 29.09\tiny{$\pm$0.1} & 7.59\tiny{$\pm$0.2} & 79.48\tiny{$\pm$0.1} & 37.69\tiny{$\pm$0.2} & 58.59\tiny{$\pm$0.2} \\
        \hline
        Sequential FT & RTDETR-l & 70.0\tiny{$\pm$0.6} & 0.00\tiny{$\pm$0.0} & 58.90\tiny{$\pm$0.7} & 0.00\tiny{$\pm$0.0} & 40.80\tiny{$\pm$0.5} & 0.00\tiny{$\pm$0.0} & 24.64\tiny{$\pm$0.8} & 12.32\tiny{$\pm$0.3} \\
        LwF \cite{LwF} & RTDETR-l & 70.0\tiny{$\pm$0.7} & 8.73\tiny{$\pm$0.8} & 21.80\tiny{$\pm$0.5} & 6.00\tiny{$\pm$0.6} & 8.37\tiny{$\pm$0.4} & 12.47\tiny{$\pm$0.7} & 10.30\tiny{$\pm$0.3} & 11.39\tiny{$\pm$0.8} \\
        ERD \cite{ERD} & RTDETR-l & 70.0\tiny{$\pm$0.6} & 1.09\tiny{$\pm$0.7} & 57.60\tiny{$\pm$0.5} & 1.86\tiny{$\pm$0.8} & 43.10\tiny{$\pm$0.4} & 1.56\tiny{$\pm$0.7} & 27.65\tiny{$\pm$0.6} & 14.61\tiny{$\pm$0.3} \\
        \rowcolor{lightblue} \textbf{DuET (Ours)} & RTDETR-l & 70.0\tiny{$\pm$0.2} & 48.70\tiny{$\pm$0.2} & 1.86\tiny{$\pm$0.1} & 43.80\tiny{$\pm$0.2} & 1.03\tiny{$\pm$0.1} & 69.57\tiny{$\pm$0.2} & 38.91\tiny{$\pm$0.2} & 54.24\tiny{$\pm$0.2} \\
        \hline
        Sequential FT & RTDETR-x & 73.3\tiny{$\pm$0.7} & 0.00\tiny{$\pm$0.0} & 58.80\tiny{$\pm$0.8} & 0.00\tiny{$\pm$0.0} & 44.10\tiny{$\pm$0.6} & 0.00\tiny{$\pm$0.0} & 24.39\tiny{$\pm$0.7} & 12.20\tiny{$\pm$0.3} \\
        LwF \cite{LwF} & RTDETR-x & 73.3\tiny{$\pm$0.6} & 10.70\tiny{$\pm$0.7} & 9.25\tiny{$\pm$0.5} & 6.02\tiny{$\pm$0.6} & 7.66\tiny{$\pm$0.4} & 14.60\tiny{$\pm$0.7} & 9.17\tiny{$\pm$0.3} & 11.89\tiny{$\pm$0.8} \\
        ERD \cite{ERD} & RTDETR-x & 73.3\tiny{$\pm$0.7} & 7.03\tiny{$\pm$0.8} & 57.00\tiny{$\pm$0.5} & 0.93\tiny{$\pm$0.6} & 42.50\tiny{$\pm$0.4} & 9.59\tiny{$\pm$0.7} & 24.27\tiny{$\pm$0.8} & 16.93\tiny{$\pm$0.3} \\
        \rowcolor{lightblue} \textbf{DuET (Ours)} & RTDETR-x & 73.3\tiny{$\pm$0.2} & 66.30\tiny{$\pm$0.2} & 6.40\tiny{$\pm$0.1} & 58.90\tiny{$\pm$0.2} & 5.26\tiny{$\pm$0.1} & \textit{\underline{90.45}}\tiny{$\pm$0.2} & 51.19\tiny{$\pm$0.2} & \textit{\underline{70.82}}\tiny{$\pm$0.2} \\
        \hline
        Sequential FT & YOLO11n & 50.1\tiny{$\pm$0.7} & 0.12\tiny{$\pm$0.8} & 37.60\tiny{$\pm$0.5} & 0.25\tiny{$\pm$0.6} & 25.8\tiny{$\pm$0.4} & 0.24\tiny{$\pm$0.7} & 19.48\tiny{$\pm$0.8} & 9.86\tiny{$\pm$0.3} \\
        LwF \cite{LwF} & YOLO11n & 50.1\tiny{$\pm$0.6} & 39.00\tiny{$\pm$0.7} & 0.29\tiny{$\pm$0.2} & 33.90\tiny{$\pm$0.8} & 1.51\tiny{$\pm$0.3} & 77.84\tiny{$\pm$0.5} & 35.44\tiny{$\pm$0.7} & 56.64\tiny{$\pm$0.4} \\
        ERD \cite{ERD} & YOLO11n & 50.1\tiny{$\pm$0.7} & 39.60\tiny{$\pm$0.6} & 0.06\tiny{$\pm$0.2} & 34.20\tiny{$\pm$0.8} & 0.04\tiny{$\pm$0.1} & 79.04\tiny{$\pm$0.5} & 34.65\tiny{$\pm$0.7} & 56.85\tiny{$\pm$0.4} \\
        \rowcolor{lightblue} \textbf{DuET (Ours)} & YOLO11n & 50.1\tiny{$\pm$0.2} & 47.10\tiny{$\pm$0.2} & 20.10\tiny{$\pm$0.1} & 41.30\tiny{$\pm$0.2} & 19.10\tiny{$\pm$0.1} & \textbf{94.01}\tiny{$\pm$0.2} & \textbf{56.03}\tiny{$\pm$0.2} & \textbf{75.02}\tiny{$\pm$0.2} \\
        \hline
        Sequential FT & YOLO11x & 76.3\tiny{$\pm$0.7} & 0.33\tiny{$\pm$0.8} & 50.20\tiny{$\pm$0.5} & 2.62\tiny{$\pm$0.6} & 44.60\tiny{$\pm$0.4} & 0.43\tiny{$\pm$0.7} & 28.51\tiny{$\pm$0.8} & 14.47\tiny{$\pm$0.3} \\
        LwF \cite{LwF} & YOLO11x & 76.3\tiny{$\pm$0.8} & 67.50\tiny{$\pm$0.7} & 0.66\tiny{$\pm$0.2} & 50.10\tiny{$\pm$0.8} & 0.23\tiny{$\pm$0.1} & 88.47\tiny{$\pm$0.5} & 39.16\tiny{$\pm$0.7} & 63.82\tiny{$\pm$0.4} \\
        ERD \cite{ERD} & YOLO11x & 76.3\tiny{$\pm$0.7} & 67.70\tiny{$\pm$0.8} & 0.46\tiny{$\pm$0.2} & 51.30\tiny{$\pm$0.8} & 0.25\tiny{$\pm$0.1} & 88.73\tiny{$\pm$0.5} & 40.10\tiny{$\pm$0.7} & 64.42\tiny{$\pm$0.4} \\
        \rowcolor{lightblue} \textbf{DuET (Ours)} & YOLO11x & 76.3\tiny{$\pm$0.2} & 22.60\tiny{$\pm$0.1} & 43.70\tiny{$\pm$0.2} & 21.60\tiny{$\pm$0.2} & 40.00\tiny{$\pm$0.1} & 29.62\tiny{$\pm$0.2} & 40.58\tiny{$\pm$0.2} & 35.10\tiny{$\pm$0.2} \\
    \bottomrule
    \end{tabular}
    }
\end{table*}

\clearpage
\begin{table*}[!ht]
    \centering
    \caption{Computational complexity analysis of various methods evaluated on the multi-phase experiment: Watercolor [1:3] $\rightarrow$ Comic [4:6] $\rightarrow$ Clipart [7:13] $\rightarrow$ VOC [14:20]. The inference speed and memory footprint are averaged across all incremental tasks.}
    \label{tab:complexity_analysis}
    \vspace{-2mm}
    \resizebox{0.8\textwidth}{!}{%
    \begin{tabular}{cc|cccc}
    \toprule
        \multirow{2}{*}{\textbf{Method}}
         & \multirow{2}{*}{\textbf{Base Detector}} & \multirow{2}{*}{\textbf{GFLOPs}} & \multirow{2}{*}{\makecell{\textbf{Trainable} \\ \textbf{Params (M)}}} & \multirow{2}{*}{\makecell{\textbf{Avg. Inference} \\ \textbf{Speed (ms)}}} & \multirow{2}{*}{\makecell{\textbf{Avg. Memory} \\ \textbf{Footprint (GB)}}} 
        \\
        & & & & \\
        \hline
        Sequential FT & YOLO11n \cite{YOLO11} & 6.3 & 2.58 & 9.150 & 0.235 \\
        LwF \cite{LwF} & YOLO11n \cite{YOLO11} & 6.3 & 2.58 & 9.125 & 0.261 \\
        ERD \cite{ERD} & YOLO11n \cite{YOLO11} & 6.3 & 2.58 & 9.075 & 0.257 \\
        LDB \cite{LDB} & ViTDet \cite{VitDet} & 1829.61 & 110.52 & 137.92 & 1.818 \\
        CL-DETR \cite{CL-DETR} & Deformable DETR \cite{DeformableDETR} & 11.77 & 39.85 & 39.075 & 0.789 \\
        \rowcolor{lightblue} \textbf{DuET (Ours)} & YOLO11n \cite{YOLO11} & 6.3 & 2.58 & 4.4 & 0.244 \\
    \bottomrule
    \end{tabular}
    }
\end{table*}
\clearpage
\begin{table}[t]
    \centering
    \caption{Results of various methods on Watercolor [1:3] $\rightarrow$ Comic [4:6] $\rightarrow$ Clipart [7:13] $\rightarrow$ VOC [14:20] with different base detectors. Among columns, best in \textbf{bold}, second best \textit{\underline{underlined}}.}
    \label{E3_exps_table}
    \renewcommand{\arraystretch}{1.2}
    \setlength{\tabcolsep}{1.2pt}
    \resizebox{\columnwidth}{!}{%
    \begin{tabular}{cc||ccc}
    \toprule
        \textbf{Method} & \textbf{Base Detector} & \textbf{Avg RI (\%)} & \textbf{Avg GI (\%)} & \textbf{RAI (\%)} \\
    \hline
    \hline
        LDB \cite{LDB} & ViTDet & 86.08\scriptsize{$\pm$0.6} & 19.57\scriptsize{$\pm$0.5} & 52.83\scriptsize{$\pm$0.4} \\
        \rowcolor{lightblue} \textbf{DuET (Ours)} & ViTDet & 65.57\scriptsize{$\pm$0.2} & 40.44\scriptsize{$\pm$0.1} & 53.01\scriptsize{$\pm$0.2} \\
        \hline
        CL-DETR \cite{CL-DETR} & Deformable DETR & 71.73\scriptsize{$\pm$0.5} & 36.63\scriptsize{$\pm$0.6} & 54.18\scriptsize{$\pm$0.4} \\
        \rowcolor{lightblue} \textbf{DuET (Ours)} & Deformable DETR & 88.54\scriptsize{$\pm$0.2} & 34.81\scriptsize{$\pm$0.1} & 61.68\scriptsize{$\pm$0.2} \\
        \hline
        Sequential FT & RTDETR-l & 0.00\scriptsize{$\pm$0.0} & 7.76\scriptsize{$\pm$0.4} & 3.88\scriptsize{$\pm$0.3} \\
        LwF \cite{LwF} & RTDETR-l & 0.40\scriptsize{$\pm$0.2} & 13.47\scriptsize{$\pm$0.7} & 6.94\scriptsize{$\pm$0.5} \\
        ERD \cite{ERD} & RTDETR-l & 1.05\scriptsize{$\pm$0.4} & 17.91\scriptsize{$\pm$0.6} & 9.48\scriptsize{$\pm$0.3} \\
        \rowcolor{lightblue} \textbf{DuET (Ours)} & RTDETR-l & 24.75\scriptsize{$\pm$0.2} & 22.89\scriptsize{$\pm$0.2} & 23.82\scriptsize{$\pm$0.1} \\
        \hline
        Sequential FT & RTDETR-x & 0.00\scriptsize{$\pm$0.0} & 6.35\scriptsize{$\pm$0.4} & 3.18\scriptsize{$\pm$0.2} \\
        LwF \cite{LwF} & RTDETR-x & 65.51\scriptsize{$\pm$0.8} & 16.69\scriptsize{$\pm$0.3} & 41.10\scriptsize{$\pm$0.7} \\
        ERD \cite{ERD} & RTDETR-x & 0.05\scriptsize{$\pm$0.1} & 10.42\scriptsize{$\pm$0.2} & 5.24\scriptsize{$\pm$0.3} \\
        \rowcolor{lightblue} \textbf{DuET (Ours)} & RTDETR-x & 40.80\scriptsize{$\pm$0.2} & 18.88\scriptsize{$\pm$0.2} & 29.84\scriptsize{$\pm$0.1} \\
        \hline
        Sequential FT & YOLO11n & 0.00\scriptsize{$\pm$0.0} & 11.05\scriptsize{$\pm$0.5} & 5.53\scriptsize{$\pm$0.3} \\
        LwF \cite{LwF} & YOLO11n & 52.66\scriptsize{$\pm$0.6} & 17.01\scriptsize{$\pm$0.4} & 34.84\scriptsize{$\pm$0.3} \\
        ERD \cite{ERD} & YOLO11n & 54.76\scriptsize{$\pm$0.5} & \textit{\underline{41.13}}\scriptsize{$\pm$0.4} & 47.95\scriptsize{$\pm$0.7} \\
        \rowcolor{lightblue} \textbf{DuET (Ours)} & YOLO11n & \textit{\underline{89.30}}\scriptsize{$\pm$0.2} & \textbf{42.60}\scriptsize{$\pm$0.1} & \textbf{65.95}\scriptsize{$\pm$0.2} \\
        \hline
        Sequential FT & YOLO11x & 0.00\scriptsize{$\pm$0.0} & 10.21\scriptsize{$\pm$0.4} & 5.11\scriptsize{$\pm$0.3} \\
        LwF \cite{LwF} & YOLO11x & 10.56\scriptsize{$\pm$0.3} & 17.46\scriptsize{$\pm$0.5} & 14.01\scriptsize{$\pm$0.2} \\
        ERD \cite{ERD} & YOLO11x & 54.19\scriptsize{$\pm$0.6} & 8.49\scriptsize{$\pm$0.2} & 31.34\scriptsize{$\pm$0.8} \\
        \rowcolor{lightblue} \textbf{DuET (Ours)} & YOLO11x & \textbf{96.72}\scriptsize{$\pm$0.2} & 26.49\scriptsize{$\pm$0.1} & \textit{\underline{61.61}}\scriptsize{$\pm$0.2} \\
    \bottomrule
    \end{tabular}
    }
\end{table}

\begin{table}[t]
    \centering
    \caption{Results of various methods on Night Sunny [1:2] $\rightarrow$ Daytime Sunny [3:4] $\rightarrow$ Daytime Foggy [5:7] with different base detectors. Among columns, best in \textbf{bold}, second best \textit{\underline{underlined}}.}
    \label{E6_exps_table}
    \renewcommand{\arraystretch}{1.2}
    \setlength{\tabcolsep}{1.2pt}
    \resizebox{\columnwidth}{!}{%
    \begin{tabular}{cc||ccc}
    \toprule
        \textbf{Method} & \textbf{Base Detector} & \textbf{Avg RI (\%)} & \textbf{Avg GI (\%)} & \textbf{RAI (\%)} \\
    \hline
    \hline
    LDB \cite{LDB} & ViTDet & 50.50\scriptsize{$\pm$0.6} & 5.42\scriptsize{$\pm$0.7} & 27.96\scriptsize{$\pm$0.5} \\
    \rowcolor{lightblue} \textbf{DuET (Ours)} & VitDet & 39.87\scriptsize{$\pm$0.2} & 17.12\scriptsize{$\pm$0.1} & 28.50\scriptsize{$\pm$0.2} \\
    \hline
    CL-DETR \cite{CL-DETR} & Deformable DETR & \textit{\underline{64.26}}\scriptsize{$\pm$0.3} & \textbf{43.46}\scriptsize{$\pm$0.5} & \textit{\underline{53.86}}\scriptsize{$\pm$0.6} \\
    \rowcolor{lightblue} \textbf{DuET (Ours)} & Deformable DETR & 62.99\scriptsize{$\pm$0.2} & 45.11\scriptsize{$\pm$0.1} & 54.05\scriptsize{$\pm$0.2} \\
    \hline
    Sequential FT & RTDETR-l & 0.00\scriptsize{$\pm$0.0} & 14.97\scriptsize{$\pm$0.4} & 7.49\scriptsize{$\pm$0.3} \\
    LwF \cite{LwF} & RTDETR-l & 14.76\scriptsize{$\pm$0.2} & 1.24\scriptsize{$\pm$0.3} & 8.00\scriptsize{$\pm$0.5} \\
    ERD \cite{ERD} & RTDETR-l & 5.40\scriptsize{$\pm$0.4} & 15.83\scriptsize{$\pm$0.6} & 10.62\scriptsize{$\pm$0.3} \\
    \rowcolor{lightblue} \textbf{DuET (Ours)} & RTDETR-l & 20.76\scriptsize{$\pm$0.2} & 10.55\scriptsize{$\pm$0.2} & 15.66\scriptsize{$\pm$0.1} \\
    \hline
    Sequential FT & RTDETR-x & 0.00\scriptsize{$\pm$0.0} & 22.74\scriptsize{$\pm$0.4} & 11.37\scriptsize{$\pm$0.3} \\
    LwF \cite{LwF} & RTDETR-x & 7.62\scriptsize{$\pm$0.3} & 8.86\scriptsize{$\pm$0.2} & 8.24\scriptsize{$\pm$0.4} \\
    ERD \cite{ERD} & RTDETR-x & 3.02\scriptsize{$\pm$0.2} & 19.16\scriptsize{$\pm$0.3} & 11.09\scriptsize{$\pm$0.2} \\
    \rowcolor{lightblue} \textbf{DuET (Ours)} & RTDETR-x & 27.39\scriptsize{$\pm$0.2} & 23.65\scriptsize{$\pm$0.1} & 25.52\scriptsize{$\pm$0.2} \\
    \hline
    Sequential FT & YOLO11n & 0.00\scriptsize{$\pm$0.0} & 30.51\scriptsize{$\pm$0.6} & 15.26\scriptsize{$\pm$0.5} \\
    LwF \cite{LwF} & YOLO11n & 27.94\scriptsize{$\pm$0.7} & 23.78\scriptsize{$\pm$0.5} & 25.86\scriptsize{$\pm$0.6} \\
    ERD \cite{ERD} & YOLO11n & 44.60\scriptsize{$\pm$0.6} & 39.40\scriptsize{$\pm$0.5} & 42.00\scriptsize{$\pm$0.3} \\
    \rowcolor{lightblue} \textbf{DuET (Ours)} & YOLO11n & \textbf{88.57}\scriptsize{$\pm$0.2} & \textit{\underline{41.92}}\scriptsize{$\pm$0.1} & \textbf{65.25}\scriptsize{$\pm$0.2} \\
    \hline
    Sequential FT & YOLO11x & 0.00\scriptsize{$\pm$0.0} & 18.16\scriptsize{$\pm$0.4} & 9.08\scriptsize{$\pm$0.3} \\
    LwF \cite{LwF} & YOLO11x & 19.57\scriptsize{$\pm$0.3} & 28.44\scriptsize{$\pm$0.5} & 24.01\scriptsize{$\pm$0.2} \\
    ERD \cite{ERD} & YOLO11x & 45.85\scriptsize{$\pm$0.6} & 37.38\scriptsize{$\pm$0.2} & 41.62\scriptsize{$\pm$0.8} \\
    \rowcolor{lightblue} \textbf{DuET (Ours)} & YOLO11x & 43.46\scriptsize{$\pm$0.2} & 38.37\scriptsize{$\pm$0.1} & 40.92\scriptsize{$\pm$0.2} \\
    \bottomrule
    \end{tabular}
    }
\end{table}

\begin{algorithm}[htbp]
\small
\caption{DuET Training Algorithm}
\label{alg:duet_training_algo}
\KwIn{Pre-trained model weights: \(\theta_0\), Sequence of tasks: \(\{T_1, T_2, \dots, T_T\}\).}
\KwOut{Final model weights: \(\theta_T\).}

\SetAlgoLined

Initialize model with pre-trained weights: \(\theta_0\)\;
\For{\(t = 1, 2, \dots, T\)}{
    \eIf{\(t = 1\)}{
        Train model on task \(T_1\) using \(\mathcal{L}_{\text{Detector}}\)\;
        Update weights: \(\theta_1 \gets \theta_0 - \eta \cdot \nabla_{\theta} \mathcal{L}_{\text{Detector}}\)\;
        Decompose weights: \\
        \(\theta_0 \rightarrow [\theta_{s_0}, \theta_{\tau_0}]\), \quad \(\theta_1 \rightarrow [\theta_{s_1}, \theta_{\tau_1}]\)\;
        Compute shared task vector: \(\tau_{s_1} = \theta_{s_1} - \theta_{s_0}\)\;
        Compute task-specific task vector: \(\tau_{\tau_1} \gets \theta_{\tau_1}\)\;
    }{
        Initialize: \(\theta_t \gets \theta_{t-1}\) \;
        Train model on task \(T_t\) using \(\mathcal{L}_{\text{Total}}\)\;
        Update weights: \(\theta_t \gets \theta_{t-1} - \eta \cdot \nabla_{\theta} \mathcal{L}_{\text{Total}}\)\;
        Decompose weights: \\
        \(\theta_{t-1} \rightarrow [\theta_{s_{t-1}}, \theta_{\tau_{t-1}}]\), \quad
        \(\theta_t \rightarrow [\theta_{s_t}, \theta_{\tau_t}]\)\;
        Compute shared task vectors: \\ 
        \(\tau_{\text{old}} = \tau_{s_{t-1}} = \theta_{s_{t-1}} - \theta_{s_0}\)\;
        \(\tau_{\text{curr}} = \tau_{s_t} = \theta_{s_t} - \theta_{s_0}\)\;
        Compute task-specific task vectors: \\
        \(\tau_{t_{\text{old}}} = \theta_{\tau_{t-1}}\), \quad \(\tau_{t_{\text{curr}}} = \theta_{\tau_{t}}\)\;
        Update shared weights using DuET:
        \((\theta_{s_t})_{\text{incre}} \gets \textbf{DuET}(\tau_{\text{curr}}, \tau_{\text{old}}, \theta_{s_0})\)\;
        Update task-specific weights:
        \((\theta_{\tau_t})_{\text{incre}} \gets [\tau_{t_{\text{old}}}, \tau_{t_{\text{curr}}}]\)\;
        Load new updated weights:
        \(\theta_t \gets [(\theta_{s_t})_{\text{incre}}, (\theta_{\tau_t})_{\text{incre}}]\)\;
    }
}
\Return \(\theta_T\)\;
\end{algorithm}
\begin{algorithm}[htbp]
\small
\caption{DuET Task Arithmetic Algorithm}
\label{alg:duet_ta_algo}
\SetAlgoLined
\KwIn{\textbf{Parameters:} Shared pre-trained weights: \(\theta_{s_0}\), Old Task Vector: \(\tau_{\text{old}}\), Current Task Vector: \(\tau_{\text{curr}}\) \newline
\textbf{Hyperparameters:} Limiting Factor: \(\gamma\), Base Scaling Coefficient: \(\alpha_{\text{base}}\), Numerical Stability constant: \(\epsilon\)}
\KwOut{Updated inremental shared weights: \(\theta_{s_t}^{\text{incre}}\)}

\SetAlgoLined
\For{\textbf{Model Layer:} \(l = 1,2,\dots,L\)}{
    \(
    p_l \;=\; \frac{\|\tau_{\text{old}}^l\| - \|\tau_{\text{curr}}^l\|}{\|\tau_{\text{old}}^l + \tau_{\text{curr}}^l\| + \epsilon},
    \)
    
    \( \delta_l = \gamma \cdot \tanh(p_l) \)
    
    \( \alpha_l = \alpha_{\text{base}} + \text{clamp} \ (\delta_l, -\gamma, \gamma) \)
    
    \( \beta_l = 1 - \alpha_l \)
    
    \( (\theta_{s_t}^l)_{\text{incre}} = \theta_{s_0}^l + \alpha_l \cdot \tau_{\text{old}}^l + \beta_l \cdot \tau_{\text{curr}}^l \)
}

\( (\theta_{s_t})_{\text{incre}} = \{(\theta_{s_t}^l)_{\text{incre}}\}_{l=1}^{L} \)

\Return \((\theta_{s_t})_{\text{incre}}\)
\end{algorithm}
\begin{table*}[!ht]
    \centering
    \caption{Results of various methods on Daytime Sunny [1:4] $\rightarrow$ Night Rainy [5:7] with different base detectors. Among columns, best in \textbf{bold}, second best \textit{\underline{underlined}}. Training time is reported on single NVIDIA A100-PCIE-40GB.}
    \label{E7_exps_table}
    \vspace{-2mm}
    \resizebox{\textwidth}{!}{
    \setlength{\tabcolsep}{1.8pt}
    \begin{tabular}{ccc||c|c|c|c|c||ccc}
    \toprule
        \multirow{4}{*}{\textbf{Method}} & \multirow{4}{*}{\textbf{Base Detector}} & \multirow{4}{*}{\makecell{\textbf{Training} \\ \textbf{Time} \\ \textbf{(\(\mathcal{T}_1 + \mathcal{T}_2\)}) \\ (hours)}} &
        \multirow{4}{*}{\makecell{\textbf{T1} \\ \textbf{Daytime} \\ \textbf{Sunny} \\ \textbf{{[1:4]}}}} &
        \multicolumn{4}{c||}{\textbf{T2: Night Rainy [5:7]}} &
        \multirow{4}{*}{\makecell{\textbf{Avg RI} \\ \textbf{(\%)}}} & 
        \multirow{4}{*}{\makecell{\textbf{Avg GI} \\ \textbf{(\%)}}} & 
        \multirow{4}{*}{\makecell{\textbf{RAI} \\ \textbf{(\%)}}} \\
        \cline{5-8}
        & & & & \makecell{\textbf{Old}} &  
        \makecell{\textbf{New}} &
        \multicolumn{2}{c||}{\textbf{Unseen}} \\
        \cline{5-8}
        & & & & \multirow{2}{*}{\makecell{\textbf{Daytime} \\ \textbf{Sunny [1:4]}}} & 
        \multirow{2}{*}{\makecell{\textbf{Night} \\ \textbf{Rainy [5:7]}}} & 
        \multirow{2}{*}{\makecell{\textbf{Night} \\ \textbf{Rainy [1:4]}}} & 
        \multirow{2}{*}{\makecell{\textbf{Daytime} \\ \textbf{Sunny [5:7]}}} & & & \\
        & & & & & & & & & & \\
        \hline
        \hline
        LDB \cite{LDB} & VitDet & 23.73 & 45.3\tiny{$\pm$0.6} & 1.40\tiny{$\pm$0.3} & 8.10\tiny{$\pm$0.4} & 0.02\tiny{$\pm$0.5} & 16.10\tiny{$\pm$0.7} & 3.09\tiny{$\pm$0.2} & 21.02\tiny{$\pm$0.3} & 12.05\tiny{$\pm$0.6} \\
        \rowcolor{lightblue} \textbf{DuET (Ours)} & VitDet & 24.13 & 45.3\tiny{$\pm$0.2} & 2.02\tiny{$\pm$0.3} & 16.10\tiny{$\pm$0.1} & 1.58\tiny{$\pm$0.3} & 15.20\tiny{$\pm$0.2} & 4.46\tiny{$\pm$0.2} & 22.47\tiny{$\pm$0.1} & 13.47\tiny{$\pm$0.2} \\
        \hline
        CL-DETR \cite{CL-DETR} & Deformable DETR & 91.96 & 46.3\tiny{$\pm$0.4} & 26.26\tiny{$\pm$0.4} & 9.75\tiny{$\pm$0.5} & 7.41\tiny{$\pm$0.6} & 14.88\tiny{$\pm$0.3} & 56.72\tiny{$\pm$0.4} & \textit{\underline{53.93}}\tiny{$\pm$0.2} & 55.33\tiny{$\pm$0.5} \\
        \rowcolor{lightblue} \textbf{DuET (Ours)} & Deformable DETR & 92.63 & 46.3\tiny{$\pm$0.2} & 26.75\tiny{$\pm$0.1} & 3.33\tiny{$\pm$0.2} & 7.98\tiny{$\pm$0.2} & 13.9\tiny{$\pm$0.1} & \textit{\underline{57.78}}\tiny{$\pm$0.2} & \textbf{54.58}\tiny{$\pm$0.1} & \textit{\underline{56.18}}\tiny{$\pm$0.2} \\
        \hline
        CL-DETR \cite{CL-DETR} & RT-DETR-l & 33.87 & 57.2\tiny{$\pm$0.4} & 5.11\tiny{$\pm$0.5} & 28.1\tiny{$\pm$0.6} & 5.04\tiny{$\pm$0.3} & 14.3\tiny{$\pm$0.4} & 8.93\tiny{$\pm$0.2} & 20.82\tiny{$\pm$0.5} & 14.88\tiny{$\pm$0.4} \\
        \rowcolor{lightblue} \textbf{DuET (Ours)} & RT-DETR-l & 33.89 & 57.2\tiny{$\pm$0.4} & 11.8\tiny{$\pm$0.1} & 14.90\tiny{$\pm$0.2} & 12.50\tiny{$\pm$0.2} & 17.30\tiny{$\pm$0.1} & 20.63\tiny{$\pm$0.2} & 36.50\tiny{$\pm$0.1} & 28.57\tiny{$\pm$0.2} \\
        \hline
        LwF \cite{LwF} & YOLO11n & 11.01 & 49.4\tiny{$\pm$0.2} & 21.50\tiny{$\pm$0.4} & 0.17\tiny{$\pm$0.6} & 9.36\tiny{$\pm$0.3} & 0.55\tiny{$\pm$0.5} & 43.52\tiny{$\pm$0.3} & 17.78\tiny{$\pm$0.7} & 30.65\tiny{$\pm$0.6} \\
        ERD \cite{ERD} & YOLO11n & 11.50 & 49.4\tiny{$\pm$0.2} & 25.60\tiny{$\pm$0.4} & 16.70\tiny{$\pm$0.3} & 12.20\tiny{$\pm$0.6} & 17.60\tiny{$\pm$0.7} & 51.82\tiny{$\pm$0.5} & 38.96\tiny{$\pm$0.3} & 45.39\tiny{$\pm$0.4} \\
        \rowcolor{lightblue} \textbf{DuET (Ours)} & YOLO11n & 11.02 & 49.4\tiny{$\pm$0.2} & 43.30\tiny{$\pm$0.1} & 2.67\tiny{$\pm$0.3} & 14.00\tiny{$\pm$0.2} & 8.93\tiny{$\pm$0.1} & \textbf{87.65}\tiny{$\pm$0.2} & 34.18\tiny{$\pm$0.1} & \textbf{60.92}\tiny{$\pm$0.2} \\
    \bottomrule
    \end{tabular}
    }
\end{table*}
\end{document}